\documentclass{article}

\PassOptionsToPackage{numbers, compress}{natbib}

\usepackage{microtype}
\usepackage{graphicx}
\usepackage{subfigure}
\usepackage{booktabs} 

\usepackage{hyperref}

\usepackage[utf8]{inputenc} 
\usepackage[T1]{fontenc}    
\usepackage{url}            
\usepackage{amsfonts}       
\usepackage{nicefrac}       
\usepackage[usenames,dvipsnames]{xcolor}
\usepackage{amsmath, amssymb, mathtools, amsthm,bm}
\usepackage{bbm}
\usepackage{algorithm} 
\usepackage{algorithmic} 
\usepackage{multirow}
\usepackage{float}
\usepackage[font=footnotesize,skip=2pt,figurename=Fig.]{caption}

\newcommand{\xhdr}[1]{\paragraph*{
  \bf {#1}}}
\newcommand{\omt}[1]{}
\newcommand{\ignore}[1]{}

\newtheorem{theorem}{Theorem}[section]
\newtheorem{lemma}[theorem]{Lemma}

\newtheorem{definition}[theorem]{Definition}
\newtheorem{exmpl}{Example}
\newtheorem{remark}[theorem]{Remark}

\newenvironment{sproof}{%
  \proof}{\endproof}

\makeatletter
\newenvironment{routine}[1][htb]{%
    \renewcommand{\ALG@name}{Routine}
   \begin{algorithm}[#1]%
  }{\end{algorithm}}
\makeatother

\newcommand{\rewfn}{r}            
\newcommand{\Totalrew}{U}         
\newcommand{\fairfn}{\pi}         
\newcommand{\fairc}{\tau}         
\newcommand{\Totalu}{U}           

\newcommand{\rewob}{\tilde{\rewfn}}            
\newcommand{\status}{\text{fairness measure }}

\newcommand{\action}{action-dependent bandits}
\newcommand{\history}{history-dependent bandits}
\newcommand{\vp}{\bm{\mathrm{p}}}
\newcommand{\vq}{\bm{\mathrm{q}}}

\newcommand{\va}{\bm{\mathrm{a}}}

\newcommand{\vl}{\bm{\mathrm{l}}}
\newcommand{\vr}{\bm{\mathrm{r}}}
\newcommand{\vf}{\bm{\mathrm{f}}}

\newcommand{\N}{\mathbb{N}}
\newcommand{\R}{\mathbb{R}}

\newcommand{\cA}{\mathcal{A}}
\newcommand{\cB}{\mathcal{B}}
\newcommand{\cD}{\mathcal{D}}

\newcommand{\cE}{\mathcal{E}}
\newcommand{\cH}{\mathcal{H}}
\newcommand{\cM}{\mathcal{M}}
\newcommand{\cI}{\mathcal{I}}
\newcommand{\cJ}{\mathcal{J}}
\newcommand{\cO}{\mathcal{O}}
\newcommand{\cP}{\mathcal{P}}
\newcommand{\cS}{\mathcal{S}}

\newcommand{\cL}{\mathcal{L}}
\newcommand{\cT}{\mathcal{T}}

\newcommand{\Esymb}{\mathbb{E}}
\newcommand{\Psymb}{\mathbb{P}}

\newcommand{\Rsymb}{\mathbb{R}}
\newcommand{\Nsymb}{\mathbb{N}}
\DeclareMathOperator*{\E}{\Esymb}

\DeclareMathOperator*{\argmax}{arg\,max}
\DeclareMathOperator*{\argmin}{arg\,min}

\renewcommand{\hat}{\widehat}

\newcommand{\UCB}{\texttt{UCB}}
\newcommand{\est}{\texttt{est}}
\newcommand{\err}{\texttt{err}}
\newcommand{\DE}{\texttt{DE}}

\newcommand{\Ber}{\texttt{Bernoulli}}

\newcommand{\KL}{\texttt{KL}}
\newcommand{\Np}{N_\texttt{p}}

\newcommand{\Esymbr}{\mathbb{E}_{\texttt{rand}}}

\newcommand{\CombLipBC}{\texttt{CombLipBwC}}
\newcommand{\CombBC}{\texttt{CombBwC}}

\newcommand{\Beta}{\texttt{Beta}}
\newcommand{\suf}{\texttt{suf}}
\newcommand{\und}{\texttt{und}}

\def\DUCB{{\sf \scalebox{0.5}{DUCB}}}

\def\reg{\mathrm{Reg}}

\newcommand{\overbar}[1]{\mkern 1.5mu\overline{\mkern-1.5mu#1\mkern-1.5mu}\mkern 1.5mu}

\newcommand{\squishlist}{
\begin{list}{{{\small{$\bullet$}}}}
{\setlength{\itemsep}{3pt}      \setlength{\parsep}{1pt}
\setlength{\topsep}{1pt}       \setlength{\partopsep}{0pt}
\setlength{\leftmargin}{1em} \setlength{\labelwidth}{1em}
\setlength{\labelsep}{0.5em} } }
\newcommand{\squishend}{  \end{list}}

\newcounter{relctr} 
\everydisplay\expandafter{\the\everydisplay\setcounter{relctr}{0}} 
\newcommand\labelrel[2]{%
  \begingroup
    \refstepcounter{relctr}%
    \stackrel{\textnormal{(\alph{relctr})}}{\mathstrut{#1}}%
    \originallabel{#2}%
  \endgroup
}
\AtBeginDocument{\let\originallabel\label} 

\usepackage[final]{neurips_2021}

\title{Bandit Learning with Delayed Impact of Actions}

\author{
  Wei Tang$^\dag$, Chien-Ju Ho$^\dag$, and Yang Liu$^*$\\
  $^\dag$Washington University in St. Louis,  $^*$University of California, Santa Cruz\\
  \texttt{\{w.tang, chienju.ho\}@wustl.edu, yangliu@ucsc.edu}
}

\begin{document}

\maketitle

\begin{abstract}

We consider a stochastic multi-armed bandit (MAB) problem with delayed impact of actions. In our setting, actions taken in the past
impact the arm rewards in the subsequent future.
This delayed impact of actions is prevalent in the real world.
For example, the capability to pay back a loan for people in a certain social group might depend on historically how frequently that group has been approved loan applications. 
If banks keep rejecting loan applications to people in a disadvantaged group, it could create a feedback loop and further damage the chance of getting loans for people in that group. 
In this paper, we formulate this delayed and long-term impact of actions within the context of multi-armed bandits. We generalize the bandit setting to encode the dependency of this ``bias" due to the action history during learning. 
The goal is to maximize the collected utilities over time while taking into account the dynamics created 
by the delayed impacts of historical actions. 
We propose an algorithm that achieves a regret of $\tilde{\cO}(KT^{2/3})$ and show a matching regret lower bound of $\Omega(KT^{2/3})$, where $K$ is the number of arms and $T$ is the learning horizon. Our results complement the bandit literature by adding techniques to deal with actions with long-term impacts and have implications in designing fair algorithms.

\end{abstract}

\section{Introduction}
Algorithms have been increasingly involved in high-stakes decision making. 
Examples include 
approving/rejecting loan applications~\citep{fuster2018predictably,kleinberg2018discrimination}, 
deciding on employment and compensation~\citep{bartik2016credit,cowgill2009incentive}, 
and recidivism and bail decisions~\citep{angwin2016machine}.
Automating these high-stakes decisions has raised ethical concerns on whether it amplifies the discriminative bias against protected classes~\citep{obermeyer2019dissecting,chouldechova2017fair}. 
There have also been growing efforts towards studying algorithmic approaches to mitigate these concerns.
Most of the above efforts have focused on static settings:
a utility-maximizing decision maker needs to ensure her actions satisfy some fairness criteria at the decision time, 
without considering the long-term impacts of actions.
However, in practice, these decisions may often introduce long-term impacts to the rewards and well-beings for the human agents involved.
For example,

\squishlist
    \item A regional financial institute may decide on the fraction of loan applications from different social groups to approve. 
    These decisions could affect the development of these groups: 
    The capability of applicants from a group to pay back a loan might depend on the group's socio-economic status, 
    which is influenced by how frequently applications from this group have been approved~\citep{bartlett2018consumer,cowgill2019economics}.
    \item The police department may decide on the amount of patrol time or the probability of patrol in a neighborhood (primarily populated with a demographic group). 
    The likelihood to catch a crime in a neighborhood might depend on how frequent the police decides to patrol this area~\citep{goel2016precinct,gelman2007analysis}. 
\squishend

These observations raise the following concerns. If being insensitive with the long-term impact of actions, the decision maker risks treating a historically disadvantaged group unfairly. 
Making things even worse, these unfair and oblivious decisions might reinforce existing biases and make it harder to observe the true potential for a disadvantaged group. While being a relatively under-explored (but important) topic, several recent works have looked into this problem of delayed impact of actions in algorithm design. However, these studies have so far focused on understanding the impact in a one-step delay of actions~\citep{liu2018delayed, kannan2019downstream, heidari2019long}, or a sequential decision making setting without uncertainty~\citep{liufair,hu2018short,mouzannar2019fair, zhang2019group, liu2020disparate,d2020fairness,zhang2020fair}. 

Our work departs from the above line of efforts by studying the long-term impact of actions in sequential decision making under uncertainty.
We generalize the multi-armed bandit setting by introducing the \emph{impact functions} that encode the dependency of the ``bias'' due to the action history of the learning to the arm rewards. Our goal is to learn to maximize the rewards obtained over time, in which the rewards' evolution could depend on the past actions. 

The history-dependency reward structure makes our problem substantially more challenging. In particular, we first show that applying standard bandit algorithms leads to linear regret, i.e., existing approaches will obtain low rewards with a biased learning process.
To address this challenge, under relatively mild conditions for the dependency dynamics, we present an algorithm, based on a phased-learning template which smoothes out the historical bias during learning, that achieves a regret of $\tilde{\cO}(KT^{2/3})$. Moreover, we show a matching lower regret bound of $\Omega(KT^{2/3})$ that demonstrates that our algorithm is order-optimal. Finally, we conduct a series of simulations showing that our algorithms compare favorably to other state-of-the-art methods proposed in other application domains.
From a policy maker's point of view, our paper explores solutions to learn the optimal sequential intervention when the actions taken in the past impact the learning environment in an unknown and long-term manner. We believe our work nicely complements the existing literature that focuses more on the ``understanding'' of the dynamics~\citep{hu2018short,liu2018delayed,zhang2019group,zhang2020fair}.

\xhdr{Related work.}
Our work contributes to algorithmic fairness studied in sequential settings.
Prior works either study fairness in sequential learning settings without considering long-term impact of actions~\cite{joseph2016fairness, liu2017calibrated,gillen2018online,bechavod2019equal,  gupta2019individual,patil2020achieving} or explore the delayed impacts of actions with focus on addressing the one-step delayed impacts or sequential learning with full information~\citep{hu2018short,liu2018delayed,bartlett2018consumer, heidari2019long,mouzannar2019fair,cowgill2019economics,d2020fairness}.
Our work differs from the above and studies delayed impacts of actions in sequential decision making under uncertainty.
Our formulation bears similarity to reinforcement learning since our impact function encodes memory 
(and is in fact Markovian~\citep{ortner2012regret,tekin2010online}), although
we focus on studying the exploration-exploitation tradeoff in bandit formulation.
Our learning formulation builds on the rich bandit learning literature~\citep{lai1985asymptotically,auer2002finite} and is related to non-stationary bandits~\cite{slivkins2008adapting,besbes2014stochastic,besbes2015non,levine2017rotting,kleinberg2018recharging}.
Our techniques share similar insights with Lipschitz bandits~\citep{kleinberg2008multi,slivkins2014contextual} and combinatorial bandits~\citep{chen2016combinatorial} in that we also assume the Lipschitz reward structure and consider combinatorial action space.
There are also recent works that have formulated delayed action impact in bandit learning~\citep{pike2018bandits,kleinberg2018recharging}, but in all of these works, the setting and the formulation are different from the ones we consider in the present work. 
More discussions on related work can be found in Appendix~\ref{sec:related_full}.

\section{Problem Setting}
\label{sec:setting}

We formulate the setting in which an institution sequentially determines how to allocate resource to different groups. 
For example, a regional financial institute may decide on 
overall frequency of loan applications to approve from different social groups. 
The police department may decide on the amount of patrol time 
allocated to different regions.

The institution is assumed to be a utility maximizer, aiming to maximize the expected reward associated with the allocation policy over time.
If we assume the reward\footnote{The reward could be whether a crime has been stopped or whether the borrower pays the monthly payment on time. For applications that require longer time periods to assess the rewards, the duration of a time step, i.e., the frequency to update the policy, would also need to be adjusted accordingly.} 
for allocating a unit of resource to a group is i.i.d. drawn from some unknown distribution,
this problem can be reduced to a standard bandit problem,
with each group representing an \emph{arm}. 
The goal of the institution is then to learn a sequence of arm selections to maximize its cumulative rewards.

In this work, we extend the bandit setting and consider the delayed impact of actions.
Below we formalize our setup which introduces \emph{impact functions} to bandit framework.

\textbf{Action space.}
There are $K$ \emph{base arms}, indexed from $k=1$ to $K$, with each base arm representing a group.
At each discrete time $t$, the institution chooses an action, called a \emph{meta arm},
which specifies the probability to activate each base arm.
Let $\cP = \Delta([K])$ be the $(K-1)$-dimensional probability simplex.
We denote the meta arm as $\vp(t)=\{p_1(t),\ldots,p_K(t)\}\in\cP$.
Each base arm $k$ is activated independently according to their $p_k(t)$ in $\vp(t)$. 
The institution only observes the reward from the arms that are activated. 
Our feedback model deviates slightly from the classical bandit feedback and shares similarity to combinatorial bandits:
instead of assuming always observing one arm's reward each time, we observe the reward of one arm in expectation, i.e., we can potentially observe no arm's reward or multiple arms' rewards. 
This modeling choice is mainly needed to resolve a technicality issue and has been adopted in the literature~\cite{wang2017improving,chen2016combinatorial}.
Most of our algorithms and results extend to the case where only one arm is activated according to $\vp(t)$.\footnote{The upper bound becomes  $\tilde{\cO}(K^{4/3}T^{2/3})$, which is slightly worse than  $\tilde{\cO}(KT^{2/3})$ in $K$. But the upper bound is still tight in $T$. We provide discussions in Remark~\ref{rmk_new_Hoeffding}}

\begin{remark}
We can also interpret the meta-arm as specifying the proportion of resources allocated to each base arm. 
The interpretation impacts the way the rewards are generated (i.e., instead of observing the rewards of the realized base arms, the institution observes the rewards of all base arms with non-zero allocations.)  
Our analysis utilizes the idea of importance weighting and could deal with both cases in the same framework.
To simplify the presentation, we focus on the case of interpreting the meta-arm as probabilities, though our results apply to both interpretations.
\end{remark}

\textbf{Delayed impacts of actions.}
We consider the scenario in which the rewards of actions are unknown a priori and are influenced by the action history.
Formally, let $\cH(t)=\{\vp(s)\}_{s\in[t]}$ be the action history at time $t$. 
We define the \emph{impact function} \underline{$\bm{\mathrm{f}}(t) = F(\cH(t))$} to summarize the impact of the learner's actions to the reward generated in each group, where $F(\cdot)$ is the function mapping the action history to its current impact on arms' rewards.
In the following discussion, we make $F(\cdot)$
implicit and use the vector $\bm{\mathrm{f}}(t)=\{f_1(t),\ldots,f_K(t)\}$ to denote the impact to each group, where $f_k(t)$ captures the impact of action history to arm $k$.

\textbf{Rewards and regret.}
The reward for selecting group $k$ at time $t$ depends on both $p_k(t)$ and the historical impact $f_k(t)$. 
In particular, when the arm representing group $k$ is activated, 
the institution observes a reward (the instantaneous reward is bounded within $[0, 1]$) drawn i.i.d. from a unknown distribution with mean \underline{$\rewfn_k\left(f_k(t)\right)\in[0,1]$} and claims the sum of rewards from activated arms as total rewards.
$\rewfn_k(\cdot)$ is unknown a priori but is Lipschitz continuous (with known Lipschitz constant $L_k \in(0,1]$) with respect to its input, 
i.e., a small deviation of the institution's actions has small 
impacts on the unit reward from each group.
When action $\vp(t)$ is taken at time $t$,
the institution obtains an expected reward 
\begin{align}\label{eq:reward-structure}
    \Totalrew_t(\vp(t)) = \sum_{k=1}^K p_k(t) \cdot \rewfn_k \left(f_k(t)\right).
\end{align}
As for the impact function, 
we focus on the setting in which $\bm{\mathrm{f}}(t)$ is a time-discounted average, with each component $f_k(t)$ defined as
\begin{align}\label{fn:impact_fn}
    f_k(t) = \frac{\sum_{s=1}^{t} p_k(s) \gamma^{t-s} }{\sum_{s=1}^{t} \gamma^{t-s}},
\end{align}
where $\gamma\in[0,1)$ is the time-discounting factor.
\footnote{Here we follow the tradition to define $0^0 = 1$ when $\gamma = 0$.} 
Intuitively, $f_k(t)$ is a weighted average with more weights on recent actions. 
We would like to highlight that our results extend to a more general family of impact functions and do not require the exact knowledge of impact functions
(see discussion in Section~\ref{sec:history-discussion}). We also note that when $\gamma =0$, our setting reduces to a special case where the impact function only depends on the current action $p_k(t)$ (\emph{action dependent}), instead of the entire history of actions (discounted by 0 right away). 
We study this special case of interest in Section~\ref{sec:action}.

Let $\cA$ be the algorithm the institution deploys. 
The goal of $\cA$ is to choose a sequence of actions $\{\vp(t)\}$ that maximizes the total utility.
The performance of $\cA$ is characterized by regret, defined as
\begin{align}\label{reg_defn}
    \reg(T) = \sup_{\vp \in \cP} \sum_{t=1}^T \Totalu_t(\vp) - \mathbb{E}\bigg[ \sum_{t = 1}^T \Totalu_t(\vp(t))\bigg],
\end{align}
where the expectation is taken on the randomness of algorithm $\cA$ and the utility realization.\footnote{In this paper, we adopt the standard regret definition and compare against the optimal fixed policy. Another possible regret definition is to compare against the optimal dynamic policy that could change based on the history.  However, calculating the optimal dynamic policy in our setting is nontrivial as it requires to solve an MDP with continuous states.  
}

\subsection{Exemplary Application of Our Setup}
We provide an illustrative example to instantiate our model.
Consider a police department who needs to dispatch a number of police officers to $K$ different districts. 
Each district has a different crime distribution, and the goal (absent additional fairness
constraints) might be to maximize the number of crimes caught~\citep{elzayn2019fair}.~\footnote{
As discussed by \citet{elzayn2019fair}, there might be other goals besides simply catching criminals, including preventing crime, fostering community relations, and promoting public safety. We use the same goal they adopted for the illustrative purpose.}
The effects of police patrol resource allocated to each district may aggregate over time and then impact the crime rate of that district. 
In other words, the crime rate in each district depends on how frequently the police officers have been dispatched historically in this district. 

To simplify the discussion, we normalize the expected police resource to be one unit.
Each district $k$ has a default average crime rate $\overline{r}_k  \in (0, 1)$ at the beginning of the learning process.
This crime rate can (at most) be decreased to $\underline{r}_k \in (0, \overline{r}_k)$.
All of these are unknown to the police department. 
The police department makes a resource allocation decision at each time step. 
We use $r_k(t)\in(0, 1)$ to denote the crime rate in district $k$ at time $t$, taking into account the impact of historical decisions. 
Assume $p_k(t)$ is the amount of police resource dispatched to district $k$ at time $t$ ($\sum_k p_k(t)=1$ for all $t$), the expected number of crimes caught at district $k$ at time $t$ would be $p_k(t)r_k(t)$. 
Note that here $p_k(t)$ can be interpreted as the probability of allocating police resource (randomly sending the patrol team to each of the $K$ districts) or the fraction of allocated police resource.

Below we provide one natural example of the interaction between the impact function and the reward.
At time step $t+1$, let $\cH_k(t):=\{p_k(1), \ldots, p_k(t)\}$ denote the historical decisions of the police department for district $k$.
Now given $\cH_k(t+1) = \{\cH_k(t) \cup p_k(t+1)\}$ where $p_k(t+1)$ is the current decision for district $k$, assume that the crime rate at time $t+1$ in district $k$ is in the following form:
\begin{align}
    r_k(t+1) = \overline{r}_k - f_k(\cH_k(t+1)) \times (\overline{r}_k - \underline{r}_k), \label{payback_rate}
\end{align}
where $f_k(\cdot): [0, 1]^t \rightarrow [0, 1]$ is the impact function that summarizes how historical actions would impact the current crime rate.
One possible example is $f_k(\cH_k(t)) = \frac{\sum_{s= 1}^t p_k(s)\gamma^{t-s}}{\sum_{s = 1}^t \gamma^{t-s}}$ as we defined in Equation~\eqref{fn:impact_fn}.
This impact function has two natural properties:
\squishlist
    \item When $f_k(\cH_k(t)) = 1$ (e.g., $p_k(s) = 1, \forall s\le t$), the police department keeps dispatching the police officers to district $k$ with probability $1$, then district $k$ will reach its lowest crime rate.
    \item When $f_k(\cH_k(t)) \rightarrow 0$ (e.g., $p_k(s) \rightarrow 0, \forall s\le t$), the police department rarely dispatch police officers to district $k$,  The crime rate in district $k$ will reach its highest level.
\squishend

In this example, treating each district as an arm and directly applying standard bandit algorithms might reach suboptimal solutions since the reward dynamic is not considered.
In this paper, we develop algorithms that can take into account this history-dependent reward dynamic and achieve no-regret learning. Our results hold for a general class of impact functions (under mild conditions) and do not need to assume the exact knowledge of the impact function.

\section{Overview of Main Results}
\label{sec:neg_res}
We summarize our main results in this section.
First, we present an important, though perhaps not surprising, negative result:
if the institution is not aware of the delayed impact of actions,
applying existing standard bandit algorithms in our setting leads to linear regrets.
This negative result highlights the importance of designing new algorithms when delayed impact of actions are present.
The formal statement and analysis are in Appendix~\ref{app:negative_res}.
\begin{lemma}[Informal]\label{lem:negative-result} 
If the institution is unaware of the delayed impact of actions, applying standard bandit algorithms (including UCB, Thompson Sampling) leads to linear regrets.
\end{lemma}

While the negative result might not be surprising, as it resembles similarity to the negative results on applying classic bandit algorithm to a non-stationary setting, it points out the need to design new algorithms for settings with delayed impact of actions.
The key challenge introduced by our setting is in estimating the arm rewards: when pulling the same meta arm at different time steps, the institution does not guarantee to obtain rewards drawn from the targeted distribution according to the chosen meta arm, as the arm reward depends on the impact function $\vf(t)$. 
To address this challenge, we note that if the institution keeps pulling the same meta-arm repeatedly, the impact function (and thus the arm reward associated with the meta-arm) would converge to some value.
This observation leads to our approaches. 
We first develop a bandit algorithm that works with impacts that converge ``immediately" (or equivalently only depend on ``immediate'' actions, echoing the case with $\gamma = 0$ in Equation~\eqref{fn:impact_fn}).
We then propose a phased-learning reduction template that reduces our general setting to the above one 
and achieves a sublinear regret.

\begin{theorem}[Informal]
There is an algorithm that achieves an optimal regret bound $\tilde{\cO}(KT^{2/3})$ for the bandit problem with the impact function defined in Equation~\eqref{fn:impact_fn}. In addition, there is a matching lower bound of $\Omega(KT^{2/3})$.
\end{theorem}

To provide an overview of our approaches, 
we start with \emph{\action} (Section~\ref{sec:action}), where the impact at time $t$  depends only on the action at $t$, i.e., $\vf(t)=\vp(t)$, namely $\gamma = 0$ in Equation~\eqref{fn:impact_fn}.
This setting not only captures the one-step impact but also offers a backbone for the phase-learning template for the general history-dependent scenario. 
In this setting, when a meta-arm $\vp=\{p_1,\ldots,p_K\}$ is selected, 
each base arms $k$ is activated with probability $p_k$, 
and the institution observes the realized rewards for all activated base arms
and receives the sum of them as total rewards.
Since we know the probability $p_k$ for activating each base arm, 
we may apply importance weighting to simulate the case as if the learner is selecting $K$ probabilities and obtain $K$ signals at each time step.
This interpretation transforms our problem structure to a setting similar to combinatorial bandits.
Furthermore, since both $\rewfn_k(\cdot)$ are Lipschitz continuous, we adopt the idea from Lipschitz bandits to discretize the continuous space of each $p_k$.
With these ideas combined, we design a UCB-like algorithm that achieves a regret of
\underline{$\cO(KT^{2/3}(\ln T)^{1/3})$}.

With the solution of \action, we explore the general \emph{\history}  with impact functions following Equation~\eqref{fn:impact_fn} (Section~\ref{sec:history}).
The main idea is to divide total time rounds into phases, and then selecting the same actions in each phase to smooth out impacts of historically made actions, which will then help reduce the problem to an action-dependent one.
One challenge is to construct appropriate confidence bound and adjust the length of each phase to account for the historical action bias.
With a careful combination with our results for \action, we present an algorithm which can also achieve a regret of the order \underline{$\tilde{\cO}(KT^{2/3})$}. 
We further proceed to show that this bound is tight and provide numerical experiments.

\section{Action-Dependent Bandits}
\label{sec:action}

In this section, we study \action, in which the impact function $\bm{\mathrm{f}}(t)=\vp(t)$, corresponding to $\gamma = 0$ in Equation~\eqref{fn:impact_fn}.
Our algorithm starts with a discretization over the space $\cP$.
Formally, we uniformly discretize $[0,1]$ for each base arm into intervals of a fixed length $\epsilon$, with carefully chosen $\epsilon$ such that $1/\epsilon$ is an positive integer.\footnote{Smarter discretization generally does not lead to better regret bounds~\cite{kleinberg2008multi}.}
Let $\cP_\epsilon$ be the space of discretized meta arms,
i.e., for each $\vp =\{p_1,\ldots,p_K\}\in\cP_\epsilon$, 
$\sum_{k=1}^K p_k=1$ and $p_k \in \{\epsilon, 2\epsilon, \ldots, 1\}$ for all $k$.
Let $\vp_\epsilon^* := \sup_{\vp \in \cP_\epsilon} \sum_{k=1}^K p_k \cdot \rewfn_k(p_k)$
denote the optimal strategy in discretized space $\cP_\epsilon$. 
After a meta arm $\vp(t)=\{p_1(t),\ldots,p_K(t)\}\in\cP_\epsilon$ is selected,
each arm $k$ is {\em independently} activated with probability $p_k(t)$.
From now, we use $\rewob_t(\cdot)$ to denote the realization 
of corresponding reward.
The learner observes activated arms, 
and observes the instantaneous reward $\rewob_t(p_k(t))$ of 
each activated arm $k$. 
We use importance weighting~\cite{gretton2009covariate} to construct the unbiased realized reward for each of the $K$ elements in $\vp$: 
\begin{equation}\label{eq:reward_impor_weights}
\hat{\rewfn}_t(p_k(t))  = \left \{
  \begin{aligned}
    & \rewob_t(p_k(t))/{p_k(t)}, && \text{arm } k \text{ is activated} \\
    & 0. && \text{arm } k \text{ is not activated}
  \end{aligned} \right.
\end{equation} 
Since the probability activating arm $k$ is $p_k(t)$, it is easy to see that $\Esymb[\hat{\rewfn}_t(p_k(t))] = \Esymb[\rewob_t(p_k(t))]$. 
Given the importance-weighted rewards $\{\hat{\rewfn}_t(p_k(t))\}$,
we re-frame our problem as choosing a $K$-dimensional probability measure (one value for each base arm).
In particular, for each base arm $k$, $p_k$ will take the value from $\{\epsilon, 2\epsilon,\ldots,1\}$, and we refer to $p_k$ as the \emph{discretized arm}.

\begin{remark}
The above importance-weighting technique enables us to ``observe'' samples of $\rewfn_k(p_k)$ for all base arms $k$ when selecting $\vp=\{p_1,\ldots,p_K\}$. 
This technique helps to bridge the gap between the interpretation of whether $\vp$ is a probability distribution or an allocation over base arms.
Our following techniques can be applied in either interpretation.
\end{remark}

\begin{algorithm}[h]
\caption{Action-Dependent UCB} 
\label{algo:action_UCB}
\begin{algorithmic}[1]
\STATE \textbf{Input:} $K$, $\epsilon$
\STATE \textbf{Initialization:} 
For each discretized arm, play an arbitrary meta arm such that this discretized arm is included (if the selection of the arm is not realized, then simply initialize its reward to $0$; otherwise initialize it to the observed reward divided/reweighted by the selection probability).
\FOR{$t = \left \lceil K/\epsilon \right \rceil + 1, ..., T$}
  \STATE Select $\vp(t) = \argmax_{\vp\in \cP_\epsilon} \UCB_t(\vp)$ where $\UCB_t(\vp)$ is defined as in~\eqref{eq:mean-update}.
  \STATE Arm $k$ is activated w.p. $p_k(t)$ and observe its realized reward $\rewob_t(p_{k}(t))$. 
  \STATE Update the importance-weighted rewards $\{\hat{\rewfn}_t(p_k(t))\}$ as in~\eqref{eq:reward_impor_weights} and update the empirical mean $\{\bar{\rewfn}_t(p_k(t))\}$
  for each base arm as in~\eqref{eq:mean-update}.
\ENDFOR
\end{algorithmic}
\end{algorithm}
By doing so, our problem is now similar to combinatorial bandits,
in which we are choosing $K$ discretized arms and observe the corresponding rewards. 
Below we describe our UCB-like algorithm based on the reward estimation 
of discretized arms.
We define the set $\cT_t(p_k) = \{s\in[t]: p_k\in\vp(s)\}$ to record all the time steps such that the deployed meta arm $\vp(s)$ contains the discretized arm $p_k$. 
We can maintain the empirical estimates of the mean reward 
for each discretized arm and compute the $\UCB$ index for each meta arm $\vp \in \cP_\epsilon$:
\begin{align}
\bar{\rewfn}_t(p_k) = \frac{\sum_{s\in\cT_t(p_k)}\hat{\rewfn}_s(p_k)} {n_t(p_k)}, \quad
\UCB_t(\vp) =  \sqrt{ \frac{K \ln t}{\min_{p_k \in \vp} n_t(p_k)}} + \sum_{p_k \in \vp} p_k \cdot \bar{\rewfn}_t(p_k),
\label{eq:mean-update}
\end{align} 
where $n_t(p_k)$ is the cardinality of set $\cT_t(p_k)$.
With the $\UCB$ index in place, we are now ready to state our algorithm in Algorithm~\ref{algo:action_UCB}.
The next theorem provides the regret bound of Algorithm \ref{algo:action_UCB}.
\begin{theorem} \label{theorem:key_theorem_vanilla}
Let $\epsilon=\Theta\big((\ln T/T)^{1/3}\big)$. 
The regret of Algorithm \ref{algo:action_UCB} (with respect to the optimal arm in non-discretized $\cP$) is upper bounded as follows:
$\reg(T) = \cO\big(KT^{2/3}(\ln T)^{1/3}\big)$.
\end{theorem}

\begin{sproof}
Similar to the proofs of the family of UCB-style algorithms for MAB, after an appropriate discretization,  we can derive the regret as the sum of the \emph{badness} (suboptimality of a meta arm) for all (discretized) suboptimal meta arm selection.
However, this will cost us an exponential $K$ in the order of final regret bound: this is because we need to take the summation over all feasible suboptimal meta arms, which the number grows exponentially with $K$. 
To tackle this challenge, we focus on the derivations of badness via tracking the \emph{minimum} suboptimal selections in the space of realized actions (base arms), which enables us to reduce the exponential $K$ to a polynomial $K$.
On a high level, our proof proceeds in the following steps:
\squishlist
\item In Step $1$, we obtain a high probability bound of the estimation error for the expected rewards of meta arms after discretization.
\item In Step $2$, we bound the probability on deploying a suboptimal meta arm when selected sufficiently many number of times, 
where we quantify such sufficiency via $\min_{p_k \in\vp} n_t(p_k)$, which is the minimum number of selection of a discretized arm contained in a suboptimal meta arm.
\item In Step $3$ and Wrapping-up step, we bound the expected value of $\min_{p_k \in\vp}n_t(p_k)$ and connect the regret for playing suboptimal meta arms $\vp$ with the regret incurred by including discretized arms $p_k \notin \vp_\epsilon^*$ which are not in optimal strategy (in discretized space).
\squishend
Finally, the regret bound of Algorithm~\ref{algo:action_UCB} can be 
achieved by optimizing the discretization parameter. 
\end{sproof}

\vspace{-8pt}
\xhdr{Discussions}
\label{sec:discussion-action-dependent}
Our techniques have close connections to Lipschitz bandits~\citep{combes2020unimodal,magureanu2014lipschitz} and combinatorial bandits~\citep{chen2016combinatorial,chen2016combinatorial2}.
Given the Lipschitz property of $\rewfn_k(\cdot)$,
we are able to utilize the idea of Lipschitz bandits to discretize the strategy space and achieve sublinear regret with respect to the optimal strategy in the non-discretized strategy space. 
Moreover, we achieve a significantly improved regret bound by utilizing the connection between our problem setting and combinatorial bandits.
In combinatorial bandits, the learner selects $K$ actions out of action space $\cM$  at each time step,
where $|\cM| = \Theta(K/\epsilon)$ in our setting.
Directly applying state-of-the-art combinatorial bandit algorithms~\citep{chen2016combinatorial} in our setting would achieve an instance-independent regret bound of 
$\cO\big(K^{3/4}T^{3/4}(\ln T)^{1/4}\big)$, 
while we achieve a lower regret of $\cO\big(KT^{2/3}(\ln T)^{1/3}\big)$.\footnote{We compare our results with a tight regret bound achieved in Theorem 2 of~\citep{chen2016combinatorial}. The detailed derivations are deferred to Appendix~\ref{cf_regret}.}
The reason for our improvement is that, for each base arm, regardless of which probability it was chosen, 
we can update the reward 
of the base arm, 
which provides information for all meta arms that select this arm with a different probability.
This reduces the exploration and helps achieving the improvement.
In addition to the above improvement, we would like to highlight that another of our main contributions is to extend the action-dependent bandits to the problem of history-dependent bandits, as discussed in Section~\ref{sec:history}. 

Another natural attempt to tackle our problem is to apply EXP3~\citep{auer2002nonstochastic}, which achieves sublinear regret even when the arm reward is generated adversarially.
However, note that the optimal policy in our setting could be a mixed strategy, while the ``sublinear'' regret of EXP3 is with respect to a fixed strategy.
Therefore, when applying EXP3 over the set of base arms, it still implies a linear regret in our setting. The other option is to apply EXP3 over the set of meta arms.
Since the number of meta arms is exponential in $K$, it would incur a regret exponential in $K$ due to the size of meta arms.

\section{History-Dependent Bandits} 
\label{sec:history}

We now describe how to utilize our results for \action~to
solve the history-dependent bandit learning problem, 
with the impact function specified in Equation~\eqref{fn:impact_fn}.
The crux of our analysis is the observation that, in \history, 
if the learner keeps selecting the same strategy $\vp$ for a long enough period of time,  
the expected one-shot utility will be approaching the utility of selecting $\vp$ in the \action.
More specifically, suppose after time $t$, the current action impact for all arms is $\vf(t) = \vp^{(\gamma)}(t) = \{p^{(\gamma)}_1(t), \ldots, p^{(\gamma)}_K(t)\}$. 
Assume that the learner is interested in learning about the utility of selecting $\vp = \{p_1, \ldots, p_K\}$ next.
Since the rewards 
are influenced by $\vf(t)$, 
selecting $\vp$ at time $t+1$ does not necessarily give us the utility samples at $U(\vp)$. 
Instead, the learner can keep pulling this meta arm for a non-negligible $s$ consecutive rounds to ensure that $\vf(t+s)$ approaches $\vp$.
Following this idea, we decompose the total number of time rounds $T$ into $\left \lfloor T/L \right \rfloor$ phases which each phase is associated with $L$ rounds.
We denote $m \in [1, \ldots, \left \lfloor T/L \right \rfloor]$ as the phase index and 
$\vp(m)$ as the selected meta-arm in the $m$-th phase. 
To summarize the above phased-learning template:

\begin{minipage}{0.58\textwidth}
\squishlist 
  \item In each phase $m$, we start with an {\em approaching stage}: the first $s_a$ rounds of the phase. 
  This stage is used to ``move" $\vf(t+s)$ with $1\leq s\leq s_a$ towards to $\vp$. 
  \item  In the second stage, namely, {\em estimation stage}, of each phase: the remaining $L-s_a$ rounds. 
       This stage is used for collecting the realized rewards and estimating the true reward mean on action $\vp$.
  \item Finally, we leverage our tools in \action~to decide what meta arm to select in each phase.
\squishend
\end{minipage}
\hspace{0.1cm}
\begin{minipage}{0.37\textwidth}
\includegraphics[width=1.1\textwidth]{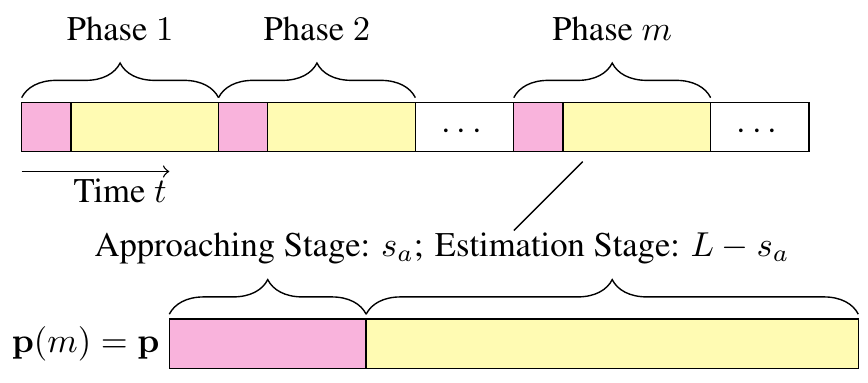}
\captionof{figure}{We deploy $\vp$ for all rounds in $m$-th phases, therefore, we use $\vp(m) = \vp$ to represent $\vp(t) = \vp$ for simplicity.}
\label{hist_algo_illustration}
\end{minipage}

\begin{algorithm}[ht]
\begin{algorithmic}[1]
\caption{Reduction Template} \label{algo:reduction_algo}
 \STATE \textbf{Input:} $K, T$; $\gamma,\epsilon, \rho\in (0,1), s_a$. 
 \STATE \textbf{Input:} A bandit algorithm $\cA$: History-Dependent UCB (Algorithm \ref{algo:history-dependent}). 
  \STATE Split all rounds into consecutive phases of $L = s_a/(1-\rho)$ rounds each.
\FOR{$m = 1, \ldots $}
  \STATE Query algorithm $\cA$ for its meta arm selection $\vp(m) = \vp$.
  \STATE Each phase is separated into two stages:\\ 
      1). Approaching stage: $t = L(m-1)+1, \ldots,L(m-1)+ s_a$;\\ 
     2). Estimation stage: $t= L(m-1)+ s_a+1, \ldots, Lm$.
  \FOR{$t = L(m-1)+1, \ldots, L(m-1)+s_a$ 
  }
    \STATE Deploy the meta arm $\vp$.
  \ENDFOR

  \FOR{$t = L(m-1)+s_a+1, \ldots, Lm$ 
  }
    \STATE Deploy the meta arm $\vp$;
    \STATE Collect the realized rewards $\rewob_t$ of activated arms
            to estimate the mean reward 
            as in~\eqref{eq:update}.
  \ENDFOR

   \STATE Update $\overbar{U}_t^\est\left(\vp\right)$ as in~\eqref{eq:update}.
\ENDFOR
\end{algorithmic}
\end{algorithm}
Note that even if we keep pulling the arm $k$ with the constant probability $p_k$ in the approaching stage, the action impact in the estimation stage is not exactly the same as meta arm we want to learn, i.e., $\vf(t+s) \neq \vp$ for $s\in(s_a, L]$, due to the finite length of the stage.
However, we can guarantee all $\vf(t+s)$ for $s\in(s_a, L]$ is close enough to $\vp$ by bounding its approximation error w.r.t $\vp$.
The above idea enables a more general reduction algorithm that is compatible with any bandit algorithm that solves the action-dependent case. 
Let $\underline{\rho = (L-s_a)/L}$ be the ratio of number of rounds in estimation stage of each phase.
We present this reduction in Algorithm \ref{algo:reduction_algo} and a graphical illustration in Figure~\ref{hist_algo_illustration}.

\subsection{History-Dependent UCB}
\label{sec:history-dependent}
In this section, 
we show how to utilize the reduction template to achieve a $\tilde{\cO}(KT^{2/3})$ regret bound for \history. We first introduce some notations.
For each discretized arm $p_k$, similar to action-dependent case, we define 
$\underline{\Gamma_m(p_k)} := \left\{s: \underline{s\in((i-1)L + s_a, iL] }\text{ where } p_k \in \vp(i), \forall i \in[m] \right\}$ 
as the set of all time indexes till the end of phase $m$ in estimation stages such that arm $k$ is pulled with probability $p_k$.
We define the following empirical 
$\bar{\rewfn}_m^\est(p_k)$ computed from our observations and the empirical utility $\overbar{U}_m^\est(\vp)$:
\footnote{$\est$ 
in superscript stands for $\est$timation stage.}
\begin{align}\label{eq:update}
\bar{\rewfn}_m^\est(p_k) = \frac{1}{n_m^{\est}(p_k)}\sum_{s\in\Gamma_m(p_k)} \hat{\rewfn}_s(p_k^{(\gamma)}(s)),\quad
\overbar{U}_m^\est(\vp)  = \sum_{p_k\in\vp} p_k \cdot \bar{\rewfn}_m^{\est}(p_k),
\end{align}
where $n_m^{\est}(p_k) := |\Gamma_m(p_k)|$ is the total number of rounds pulling arm $k$ with  probability $p_k$ in all estimation stages, and $\hat{\rewfn}_s(p^{(\gamma)}_k(s))$ is defined similarly as in Equation (\ref{eq:reward_impor_weights}).
We use the smoothed-out frequency $\{p_{k}^{(\gamma)}(s)\}_{s\in\Gamma_m(p_k)}$ in the estimation stage as an approximation for the discounted frequency right after the approaching stage.

\begin{minipage}{0.40\textwidth}
We compute our $\UCB$ for each meta arm at the end of each phase.
We define and compute \underline{$\err := K\gamma^{s_a}(L^*+1)$},
the approximation error incurred after our attempt to smooth out the historical action impact.
With these preparations,
we present the phased history-dependent UCB algorithm (in companion with Algorithm~\ref{algo:reduction_algo}) in Algorithm~\ref{algo:history-dependent}.
The main result of this section is given as follows:
\end{minipage}
\hspace{1em}
\begin{minipage}{0.45\textwidth}
\begin{algorithm}[H]
    \caption{History-Dependent UCB} \label{algo:history-dependent}
    \begin{algorithmic}[1]
    \STATE Construct $\UCB$ for each meta arm $\vp \in \cP_\epsilon$ at the end of each phase $m = 1,2,\ldots,$ as follows:
    \begin{align*}
    \UCB_{m}(\vp) = \overbar{U}_m^\est(\vp) + \err + 3\sqrt{\frac{K\ln \left( L\rho\right)}{\min_{p_k\in\vp} n_m^\est(p_k)}}.
    \end{align*}
    \STATE Select $\vp(m+1) = \argmax_{\vp} \UCB_{m}(\vp)$ with ties breaking equally. 
    \end{algorithmic}
\end{algorithm}
\end{minipage}

\begin{theorem} \label{theorem: key_theorem_neq_1}
For any constant ratio $\rho\in(0,1)$ and $\gamma\in(0,1)$, let 
$\epsilon= \Theta((\ln(T\rho)/(T\rho))^{1/3})$ and
$s_a= \Theta(\ln(\epsilon^{1/3}/K)/\ln\gamma)$. 
The regret of Algorithm \ref{algo:reduction_algo} with Algorithm \ref{algo:history-dependent} as input bounds as follows:
$\reg(T) = \cO\big( K T^{2/3}\left((\ln (T\rho))/\rho\right)^{1/3}\big)$.
\end{theorem}
For a constant ratio $\rho$, we match the optimal regret order for \action. 
When $\gamma$ is smaller, the impact function ``forgets" the impact of past-taken actions faster, therefore less rounds in approaching stage would be needed (see $s_a$'s dependence in $\gamma$) and this leads to larger $\rho$.
\begin{remark}
The dependence of our regret on the phase length $L$ is encoded in $\rho$.
When implementing our algorithm (Section~\ref{sec_simu}), we calculate $L$ via $s_a$ given the ratio $\rho$. 
We also run simulations of our algorithm on different ratios $\rho$, the results show that the performance of our algorithm are not sensitive w.r.t. specifying $\rho$s - in practice, we do not require the exact knowledge of $\rho$, instead we can afford to use a rough estimation of its upper bound to compute $L$.
\end{remark}

\subsection{Extension to General Impact Functions}
\label{sec:history-discussion}
So far, we discuss settings when the impact function is specified as in Equation~\eqref{fn:impact_fn}.
However, the same technique we presented earlier can be applied for a more general family of impact functions. 
In particular, as long as the impact function converges after the learner keeps selecting the same action,
our result holds. 
To be more precise, we only require $\vf(t)$ to satisfy the condition 
$|f_k(t+s) - g(p_k)|\le \gamma^s, \gamma\in(0,1)$ when the learner keeps pulling arm $k$ with probability $p_k$ for $s$ round.
The function $g(\cdot)$ can be an arbitrary monotone function as long as it is continuous and differentiable, for example: $g(x) = x$.
In fact, the property of $\vf(t)$ is only used when we estimate how close $\vf$ is to $g(\vp)$ after the approaching stage with repeatedly selecting $\vp$.
For a different $\vf(t)$, we define new reward mean functions $\rewfn_k'(\cdot) = \rewfn_k(g(\cdot))$,
and tune parameters $\epsilon$ and $s_a$ accordingly to bound the approximation error for $\big|U(\vp) - \overbar{U}_m^\est(\vp)\big|$ (change the Lipschitz constant). 
This way we can follow the same algorithmic template to achieve a similar regret.

Moreover, we do not require exact knowledge of the impact function $\vf(t)$. 
We only require the impact functions to satisfy the above conditions for our algorithms/analysis to hold.
With the same arguments, while we assume the reward function $\rewfn_k(\cdot)$ 
is fed with the same impact function $\vf$,
our formulation generalizes to different impact functions for $\rewfn_k(\cdot)$,
as long as these impact functions are able to stabilize given a consecutive adoption of the desired action.

\section{Matching Lower Bounds}\label{sec:lowerbound}

For both action- and history-dependent bandit learning problems,
we have proposed algorithms that achieve a regret bound of $\tilde{\cO}(KT^{2/3})$. 
We now show the above bounds are order-optimal with respect to $K$ and $T$, i.e., the lower bounds of our action- and \history~are both $\Omega(KT^{2/3} )$,
as summarized below.
\begin{theorem} \label{theorem_low_bound_all}
Let $T > 2K$ and $K \geq 4$, 
there exist problem instances that for our action- and history-dependent bandits, respectively, 
the regret for any algorithm $\cA$ follows:
$\inf_{\cA} \reg(T) \geq \Omega(K T^{2/3})$.
\end{theorem}
For the lower bound proof of action-dependent bandits (included in Appendix~\ref{sec:lower_bound}), 
we following the standard randomized problem instances construction
used in combinatorial bandits and Lipschitz bandits and use information inequality to prove the lower bound.
For history-dependent bandits, we show that for a general class of reward function $r_k(\cdot)$ which satisfies the \emph{proper} property (see Definition~\ref{defn_strict_proper_utility}), solving history-dependent bandits is as least as hard as solving action-dependent bandits.
Armed with the above derived lower bound of action-dependent bandits, we can then conclude the lower bound of history-dependent bandits.

\section{Numeric Experiments} \label{sec_simu}

\begin{figure}
    \centering
    \subfigure[Action-dependent]{\label{fig_1}\includegraphics[width=0.245\textwidth]{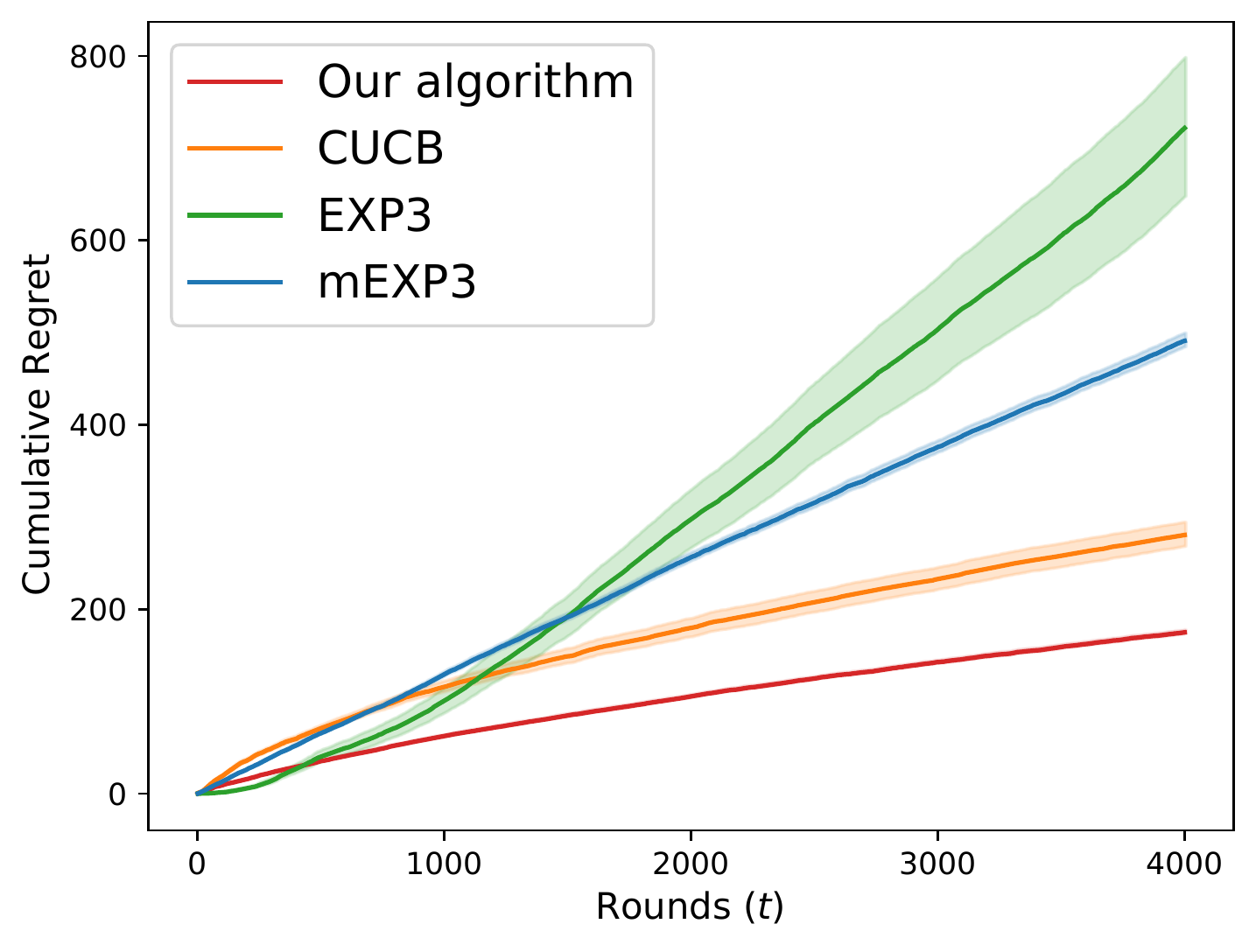}} 
    \subfigure[$\gamma = 0.2$]{\label{fig:gamma_0.2}\includegraphics[width=0.245\textwidth]{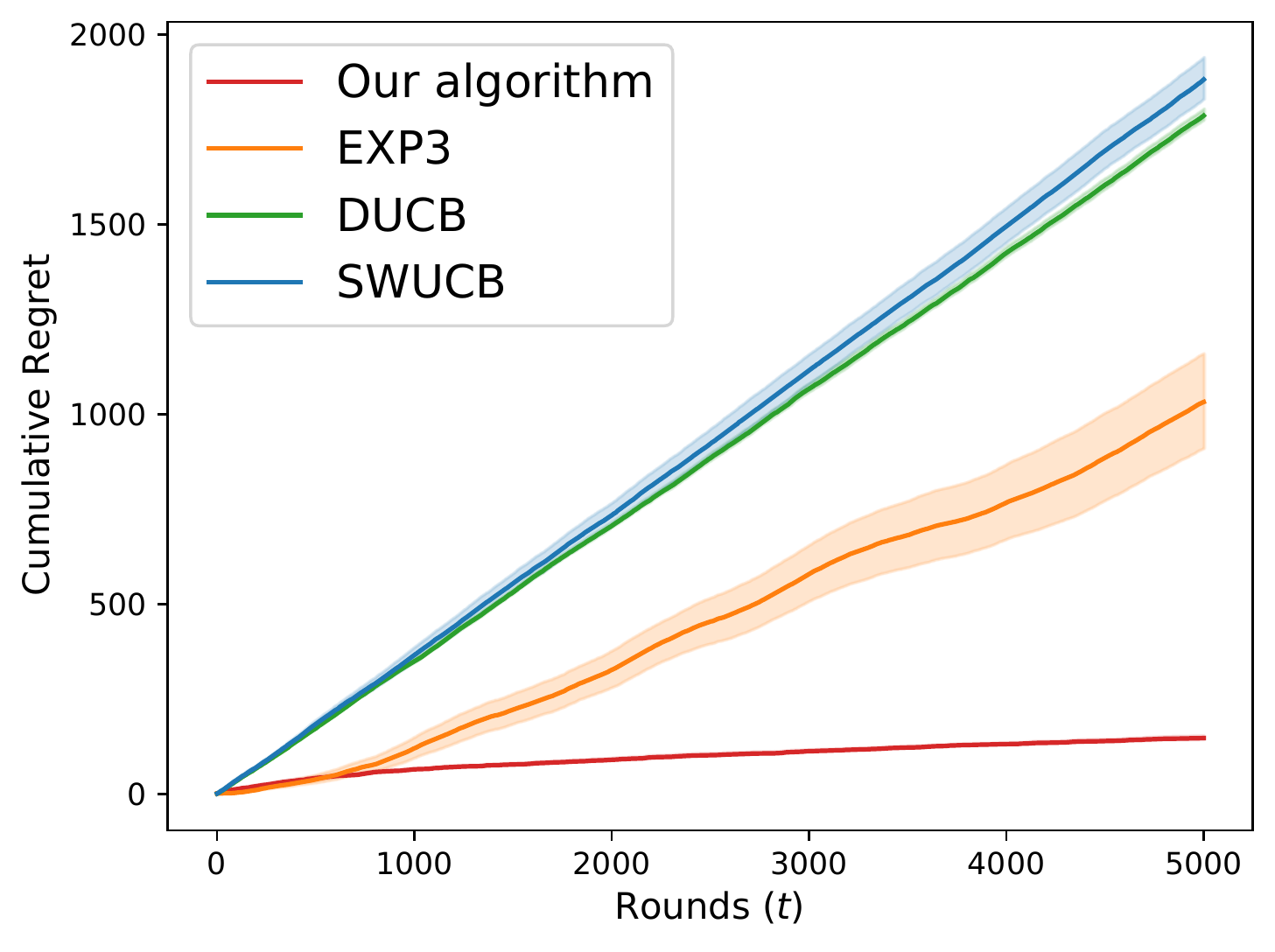}}
    \subfigure[$\gamma = 0.4$]{\label{fig:gamma_0.4}\includegraphics[width=0.245\textwidth]{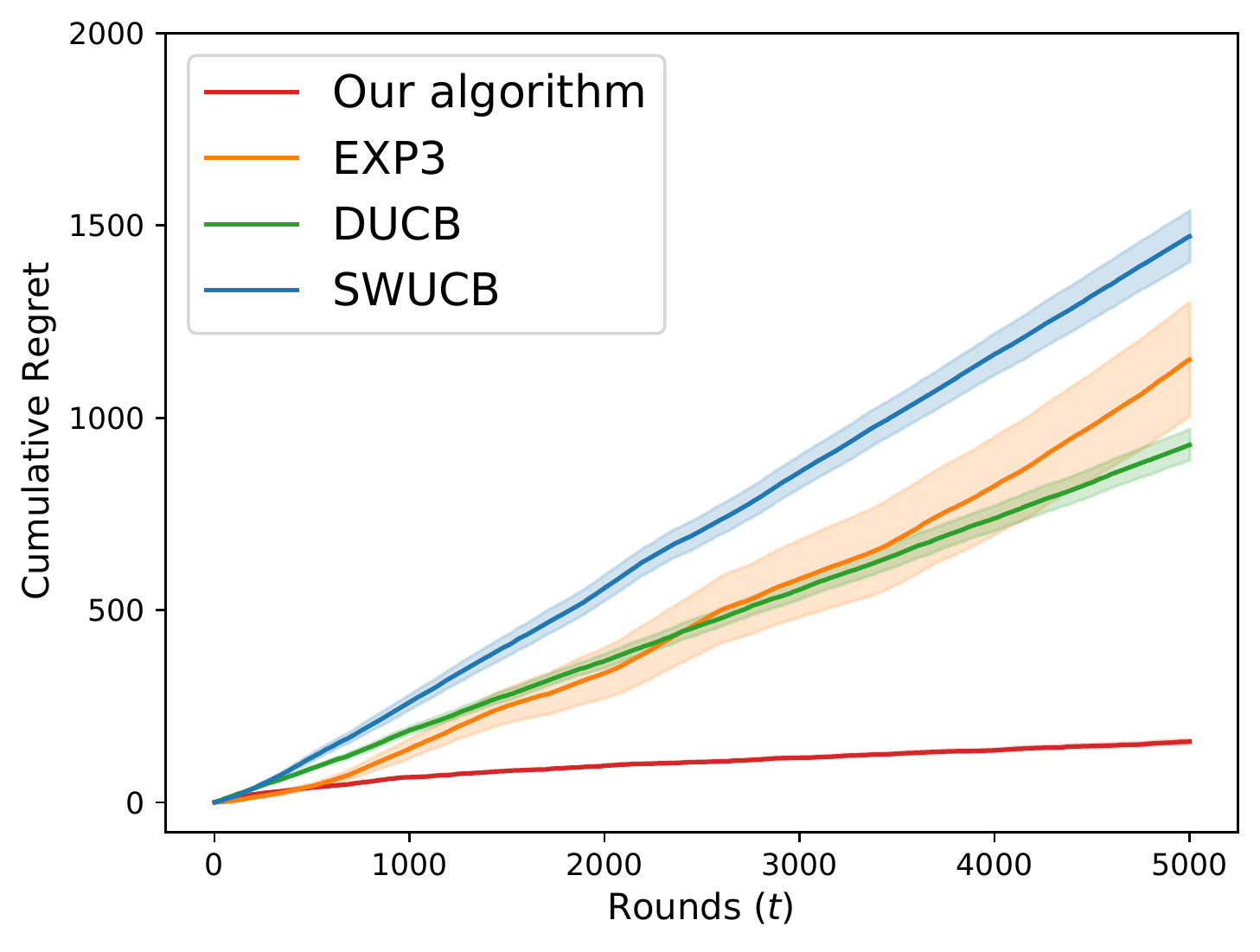}}
    \subfigure[$\gamma = 0.6$]{\label{fig:gamma_0.6}\includegraphics[width=0.245\textwidth]{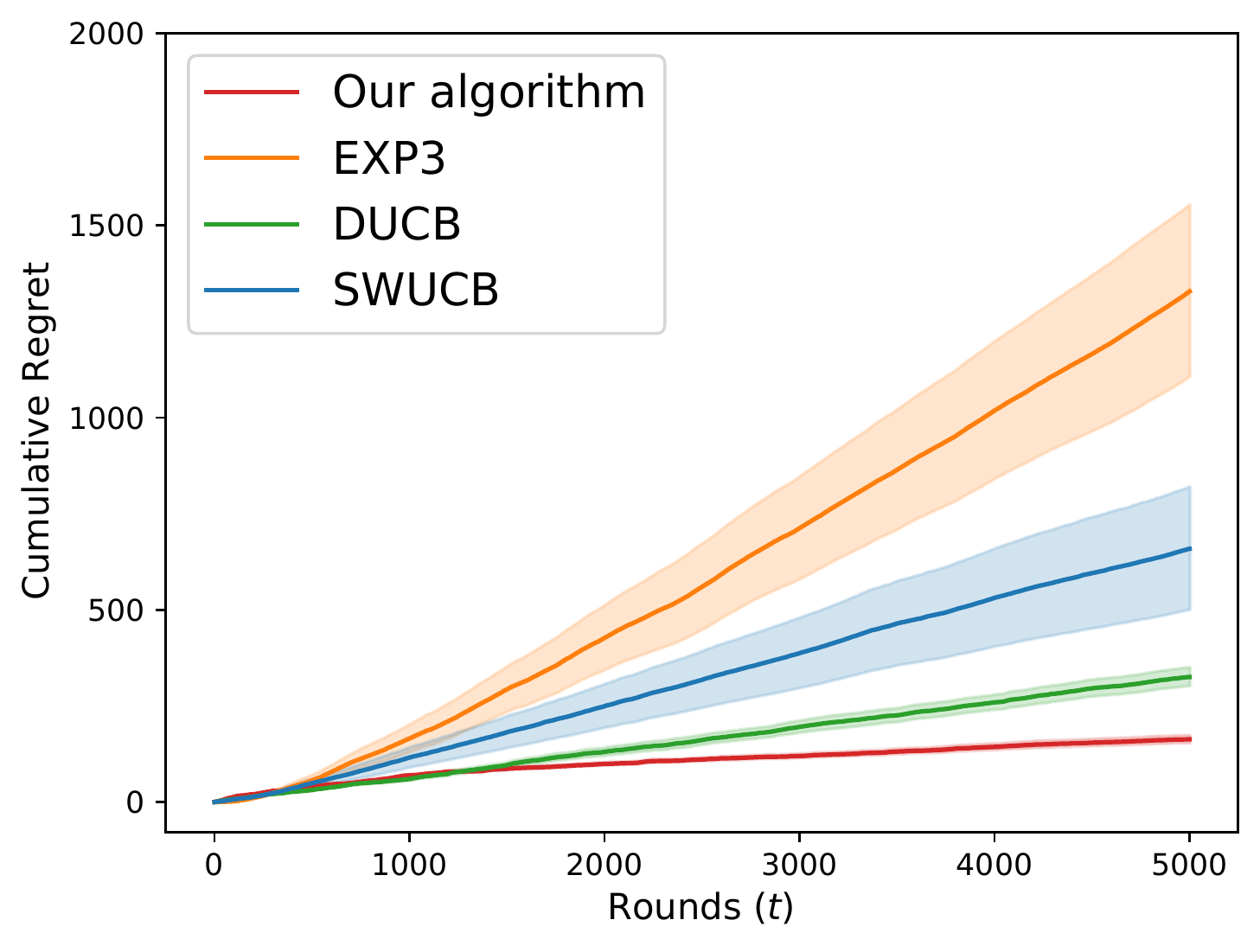}}
    \caption{(a): \textmd{Behavior of the different algorithms for \action.} (b)-(d): \textmd{Behavior of the different algorithms for \history~on different $\gamma$.}}
    \label{fig:main}
\end{figure}
\begin{figure}
    \setlength\belowcaptionskip{-3ex}
    \centering
    \subfigure[$K = 3$]{\label{fig:K_3}\includegraphics[width=0.245\textwidth]{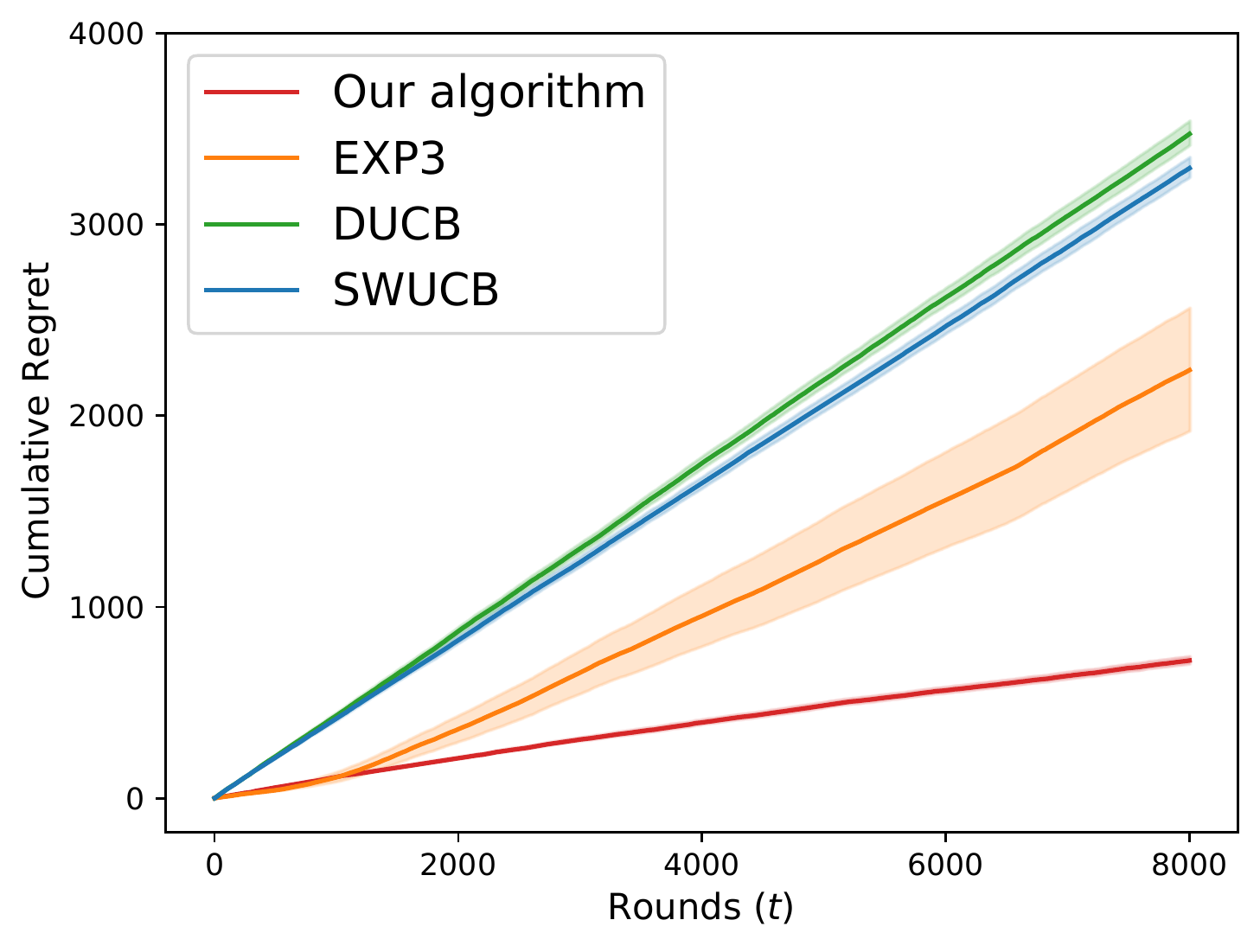}} 
    \subfigure[$K = 4$]{\label{fig:K_4}\includegraphics[width=0.245\textwidth]{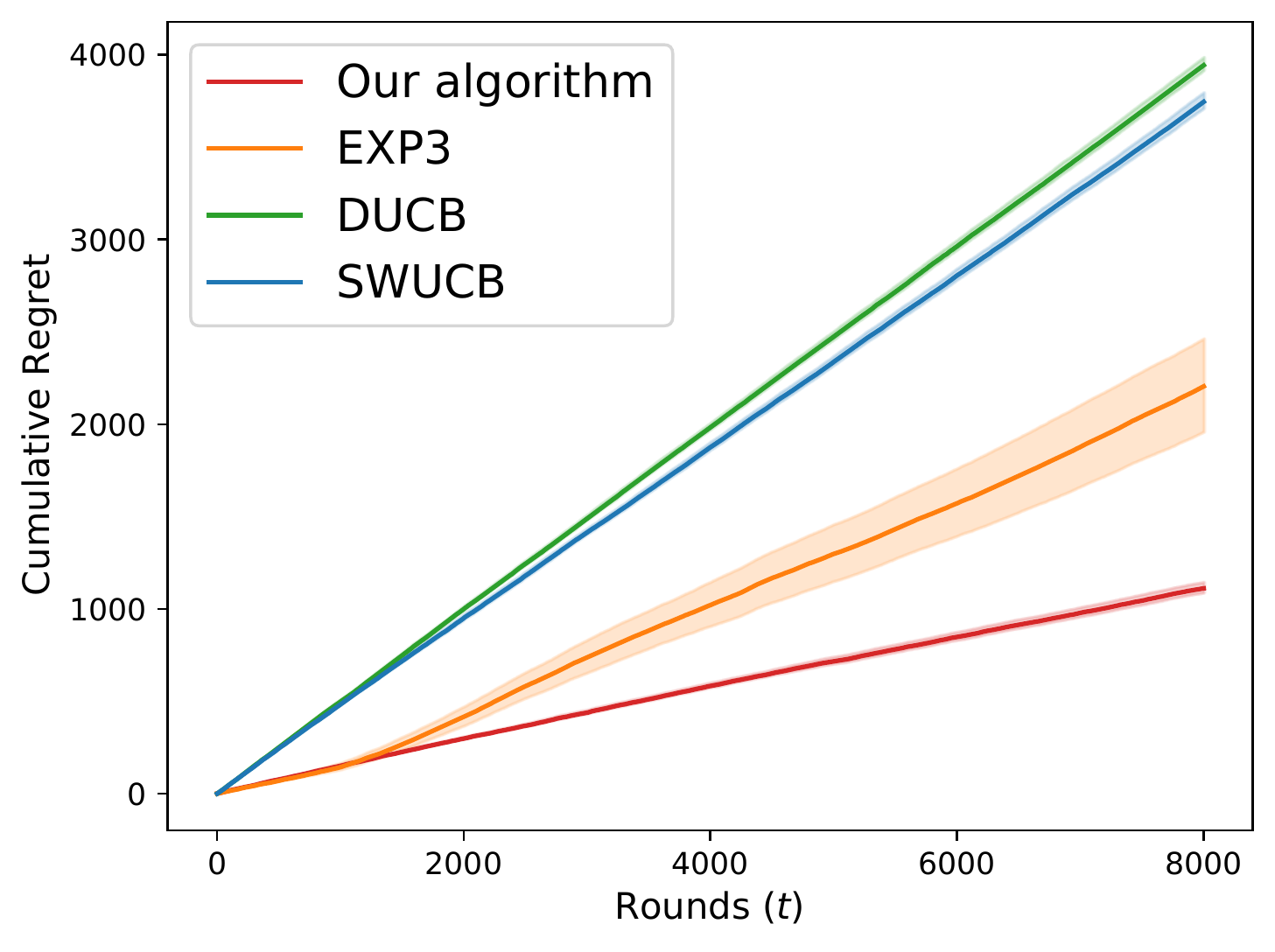}}
    \subfigure[$K = 5$]{\label{fig:K_5}\includegraphics[width=0.245\textwidth]{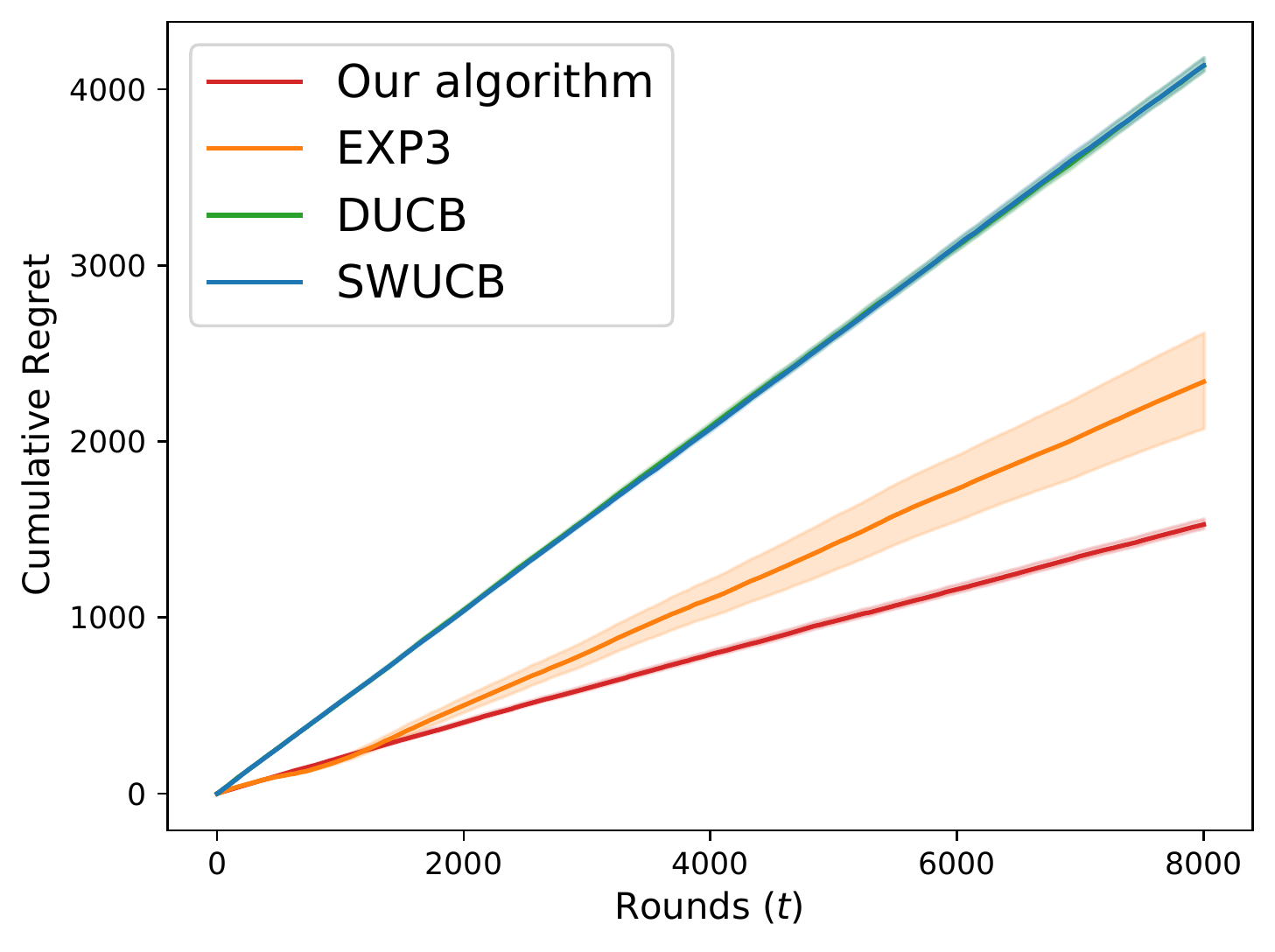}}
    \subfigure[different ratio]{\label{fig:ratio}\includegraphics[width=0.245\textwidth]{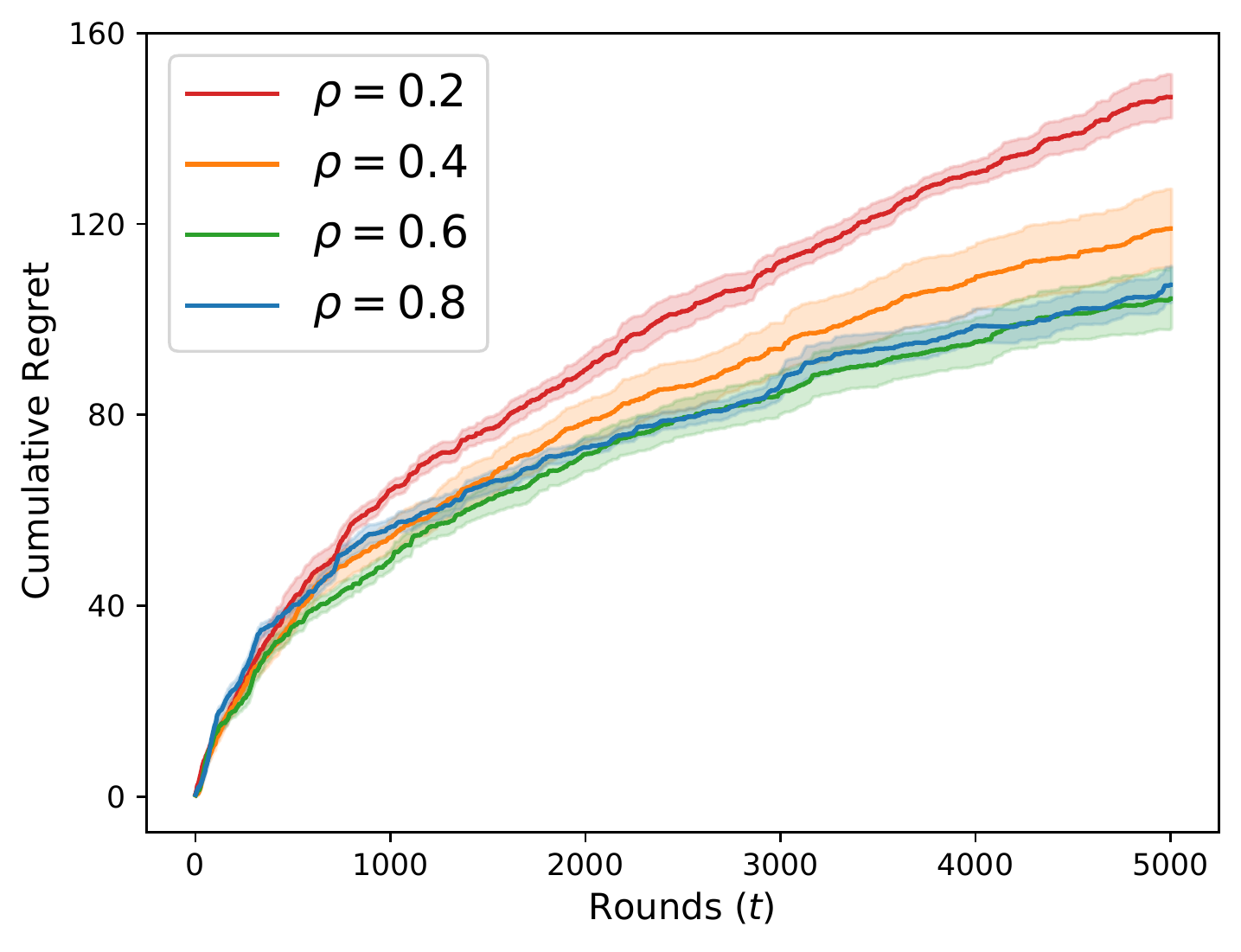}}
    \caption{(a)-(c): \textmd{Behavior of the different algorithms on different $K$, the remaining parameters are the same with the simulations in comparison on different $K$.} (d): \textmd{The performance of our algorithms on different ratios, we set $K=2$ and remaining parameters are also same as before.}}
    \label{fig:K_and_ratio}
\end{figure}

We conducted a series of simulations to understand the performance of our algorithms. 
The detailed setups and discussion are in Appendix~\ref{missing_exps}. 
We first compare our algorithm with some baselines under \action~and other non-stationary baselines under \history~with different $\gamma$ (the parameter in time-discounted frequency). 
The results, as shown in Fig.~\ref{fig:main}, demonstrate that our proposed algorithm 
consistently outperforms the baseline methods. 
We also note that the performance of our algorithm is relatively robust w.r.t. 
difference choices of $\gamma$, i.e., the time-discounting factor for the impact.
One explanation is that our algorithm utilizes repeated pulling to smooth out historical bias. 
Given the exponential-decaying nature of time discounting, 
the amount of pulling required for the impact to converge 
does not depend on $\gamma$ too heavily. 
As shown in our theoretical regret upper bounds, gamma can be absorbed with other numeric constants, 
and when the time horizon increases, 
the effect of gamma on our algorithm's performance is diminishing, 
which aligns with our empirical observation. 
We also examine our algorithm with larger number of base arms $K$ and different ratios $\rho$. 
The results, as in Figure~\ref{fig:K_and_ratio}, 
show that our algorithm outperforms other baselines when $K$ goes large.
Furthermore, in our regret bounds (see Theorem~\ref{theorem: key_theorem_neq_1}), 
the regret scales linearly w.r.t $K$. 
Though the presented results absorb other numeric constants, 
it is expected to see that the slope of the regret curve is proportionally 
increasing along with increasing $K$.
The results also suggest that our algorithm is not sensitive to different $\rho$, though one could see the regret is slightly lower when $\rho$ is increasing, which is expected from our regret bound.

\section{Conclusion and Future Work}

\label{sec:conclusion}
We explore a multi-armed bandit problem in which actions have delayed impacts to the arm rewards.
We propose algorithms that achieve a regret of $\tilde{\cO}(KT^{2/3})$ and provide a matching lower regret bound of $\Omega(KT^{2/3})$.
Our results complement the bandit literature by exploring the action history dependent biases in bandits. 
While our model have its limitations, 
it captures an important but relatively under-explored angle in algorithmic fairness, 
the long-term impact of actions in sequential learning settings.
We hope our study will open more discussions along this direction.

\begin{ack}
We thank the anonymous reviewers for their valuable comments.
This work is supported in part by the Office of Naval Research Grant N00014-20-1-2240 and the National Science Foundation (NSF) FAI program in collaboration with Amazon under grant IIS-1939677 and IIS-2040800. 
\end{ack}

\bibliography{0_mybib.bib}

\begin{thebibliography}{67}
\providecommand{\natexlab}[1]{#1}
\providecommand{\url}[1]{\texttt{#1}}
\expandafter\ifx\csname urlstyle\endcsname\relax
  \providecommand{\doi}[1]{doi: #1}\else
  \providecommand{\doi}{doi: \begingroup \urlstyle{rm}\Url}\fi

\bibitem[Angwin et~al.(2016)Angwin, Larson, Mattu, and
  Kirchner]{angwin2016machine}
Angwin, J., Larson, J., Mattu, S., and Kirchner, L.
\newblock Machine bias.
\newblock \emph{ProPublica, May}, 23:\penalty0 2016, 2016.

\bibitem[Audibert \& Bubeck(2010)Audibert and Bubeck]{audibert2010regret}
Audibert, J.-Y. and Bubeck, S.
\newblock Regret bounds and minimax policies under partial monitoring.
\newblock \emph{Journal of Machine Learning Research}, 11:\penalty0 2785--2836,
  2010.

\bibitem[Auer et~al.(2002{\natexlab{a}})Auer, Cesa-Bianchi, and
  Fischer]{auer2002finite}
Auer, P., Cesa-Bianchi, N., and Fischer, P.
\newblock Finite-time analysis of the multiarmed bandit problem.
\newblock \emph{Machine learning}, 47\penalty0 (2-3):\penalty0 235--256,
  2002{\natexlab{a}}.

\bibitem[Auer et~al.(2002{\natexlab{b}})Auer, Cesa-Bianchi, Freund, and
  Schapire]{auer2002nonstochastic}
Auer, P., Cesa-Bianchi, N., Freund, Y., and Schapire, R.~E.
\newblock The nonstochastic multiarmed bandit problem.
\newblock \emph{SIAM journal on computing}, 32\penalty0 (1):\penalty0 48--77,
  2002{\natexlab{b}}.

\bibitem[Bartik \& Nelson(2016)Bartik and Nelson]{bartik2016credit}
Bartik, A. and Nelson, S.
\newblock Credit reports as resumes: The incidence of pre-employment credit
  screening.
\newblock 2016.

\bibitem[Bartlett et~al.(2018)Bartlett, Morse, Stanton, and
  Wallace]{bartlett2018consumer}
Bartlett, R., Morse, A., Stanton, R., and Wallace, N.
\newblock Consumer-lending discrimination in the era of fintech.
\newblock \emph{Unpublished working paper. University of California, Berkeley},
  2018.

\bibitem[Bechavod et~al.(2019)Bechavod, Ligett, Roth, Waggoner, and
  Wu]{bechavod2019equal}
Bechavod, Y., Ligett, K., Roth, A., Waggoner, B., and Wu, S.~Z.
\newblock Equal opportunity in online classification with partial feedback.
\newblock In \emph{Advances in Neural Information Processing Systems}, pp.\
  8972--8982, 2019.

\bibitem[Besbes et~al.(2014)Besbes, Gur, and Zeevi]{besbes2014stochastic}
Besbes, O., Gur, Y., and Zeevi, A.
\newblock Stochastic multi-armed-bandit problem with non-stationary rewards.
\newblock In \emph{Advances in neural information processing systems}, pp.\
  199--207, 2014.

\bibitem[Besbes et~al.(2015)Besbes, Gur, and Zeevi]{besbes2015non}
Besbes, O., Gur, Y., and Zeevi, A.
\newblock Non-stationary stochastic optimization.
\newblock \emph{Operations research}, 63\penalty0 (5):\penalty0 1227--1244,
  2015.

\bibitem[Cella \& Cesa-Bianchi(2020)Cella and
  Cesa-Bianchi]{cella2020stochastic}
Cella, L. and Cesa-Bianchi, N.
\newblock Stochastic bandits with delay-dependent payoffs.
\newblock In \emph{International Conference on Artificial Intelligence and
  Statistics}, pp.\  1168--1177. PMLR, 2020.

\bibitem[Cesa-Bianchi \& Lugosi(2012)Cesa-Bianchi and
  Lugosi]{cesa2012combinatorial}
Cesa-Bianchi, N. and Lugosi, G.
\newblock Combinatorial bandits.
\newblock \emph{Journal of Computer and System Sciences}, 78\penalty0
  (5):\penalty0 1404--1422, 2012.

\bibitem[Chen et~al.(2016{\natexlab{a}})Chen, Hu, Li, Li, Liu, and
  Lu]{chen2016combinatorial2}
Chen, W., Hu, W., Li, F., Li, J., Liu, Y., and Lu, P.
\newblock Combinatorial multi-armed bandit with general reward functions.
\newblock In \emph{Advances in Neural Information Processing Systems}, pp.\
  1659--1667, 2016{\natexlab{a}}.

\bibitem[Chen et~al.(2016{\natexlab{b}})Chen, Wang, Yuan, and
  Wang]{chen2016combinatorial}
Chen, W., Wang, Y., Yuan, Y., and Wang, Q.
\newblock Combinatorial multi-armed bandit and its extension to
  probabilistically triggered arms.
\newblock \emph{The Journal of Machine Learning Research}, 17\penalty0
  (1):\penalty0 1746--1778, 2016{\natexlab{b}}.

\bibitem[Chouldechova(2017)]{chouldechova2017fair}
Chouldechova, A.
\newblock Fair prediction with disparate impact: A study of bias in recidivism
  prediction instruments.
\newblock \emph{Big data}, 5\penalty0 (2):\penalty0 153--163, 2017.

\bibitem[Combes et~al.(2015)Combes, Shahi, Proutiere,
  et~al.]{combes2015combinatorial}
Combes, R., Shahi, M. S. T.~M., Proutiere, A., et~al.
\newblock Combinatorial bandits revisited.
\newblock In \emph{Advances in Neural Information Processing Systems}, pp.\
  2116--2124, 2015.

\bibitem[Combes et~al.(2020)Combes, Prouti{\`e}re, and
  Fauquette]{combes2020unimodal}
Combes, R., Prouti{\`e}re, A., and Fauquette, A.
\newblock Unimodal bandits with continuous arms: Order-optimal regret without
  smoothness.
\newblock \emph{Proceedings of the ACM on Measurement and Analysis of Computing
  Systems}, 4\penalty0 (1):\penalty0 1--28, 2020.

\bibitem[Cortes et~al.(2017)Cortes, DeSalvo, Kuznetsov, Mohri, and
  Yang]{cortes2017discrepancy}
Cortes, C., DeSalvo, G., Kuznetsov, V., Mohri, M., and Yang, S.
\newblock Discrepancy-based algorithms for non-stationary rested bandits.
\newblock \emph{arXiv preprint arXiv:1710.10657}, 2017.

\bibitem[Cowgill \& Tucker(2019)Cowgill and Tucker]{cowgill2019economics}
Cowgill, B. and Tucker, C.~E.
\newblock Economics, fairness and algorithmic bias.
\newblock \emph{preparation for: Journal of Economic Perspectives}, 2019.

\bibitem[Cowgill \& Zitzewitz(2009)Cowgill and Zitzewitz]{cowgill2009incentive}
Cowgill, B. and Zitzewitz, E.
\newblock Incentive effects of equity compensation: Employee level evidence
  from google.
\newblock \emph{Dartmouth Department of Economics working paper}, 2009.

\bibitem[D'Amour et~al.(2020)D'Amour, Srinivasan, Atwood, Baljekar, Sculley,
  and Halpern]{d2020fairness}
D'Amour, A., Srinivasan, H., Atwood, J., Baljekar, P., Sculley, D., and
  Halpern, Y.
\newblock Fairness is not static: deeper understanding of long term fairness
  via simulation studies.
\newblock In \emph{Proceedings of the 2020 Conference on Fairness,
  Accountability, and Transparency}, pp.\  525--534, 2020.

\bibitem[Duran \& Verloop(2018)Duran and Verloop]{duran2018asymptotic}
Duran, S. and Verloop, I.~M.
\newblock Asymptotic optimal control of markov-modulated restless bandits.
\newblock \emph{Proceedings of the ACM on Measurement and Analysis of Computing
  Systems}, 2\penalty0 (1):\penalty0 1--25, 2018.

\bibitem[Elzayn et~al.(2019)Elzayn, Jabbari, Jung, Kearns, Neel, Roth, and
  Schutzman]{elzayn2019fair}
Elzayn, H., Jabbari, S., Jung, C., Kearns, M., Neel, S., Roth, A., and
  Schutzman, Z.
\newblock Fair algorithms for learning in allocation problems.
\newblock In \emph{Proceedings of the Conference on Fairness, Accountability,
  and Transparency}, pp.\  170--179, 2019.

\bibitem[Fuster et~al.(2018)Fuster, Goldsmith-Pinkham, Ramadorai, and
  Walther]{fuster2018predictably}
Fuster, A., Goldsmith-Pinkham, P., Ramadorai, T., and Walther, A.
\newblock Predictably unequal? the effects of machine learning on credit
  markets.
\newblock 2018.

\bibitem[Gael et~al.(2020)Gael, Vernade, Carpentier, and
  Valko]{gael2020stochastic}
Gael, M.~A., Vernade, C., Carpentier, A., and Valko, M.
\newblock Stochastic bandits with arm-dependent delays.
\newblock In \emph{International Conference on Machine Learning}, pp.\
  3348--3356. PMLR, 2020.

\bibitem[Garivier \& Moulines(2011)Garivier and Moulines]{garivier2011upper}
Garivier, A. and Moulines, E.
\newblock On upper-confidence bound policies for switching bandit problems.
\newblock In \emph{International Conference on Algorithmic Learning Theory},
  pp.\  174--188, 2011.

\bibitem[Gelman et~al.(2007)Gelman, Fagan, and Kiss]{gelman2007analysis}
Gelman, A., Fagan, J., and Kiss, A.
\newblock An analysis of the new york city police department's
  “stop-and-frisk” policy in the context of claims of racial bias.
\newblock \emph{Journal of the American statistical association}, 102\penalty0
  (479):\penalty0 813--823, 2007.

\bibitem[Gillen et~al.(2018)Gillen, Jung, Kearns, and Roth]{gillen2018online}
Gillen, S., Jung, C., Kearns, M., and Roth, A.
\newblock Online learning with an unknown fairness metric.
\newblock In \emph{Advances in Neural Information Processing Systems}, pp.\
  2600--2609, 2018.

\bibitem[Goel et~al.(2016)Goel, Rao, Shroff, et~al.]{goel2016precinct}
Goel, S., Rao, J.~M., Shroff, R., et~al.
\newblock Precinct or prejudice? understanding racial disparities in new york
  city’s stop-and-frisk policy.
\newblock \emph{The Annals of Applied Statistics}, 10\penalty0 (1):\penalty0
  365--394, 2016.

\bibitem[Gretton et~al.(2009)Gretton, Smola, Huang, Schmittfull, Borgwardt, and
  Sch{\"o}lkopf]{gretton2009covariate}
Gretton, A., Smola, A., Huang, J., Schmittfull, M., Borgwardt, K., and
  Sch{\"o}lkopf, B.
\newblock Covariate shift by kernel mean matching.
\newblock \emph{Dataset shift in machine learning}, 3\penalty0 (4):\penalty0 5,
  2009.

\bibitem[Gupta \& Kamble(2019)Gupta and Kamble]{gupta2019individual}
Gupta, S. and Kamble, V.
\newblock Individual fairness in hindsight.
\newblock In \emph{Proceedings of the 2019 ACM Conference on Economics and
  Computation}, pp.\  805--806, 2019.

\bibitem[Heidari et~al.(2019)Heidari, Nanda, and Gummadi]{heidari2019long}
Heidari, H., Nanda, V., and Gummadi, K.~P.
\newblock On the long-term impact of algorithmic decision policies: Effort
  unfairness and feature segregation through social learning.
\newblock \emph{arXiv preprint arXiv:1903.01209}, 2019.

\bibitem[Ho et~al.(2016)Ho, Slivkins, and Vaughan]{ho2016adaptive}
Ho, C.-J., Slivkins, A., and Vaughan, J.~W.
\newblock Adaptive contract design for crowdsourcing markets: Bandit algorithms
  for repeated principal-agent problems.
\newblock \emph{Journal of Artificial Intelligence Research}, 55:\penalty0
  317--359, 2016.

\bibitem[Hu \& Chen(2018)Hu and Chen]{hu2018short}
Hu, L. and Chen, Y.
\newblock A short-term intervention for long-term fairness in the labor market.
\newblock In \emph{Proceedings of the 2018 World Wide Web Conference}, pp.\
  1389--1398, 2018.

\bibitem[Joseph et~al.(2016)Joseph, Kearns, Morgenstern, and
  Roth]{joseph2016fairness}
Joseph, M., Kearns, M., Morgenstern, J.~H., and Roth, A.
\newblock Fairness in learning: Classic and contextual bandits.
\newblock In \emph{Advances in Neural Information Processing Systems}, pp.\
  325--333, 2016.

\bibitem[Joulani et~al.(2013)Joulani, Gyorgy, and
  Szepesv{\'a}ri]{joulani2013online}
Joulani, P., Gyorgy, A., and Szepesv{\'a}ri, C.
\newblock Online learning under delayed feedback.
\newblock In \emph{International Conference on Machine Learning}, pp.\
  1453--1461, 2013.

\bibitem[Kannan et~al.(2019)Kannan, Roth, and Ziani]{kannan2019downstream}
Kannan, S., Roth, A., and Ziani, J.
\newblock Downstream effects of affirmative action.
\newblock In \emph{Proceedings of the Conference on Fairness, Accountability,
  and Transparency}, pp.\  240--248, 2019.

\bibitem[Kleinberg et~al.(2018)Kleinberg, Ludwig, Mullainathan, and
  Sunstein]{kleinberg2018discrimination}
Kleinberg, J., Ludwig, J., Mullainathan, S., and Sunstein, C.~R.
\newblock Discrimination in the age of algorithms.
\newblock \emph{Journal of Legal Analysis}, 10, 2018.

\bibitem[Kleinberg \& Immorlica(2018)Kleinberg and
  Immorlica]{kleinberg2018recharging}
Kleinberg, R. and Immorlica, N.
\newblock Recharging bandits.
\newblock In \emph{2018 IEEE 59th Annual Symposium on Foundations of Computer
  Science}, pp.\  309--319, 2018.

\bibitem[Kleinberg et~al.(2008)Kleinberg, Slivkins, and
  Upfal]{kleinberg2008multi}
Kleinberg, R., Slivkins, A., and Upfal, E.
\newblock Multi-armed bandits in metric spaces.
\newblock In \emph{Proceedings of the fortieth annual ACM symposium on Theory
  of computing}, pp.\  681--690, 2008.

\bibitem[Kocsis \& Szepesv{\'a}ri(2006)Kocsis and
  Szepesv{\'a}ri]{kocsis2006discounted}
Kocsis, L. and Szepesv{\'a}ri, C.
\newblock Discounted ucb.
\newblock In \emph{2nd PASCAL Challenges Workshop}, volume~2, 2006.

\bibitem[Kolobov et~al.(2020)Kolobov, Bubeck, and Zimmert]{kolobov2020online}
Kolobov, A., Bubeck, S., and Zimmert, J.
\newblock Online learning for active cache synchronization.
\newblock In \emph{International Conference on Machine Learning}, pp.\
  5371--5380. PMLR, 2020.

\bibitem[Lai \& Robbins(1985)Lai and Robbins]{lai1985asymptotically}
Lai, T.~L. and Robbins, H.
\newblock Asymptotically efficient adaptive allocation rules.
\newblock \emph{Advances in applied mathematics}, 6\penalty0 (1):\penalty0
  4--22, 1985.

\bibitem[Levine et~al.(2017)Levine, Crammer, and Mannor]{levine2017rotting}
Levine, N., Crammer, K., and Mannor, S.
\newblock Rotting bandits.
\newblock In \emph{Advances in neural information processing systems}, pp.\
  3074--3083, 2017.

\bibitem[Li et~al.(2019)Li, Liu, and Ji]{li2019combinatorial}
Li, F., Liu, J., and Ji, B.
\newblock Combinatorial sleeping bandits with fairness constraints.
\newblock \emph{IEEE Transactions on Network Science and Engineering}, 2019.

\bibitem[Liu et~al.(2018)Liu, Dean, Rolf, Simchowitz, and
  Hardt]{liu2018delayed}
Liu, L.~T., Dean, S., Rolf, E., Simchowitz, M., and Hardt, M.
\newblock Delayed impact of fair machine learning.
\newblock In \emph{International Conference on Machine Learning}, pp.\
  3150--3158, 2018.

\bibitem[Liu et~al.(2020)Liu, Wilson, Haghtalab, Kalai, Borgs, and
  Chayes]{liu2020disparate}
Liu, L.~T., Wilson, A., Haghtalab, N., Kalai, A.~T., Borgs, C., and Chayes, J.
\newblock The disparate equilibria of algorithmic decision making when
  individuals invest rationally.
\newblock In \emph{Proceedings of the 2020 Conference on Fairness,
  Accountability, and Transparency}, pp.\  381--391, 2020.

\bibitem[Liu(2017)]{liufair}
Liu, Y.
\newblock Fair optimal stopping policy for matching with mediator.
\newblock In \emph{Uncertainty in Artificial Intelligence}, 2017.

\bibitem[Liu \& Ho(2018)Liu and Ho]{liu2018incentivizing}
Liu, Y. and Ho, C.-J.
\newblock Incentivizing high quality user contributions: New arm generation in
  bandit learning.
\newblock In \emph{Thirty-Second AAAI Conference on Artificial Intelligence},
  2018.

\bibitem[Liu et~al.(2017)Liu, Radanovic, Dimitrakakis, Mandal, and
  Parkes]{liu2017calibrated}
Liu, Y., Radanovic, G., Dimitrakakis, C., Mandal, D., and Parkes, D.~C.
\newblock Calibrated fairness in bandits.
\newblock \emph{Proceedings of the 4th Workshop on Fairness, Accountability,
  and Transparency in Machine Learning}, 2017.

\bibitem[Magureanu et~al.(2014)Magureanu, Combes, and
  Proutiere]{magureanu2014lipschitz}
Magureanu, S., Combes, R., and Proutiere, A.
\newblock Lipschitz bandits: Regret lower bound and optimal algorithms.
\newblock In \emph{Conference on Learning Theory}, pp.\  975--999, 2014.

\bibitem[Mouzannar et~al.(2019)Mouzannar, Ohannessian, and
  Srebro]{mouzannar2019fair}
Mouzannar, H., Ohannessian, M.~I., and Srebro, N.
\newblock From fair decision making to social equality.
\newblock In \emph{Proceedings of the Conference on Fairness, Accountability,
  and Transparency}, pp.\  359--368, 2019.

\bibitem[Obermeyer et~al.(2019)Obermeyer, Powers, Vogeli, and
  Mullainathan]{obermeyer2019dissecting}
Obermeyer, Z., Powers, B., Vogeli, C., and Mullainathan, S.
\newblock Dissecting racial bias in an algorithm used to manage the health of
  populations.
\newblock \emph{Science}, 366\penalty0 (6464):\penalty0 447--453, 2019.

\bibitem[Ortner et~al.(2012)Ortner, Ryabko, Auer, and Munos]{ortner2012regret}
Ortner, R., Ryabko, D., Auer, P., and Munos, R.
\newblock Regret bounds for restless markov bandits.
\newblock In \emph{International Conference on Algorithmic Learning Theory},
  pp.\  214--228, 2012.

\bibitem[Patil et~al.(2020)Patil, Ghalme, Nair, and
  Narahari]{patil2020achieving}
Patil, V., Ghalme, G., Nair, V., and Narahari, Y.
\newblock Achieving fairness in the stochastic multi-armed bandit problem.
\newblock In \emph{Proceedings of the AAAI Conference on Artificial
  Intelligence}, volume~34, pp.\  5379--5386, 2020.

\bibitem[Pike-Burke \& Gr{\"u}new{\"a}lder(2019)Pike-Burke and
  Gr{\"u}new{\"a}lder]{pike2019recovering}
Pike-Burke, C. and Gr{\"u}new{\"a}lder, S.
\newblock Recovering bandits.
\newblock \emph{arXiv preprint arXiv:1910.14354}, 2019.

\bibitem[Pike-Burke et~al.(2018)Pike-Burke, Agrawal, Szepesvari, and
  Grunewalder]{pike2018bandits}
Pike-Burke, C., Agrawal, S., Szepesvari, C., and Grunewalder, S.
\newblock Bandits with delayed, aggregated anonymous feedback.
\newblock In \emph{International Conference on Machine Learning}, pp.\
  4105--4113, 2018.

\bibitem[Schmit \& Riquelme(2018)Schmit and Riquelme]{schmit2018human}
Schmit, S. and Riquelme, C.
\newblock Human interaction with recommendation systems.
\newblock In \emph{International Conference on Artificial Intelligence and
  Statistics}, pp.\  862--870, 2018.

\bibitem[Seznec et~al.(2019)Seznec, Locatelli, Carpentier, Lazaric, and
  Valko]{seznec2019rotting}
Seznec, J., Locatelli, A., Carpentier, A., Lazaric, A., and Valko, M.
\newblock Rotting bandits are no harder than stochastic ones.
\newblock In \emph{The 22nd International Conference on Artificial Intelligence
  and Statistics}, pp.\  2564--2572, 2019.

\bibitem[Slivkins(2014)]{slivkins2014contextual}
Slivkins, A.
\newblock Contextual bandits with similarity information.
\newblock \emph{The Journal of Machine Learning Research}, 15\penalty0
  (1):\penalty0 2533--2568, 2014.

\bibitem[Slivkins \& Upfal(2008)Slivkins and Upfal]{slivkins2008adapting}
Slivkins, A. and Upfal, E.
\newblock Adapting to a changing environment: the brownian restless bandits.
\newblock In \emph{Conference on Learning Theory}, pp.\  343--354, 2008.

\bibitem[Tang \& Ho(2019)Tang and Ho]{tang2019bandit}
Tang, W. and Ho, C.-J.
\newblock Bandit learning with biased human feedback.
\newblock In \emph{Eighteenth International Conference on Autonomous Agents and
  Multi-Agent Systems}, 2019.

\bibitem[Tekin \& Liu(2010)Tekin and Liu]{tekin2010online}
Tekin, C. and Liu, M.
\newblock Online algorithms for the multi-armed bandit problem with markovian
  rewards.
\newblock In \emph{Proceedings of the 48th Annual Allerton Conference on
  Communication, Control, and Computing (Allerton)}, 2010.

\bibitem[Verloop et~al.(2016)]{verloop2016asymptotically}
Verloop, I.~M. et~al.
\newblock Asymptotically optimal priority policies for indexable and
  nonindexable restless bandits.
\newblock \emph{The Annals of Applied Probability}, 26\penalty0 (4):\penalty0
  1947--1995, 2016.

\bibitem[Vernade et~al.(2017)Vernade, Capp{\'e}, and
  Perchet]{vernade2017stochastic}
Vernade, C., Capp{\'e}, O., and Perchet, V.
\newblock Stochastic bandit models for delayed conversions.
\newblock \emph{arXiv preprint arXiv:1706.09186}, 2017.

\bibitem[Wang \& Chen(2017)Wang and Chen]{wang2017improving}
Wang, Q. and Chen, W.
\newblock Improving regret bounds for combinatorial semi-bandits with
  probabilistically triggered arms and its applications.
\newblock In \emph{Advances in Neural Information Processing Systems}, pp.\
  1161--1171, 2017.

\bibitem[Zhang et~al.(2019)Zhang, Khaliligarekani, Tekin,
  et~al.]{zhang2019group}
Zhang, X., Khaliligarekani, M., Tekin, C., et~al.
\newblock Group retention when using machine learning in sequential decision
  making: the interplay between user dynamics and fairness.
\newblock In \emph{Advances in Neural Information Processing Systems}, pp.\
  15243--15252, 2019.

\bibitem[Zhang et~al.(2020)Zhang, Tu, Liu, Liu, Kjellstr{\"o}m, Zhang, and
  Zhang]{zhang2020fair}
Zhang, X., Tu, R., Liu, Y., Liu, M., Kjellstr{\"o}m, H., Zhang, K., and Zhang,
  C.
\newblock How do fair decisions fare in long-term qualification?
\newblock 2020.

\end{thebibliography}
\bibliographystyle{icml2021}

\newpage
\appendix

\begin{table*}[h!]
\centering
\hfill
\renewcommand{\arraystretch}{1.1}
\renewcommand{\thefootnote}{\fnsymbol{footnote}}
\vspace{4mm}
\resizebox{\textwidth}{!}{%
\begin{tabular}{c|c|l l}
\hline
Setup & Notations & \multicolumn{2}{c}{Explanations} \\
\hline
\multirow{9}{*}{\shortstack[l]{basic setup}} 
& $ K;T $ &  the number of (base) arms; time horizon & \\
& $ k;t $ &  arm index; time round $t$    & $k \in [K], t\in[T]$ \\
& $ \cP $ &  probability simplex          & $\cP\in[0,1]^{K}$ \\
& $ \epsilon $      & discretization parameter & $\epsilon \in [0,1]$ \\
& $ \cP_\epsilon $  & probability simplex after discretization with $\epsilon$ & $\cP_\epsilon \subset \cP$ \\
& $ \vp $ &  meta arm/mixed strategy      & $\vp\in\cP$\\
& $ \vp^* $ &  optimal meta arm      & \\
& $ \vp_\epsilon^* $ &  optimal meta arm in  $ \cP_\epsilon $ & \\
& $ \rewfn_k(\cdot)$    & expected reward function of arm $k$ & $\rewfn_k: [0,1]\rightarrow [0,1]$ \\  
& $ \lambda_k$         & arm $k$'s hyperparameter on tradeoff the expected reward and fairness  &  \\  
& $ p_k,p_k(t) $    & the probability on pulling arm $k$, at time $t$ & $p_k,p_k(t)\in[0,1]$\\
& $ L_k^\rewfn $        & the Lipschitz constant of $\rewfn_k$  \\ 
& $ L_k^\fairfn $       & the Lipschitz constant of $\fairfn_k$  \\ 
& $ L^* $       & the maximum $L^* = \max (1+L_k^\rewfn + |\lambda_k|L_k^\fairfn)$ &  \\ 
& $ \rewob_t $       & the realized reward at time $t$  \\ 
& $ \hat{\rewfn}_t $ & the importance weighted reward  \\
& $ \bar{\rewfn}_t(p_k) $ & the empirical reward mean of discretized arm $p_k$ \\
& $ \reg(t) $      & cumulative regret till time $t$ &  \\
& $ \Delta_{\vp} $      & the badness of meta arm $\vp$ $t$ &  \\
& $ \vp(t) = \{p_k(t)\}_{k\in[K]}$ & the meta arm deployed in time round $t$  &  \\ 
& $ \vf(t)=\{f_k(t)\}_{k\in[K]} $       & actions impact function &  \\ 
\hline
\multirow{4}{*}{\shortstack[l]{action  \\ dependent \\ bandit}} 
& $ n_t(p_k) $ & the number of times when pulling arm $k$ with prob $p_k$ till time $t$  \\
& $ N_t(\vp)  $ & number of pulls of meta arm $\vp$ till time $t$  \\
& $\cS(p_k)$ & the set of all meta arms which contain $p_k$ & $\cS(p_k) = \{\vp, p_k\in\vp\}$ \\
& $ N_t(\cS(p_k))  $ & total number of pulls of all meta arms in $\cS(p_k)$ & $N_t(\cS(p_k)) = \sum_{\vp\in\cS(p_k)}N_t(\vp)$   \\
& $ p_{\min}(\vp)$ & $p_{\min}(\vp) = \argmin_{p\in\vp} n_t(p_k)$ for some $t$  & \\ 
& $ \overbar{U}_t(\vp) $ & the empirical reward mean of meta arm $\vp$ \\
& $ U(\vp) $ & the expected reward of meta arm $\vp$ \\
\hline
\multirow{4}{*}{\shortstack[l]{history  \\ dependent \\ bandit}} 
& $ \gamma $ & time-discounted factor & $\gamma\in (0, 1)$   \\
& $ L $      & the length of phase & $L\in \N^+$ \\
& $ s_a $    & the length of approaching stage & $s_a\in \N^+$\\
& $ \rho $   & the ratio of estimation stages over each phase & $\rho\in(0, 1)$\\
& $ m $      & the index of each phase \\
& $ \Gamma_t(p_k) $ &  the set of all indexes which arm $k$ is pulled with prob $p_k$\\
& $ \vp^{(\gamma)} = \{p^{(\gamma)}_k\}_{k\in[K]}$  & the time-discounted empirical frequency\\
& $\bar{\rewfn}_m^\est(p_k)$  & the empirical reward mean of discretized arm $p_k$ in all estimation stages\\
& $ \overbar{U}_t^\est(\hat{\vp})$  & the empirical reward mean of meta arm $\vp$ in all estimation stages\\
& $ n_m^\est(p_k)$  & the number of rounds that arm $k$ is pulled with prob $p_k$ in the first $m$ phases\\
\hline
\end{tabular}
}
\caption{The summary of notations.}
\end{table*}
\vspace{-8pt}

\section{Related Work} \label{sec:related_full}

Our learning framework is based on the rich bandit learning literature~\citep{auer2002finite,lai1985asymptotically}.
However, instead of making the standard assumption of i.i.d. or adversarial rewards, 
we consider the setting in which the arm reward depends on the action history.
The settings most similar to ours are non-stationary bandits, 
including \emph{restless} bandits~\cite{slivkins2008adapting,besbes2014stochastic,garivier2011upper, verloop2016asymptotically,duran2018asymptotic}, 
in which the reward of each arm changes over time regardless of whether the arm is pulled, and
\emph{rested} bandits~\cite{levine2017rotting,seznec2019rotting,cortes2017discrepancy}, 
in which the reward of arm evolves only when it is pulled. 
In contrast, our model encodes a generic dependency of actions taken in the past and our setting is sort of a mix between the above two. 
On one hand, the reward of each arm is \emph{restless}, because even if we do not select a particular arm at step $t$, the arm's underlying state will continue to evolve (this is represented by our definition of $\vf(t)$), which will change the expected reward to be seen in the future. 
On the other hand, the changing of rewards does depend on actions, so in this sense, it is related to rested bandit. 
Technically, due to the presence of historical bias, we allow the learner to learn the optimal strategy in a continuous space which is built on the \emph{probabilistic} simplex over all arms. 
Meanwhile, our work distinguishes from prior works in that our proposed framework does not require the exact knowledge of dependency function except to the extent of a Lipschitz property and a convergence property. 

Our formulation bears similarity to reinforcement learning since our impact function encodes memory (and is in fact Markovian~\citep{ortner2012regret,tekin2010online}), 
although
we focus on studying the exploration-exploitation tradeoff in bandit formulation.
Our techniques and approaches share similar insights with Lipschitz bandits~\citep{kleinberg2008multi,slivkins2014contextual,magureanu2014lipschitz,combes2020unimodal} and 
combinatorial bandits~\citep{cesa2012combinatorial,combes2015combinatorial,chen2016combinatorial, chen2016combinatorial2}
in that we also assume the Lipschitz reward structure and consider combinatorial action space.
However, our setting is different since the arm reward explicitly depends on the learner's action history. 
We have a detailed regret comparisons with the regret of directly applying techniques in combinatorial bandits to our setting in Section~\ref{sec:discussion-action-dependent}.

There are several works that formulate delayed feedback in online learning~\citep{joulani2013online,vernade2017stochastic,pike2018bandits,kleinberg2018recharging,pike2019recovering,gael2020stochastic,cella2020stochastic,kolobov2020online}.
We discuss the ones that are mostly related to ours.
In particular,~\citet{pike2018bandits} considers the setting in which the observed reward is a sum of a number of previously generated rewards which happen to arrive in the given round. \citet{joulani2013online} and \citet{vernade2017stochastic} focus on the setting where either feedback or reward is delayed.
Our work differs from the above works in that, in our setting, the reward of the arm is influenced by the action history while the above works still consider stationary rewards (though the reward realization could be delayed).
There have also been works that study the setting that explores different generative process of reward distribution of arms, e.g., the reward of the arm depends on strategic or biased human behavior~\cite{ho2016adaptive,liu2018incentivizing,tang2019bandit}.
The more closer works to ours include considering the arm of the reward is an increasing concave function of the time since it was last played~\citet{kleinberg2018recharging}, or decreases as it was played more time~\cite{levine2017rotting,seznec2019rotting}. 
Our work differs from the above in that we formalize an impact function that permits more general form of the reward evolvement as a function of the history of arm plays.

Our work also has implications in algorithmic fairness.
one related line of works have studied fairness in the sequential learning setting, however they do not consider long-term impact of actions~\cite{joseph2016fairness, bechavod2019equal, liu2017calibrated, gupta2019individual,gillen2018online, li2019combinatorial}. 
For the explorations of delayed impacts of actions, the studies so far have focus on addressing the one-step delayed impacts or a multi-step sequential setting with full information \citep{heidari2019long,hu2018short,liu2018delayed,mouzannar2019fair,cowgill2019economics,bartlett2018consumer}.
Our work differs from the above and studies delayed impacts of actions in sequential decision making under uncertainty.

\section{Lagrangian Formulation} \label{sec:lag_for}

While our setting follows standard bandit settings and aims to maximize the utility,
it can be extended to incorporate fairness constraints as commonly seen in the discussion of algorithmic fairness. 
For example, consider the notion of group fairness, 
which aims to achieve approximate parity of certain measures across groups. 
Let $\fairfn_i(f_i(t))\in[0,1]$ be the fairness measure for group $i$ (which could reflect the socioeconomic status of the group).
One common approach is to impose constraints to avoid the group disparity. 
Let $\fairc \in [0,1]$ be the tolerance parameter, the fairness constraints at $t$ can be written as: 
$\left |\fairfn_i(f_i(t)) - \fairfn_j(f_j(t))\right| \le \fairc, \forall i,j \in [K].$
$\fairfn_i(\cdot)$ is unknown a priori and is dependent on the historical impact.
Incorporating the fairness constraints would transform the goal of the institution as a constrained optimization problem:
\begin{align*}
\max_{\vp\in\cP} ~~~ \sum_{t=1}^T \Totalrew_t(\vp(t)) \quad 
\text{s.t.} ~~~  \left |\fairfn_i(f_i(t)) - \fairfn_j(f_j(t))\right| \le \fairc, \forall i,j \in [K], \forall t\in[T].
\end{align*}

We can then utilize the Lagrangian relaxation: impose the fairness requirement as soft constraints and obtain an unconstrained optimization problem with a different utility function.
As long as we also observe (bandit) feedback on the fairness measures at every time step,
the techniques developed in this work can be extended to include fairness constraints.

To simplify the presentation, we fix a time $t$ and drop the dependency on $t$ in the notations.

\begin{definition}
The Lagrangian $\cL: \cP \times \Lambda^2 \rightarrow \Rsymb$ where $\Lambda \subseteq \Rsymb^{\binom{K}{2}}_+$ 
of our problem can be formulated as:
\begin{align}
\cL(\vp,\lambda):=  \sum_{k=1}^K p_k \rewfn_k(f_k) -\sum_{c=1}^{\binom{K}{2}} \lambda_c^+\left(\fairfn_{i_c}(f_{i_c}) - \fairfn_{j_c}(f_{j_c}) - \fairc\right) - \sum_{c=1}^{\binom{K}{2}} \lambda_c^-\left(\fairfn_{j_c}(f_{j_c}) - \fairfn_{i_c}(f_{i_c}) - \fairc\right), \nonumber
\end{align}
where $\lambda^+, \lambda^- \in \Lambda $. 
The notation $(i_c, j_c)\in\{(i,j)_{1\le i<j \le K}\}$
is a pair of combination and $c \in [K(K-1)/2]$ is the index of each pair of this combination.
\end{definition}
The problem then reduces to jointly maximize over $\vp\in\cP$ and minimize over $\lambda^+, \lambda^- \in\Lambda$.
Rearranging and with a slight abuse of notations, we have the following equivalent optimization problem:
\begin{align} \label{final_reduction}
\max_{\vp\in\cP} ~ \min_{\lambda^+, \lambda^-} \sum_{k=1}^K p_k(t)\rewfn_k(f_k(t)) + \lambda_k\fairfn_k(f_k(t)) +\fairc \sum_{c=1}^{\binom{K}{2}} (\lambda_c^+ + \lambda_c^-),
\end{align}
where $\lambda_k :=  - \sum_{c: i_c=k} (\lambda_c^+ - \lambda_c^-) + \sum_{c: j_c=k} (\lambda_c^+ - \lambda_c^-)$. 
Due to the uncertainty of reward function $\rewfn_k(\cdot)$ and \status $\pi_k(\cdot)$ (recall that our fairness criteria is defined as the parity of socio-economic status cross different groups, which we can only observe the realization drawn from an unknown distribution), we treat the above optimization problem as a hyperparameter optimization: similar to choosing hyperparameters (the Lagrange multipliers: $\lambda^+$ and $\lambda^-$) based on a validation set in machine learning tasks. 
Therefore, given a fixed set of $\lambda^+$ and $\lambda^-$, the problem in \eqref{final_reduction} can be reduced to the following:
\begin{align}
\max_{\vp\in\cP} \sum_{k=1}^K p_k(t)\cdot \rewfn_k(f_k(t)) + \lambda_k\cdot\fairfn_k(f_k(t)).
\end{align}

\section{Negative Results} \label{app:negative_res}
In this section, we show that an online algorithm which ignores its action's impact would suffer
linear regret. 
We consider two general bandit algorithms: TS (Thompson Sampling) and a mean-converging family of algorithms (which includes UCB-like algorithms).
These are the two most popular and robust bandit algorithms that can be applied to a wide range of scenarios.
We prove the negative results respectively.
In particular, we construct problem instances that could result in linear regret if the deployed algorithm ignore the action's impact.
\begin{exmpl}\label{exmpl_2}
Considering the following Bernoulli bandit instance with two arms, indexed by arm 1 and arm 2, i.e., $K= 2$.
For any $\epsilon\in [0, 1/2)$, define the expected reward of each arm as follows:
\squishlist
\item arm 1: $\rewfn_1(p) = p/(1-\epsilon)\cdot\mathbbm{1}(p\leq 1-\epsilon) + (2- \epsilon - p)\cdot\mathbbm{1}(p\geq 1- \epsilon), \quad \forall p \in [0,1]$ 
\item arm 2: $\rewfn_2(p) = p/(2\epsilon)\cdot\mathbbm{1}(p\leq \epsilon) + (-\frac{1}{2}p + \frac{1}{2}(1+\epsilon))\cdot\mathbbm{1}(p\geq \epsilon), \quad \forall p \in [0,1]$ 
\squishend
\end{exmpl}
It is easy to see that $\vp^* = \{1-\epsilon, \epsilon\}$ is the optimal strategy for the above bandit instance.

We first prove the negative result of Thompson Sampling using the above example.
The Thompson Sampling algorithm can be summarized as below. \vspace{-0.05in}
\begin{algorithm}
  \caption{Thompson Sampling}
\begin{algorithmic}[1]
  \STATE {$S_i=0, F_i=0$.}
\FOR{$t=1, 2, \ldots,$}
    \STATE For each arm $i=1,2$, sample $\theta_i(t)$ from the $\Beta(S_i+1, F_i+1)$ distribution. 
    \STATE Play arm $a_t := \arg \max_i \theta_i(t)$ and observe reward $\rewob_t$.
    \STATE If $\rewob_t=1$, then $S_{a_t}=S_{a_t}+1$, else $F_{a_t}=F_{a_t}+1$.
\ENDFOR
\end{algorithmic}
\end{algorithm}

\begin{lemma}\label{lem:TS_fail}
For the reward structure defined in Example \ref{exmpl_2},
Thompson Sampling would suffer linear regret if it doesn't consider the action's impact it deploys at every time round, namely, it takes the sample mean as the true mean reward of each arm.
\end{lemma}

Before we proceed, we first prove the following strong law of large numbers in Beta distribution.
We note that the below two lemmas are not new results and can be found in many statistical books. 
We provide proofs here for the sake of making the current work self-contained.

\begin{lemma} \label{lem: beta_delta_approx}
Consider the Beta distribution \textup{\Beta}$(a\alpha + 1, b\alpha + 1)$ whose pdf is defined as $f(x, \alpha) = \frac{[x^a (1-x)^b]^\alpha}{B(a\alpha + 1, b\alpha + 1)}$, where $B(\cdot)$ is the beta function, then for any positive $(a,b)$ such that $a + b = 1$, when $\alpha \rightarrow \infty$, the limit of $f(x, \alpha)$ can be characterized by Dirac delta function $\delta(x -a)$.
\end{lemma}
\begin{proof}
By Stirling's approximation, we can write the asymptotics of beta function as follows:
\begin{align*}
B(x, y) \sim \sqrt{2\pi}\frac{x^{x - 0.5} y^{y - 0.5}}{(x+y)^{x+y - 0.5}}.
\end{align*}
Thus, when $\alpha \rightarrow \infty$, i.e., for large $a\alpha+1$ and $b\alpha+1$, we can approximate the pdf $f(x, \alpha)$ in the following:
\begin{align*}
f(x, \alpha) \sim \sqrt{\frac{a + 2}{2\pi ab}} h^\alpha(x),
\end{align*}
where $h(x) :=(x/a)^a\big(\frac{1-x}{b}\big)^b$.
It's easy to see that $h(x)$ has a unique maximum at $a$, by invoking Lemma \ref{lem: delta_approx} will complete the proof.
\end{proof}

\begin{lemma} \label{lem: delta_approx}
Let $h:[0,1]\rightarrow \R^+$ be any bounded measurable non-negative function with a unique maximum at $x^*$, and suppose $h$ is continuous at $x^*$. 
For $\lambda>0$ define $h_\lambda(x) = C_\lambda h^\lambda(x)$ where $C_\lambda$ normalizes such that $\int_{0}^1h_\lambda(x) dx = 1$. 
Consider any continuous function $f$ defined on $[0,1]$ and $\epsilon > 0$, then we have
$\lim_{\lambda \rightarrow \infty}\int_{h(x)\leq h(x^*)-\epsilon }h_\lambda(x)f(x)dx=0$ and $\lim_{\lambda \rightarrow \infty}\int_{0}^1 h_\lambda(x)f(x)dx=f(x^*)$. 
\end{lemma}
\begin{proof}
For any $\delta > 0$, we have
\begin{align*}
&\bigg|\int_{0}^1 h_\lambda(x)f(x)dx - f(x^*)\bigg|\\
=~& \bigg|\int_{0}^1 h_\lambda(x)\big(f(x) - f(x^*)\big) dx\bigg| \\
\leq~&  \bigg|\int_{|x - x^*| \leq \delta} h_\lambda(x)\big(f(x) - f(x^*)\big) dx\bigg| +  \bigg|\int_{|x - x^*| > \delta} h_\lambda(x)\big(f(x) - f(x^*)\big) dx\bigg| \\
\leq~&  \bigg|\int_{|x - x^*| \leq \delta} h_\lambda(x)\big(f(x) - f(x^*)\big) dx\bigg|  + \max \big|f(x) - f(x^*)\big| \bigg| \int_{|x - x^*| > \delta} h_{\lambda}(x) dx\bigg|.
\end{align*}
For any $\delta > 0$, and due to the continuous property of $f$ on $x^*$, which further implies that there exists a constant $c >0$ such that $|f(x)- f(x^*)|<\delta/2$ whenever $|x - x^*| < c$.
Thus, given $c > \delta$, we have
\begin{align*}
\bigg|\int_{0}^1 h_\lambda(x)f(x)dx - f(x^*)\bigg| \leq  \epsilon/2 + \max \big|f(x) - f(x^*)\big| \bigg| \int_{|x - x^*| > \delta} h_{\lambda}(x) dx\bigg|.  \label{goal}
\end{align*}
It suffices to show that the second term in RHS of above inequality will converge to 0 as $\lambda \rightarrow \infty$.
Let $||h||_{\infty, \delta}$ denote the $L^\infty$ norm of $h$ when $h$ is restricted to $\{|x - x^*| > \delta\}$. 
Note that for any nonnegative integrable functions $h$, we have 
\begin{align*}
\lim_{\lambda\rightarrow \infty} \bigg(\int_{0}^1 h^\lambda(x)dx\bigg)^{1/\lambda} = ||h||_{\infty}.
\end{align*}
Recall the definition of $C_\lambda = \frac{1}{\int_{0}^1 h^\lambda(x)dx}$, thus, we have $\lim_{\lambda\rightarrow \infty} C_\lambda^{1/\lambda} = \frac{1}{||h||_{\infty}}$, which immediately showing that
\begin{align*}
\bigg(\int_{|x - x^*| > \delta} h_{\lambda}(x) dx\bigg)^{1/\lambda} = C_\lambda^{1/\lambda} \bigg(\int_{|x-x^*| > \delta} h^\lambda(x)dx\bigg)^{1/\lambda},
\end{align*} 
which further implies that $||h||_{\infty,\delta}/||h||_{\infty} < 1$. 
Thus, there must exist $\lambda_0$ such that $\forall \lambda > \lambda_0$, 
\begin{align}
\bigg(\int_{|x - x^*|>\delta} h_\lambda(x)dx\bigg)^{1/\lambda} < \gamma < 1.
\end{align}
Since $\gamma < 1$, we then have $\lim_{\lambda \rightarrow \infty} \gamma^\lambda = 0$, this implies the second term of RHS of (\ref{goal}) converging to 0 as $\lambda \rightarrow \infty$.
\end{proof}

We now ready to prove Lemma~\ref{lem:TS_fail}.
\begin{proof}
We prove this by contradiction.
Let $\reg(T)$ denote the expected regret incurred by TS up to time round $T$, and $N_t(\vp) = \sum_{s =1}^t \mathbbm{1}(\vp(s) = \vp)$ denote the number of rounds when the algorithm deploys the (mixed) strategy $\vp \in \Delta_K$.
Furthermore, let $S_i(t) (\text{resp. }F_i(t))$ denote the received $1_s(\text{resp. } 0_s)$ of arm $i$ up to time round $t$.
Recall that in Thompson Sampling, we have $\Psymb(a_t = 1) = \Psymb\big(\theta_1(t) > \theta_2(t)\big)$.
By the reward function defined in Example \ref{exmpl_2}, it's immediate to see that 
\begin{align*}
S_1(T) \geq (1-\epsilon)N_T(\vp^*); \quad F_1(T)\leq T - N_T(\vp^*); \quad S_2(T)\geq 0.5\epsilon N_T(\vp^*); \quad F_2(T)\geq 0.5\epsilon N_T(\vp^*).
\end{align*}
Now suppose Thompson Sampling achieves sublinear regret, i.e., $\reg(T) = o(T)$, which implies following
\begin{align*}
\lim_{T\rightarrow \infty} \frac{T - N_T(\vp^*)}{T} = 0.
\end{align*}
Thus, by the strong law of large numbers and invoking Lemma \ref{lem: beta_delta_approx}, the sample $\theta_1(T+1)\sim\Beta(S_1(T), F_1(T))$ and $\theta_2(T+1)\sim\Beta(S_2(T), F_2(T))$ will converge as follows:
\begin{align*}
\lim_{T\rightarrow \infty} \theta_1(T+1)  = 1; 
\quad \lim_{T\rightarrow \infty} \theta_2(T+1)  = 0.5.
\end{align*}
Then it's almost surely that $\lim_{T\rightarrow \infty} \Psymb(a_{T+1} = 1) = \lim_{T\rightarrow \infty}\Psymb\big(\theta_1(T+1) > \theta_2(T+1)\big) = 1$.
This leads to following holds for sure
\begin{align*}
S_1(s+1) = S_1(s) + 1, \forall s > T.
\end{align*}
Thus, consider the regret incurred from the $(T+1)-$th round to $(2T)-$th round, the regret will be 
\begin{align*}
\reg(2T)- \reg(T) = \sum_{s={T+1}}^{2T} U(\vp(s)) = 0.5T\epsilon,
\end{align*}
where the second equality follows that $ \vp(s) = (1, 0)$ holds almost surely from $T+1$ to $2T$.
This shows that $\lim_{T\rightarrow\infty} \frac{\Esymb[\reg(2T)]}{2T} = \epsilon/4$, which contradicts that the algorithm achieves the sublinear regret. 
\end{proof}

We now show that a general class of algorithms, which are based on \emph{mean-converging}, will suffer linear regret if it ignores the action's impact.
This family of algorithms includes UCB algorithm in classic MAB problems.
\begin{definition}[Mean-converging Algorithm~\citep{schmit2018human}]
Define $I_k(t) = \{s: a_s = k, s < t\}$ as the set of time rounds such the arm $k$ is chosen.
Let $\bar{\rewfn}_k(t) = \frac{1}{|I_k(t)|} \sum_{s\in I_k(t)} \rewob_s$ be the empirical mean of arm $k$ up to time $t$.
The mean-converging algorithm $\cA$ assigns $s_k(t)$ for each arm $k$ if following holds true:
\squishlist
\item $s_k(t)$ is the function of $\{\rewob_s: s\in I_k(t)\}$ and time $t$;
\item $\Psymb(s_k(t) = \bar{r}_k(t)) = 1$ ~~ if ~~$\liminf_{t}\frac{|I_k(t)|}{t} > 0$.
\squishend
\end{definition}
\begin{lemma}
For the reward structure defined in Example \ref{exmpl_2},
the mean-converging Algorithm will suffer linear regret if it mistakenly take the sample mean as the true mean reward of each arm.
\end{lemma}
\begin{proof}
We prove above lemma by contradiction.
Let $N^\cA_t(\vp)$ denote the number of plays with deploying the strategy $\vp$ by algorithm $\cA$ till time $t$.
Suppose a mean-converging Algorithm $\cA$ achieves sublinear regret, then it must have $\lim_{T\rightarrow \infty} {N^\cA_T(\vp^*)}/{T} > 0$ and $\lim_{T\rightarrow \infty} \big({T - N^\cA_T(\vp^*)}\big)/{T} = o(T)$. 
By the definition of mean-converging algorithm and recall the reward structure defined in Example \ref{exmpl_2}, the score $s_T(1)$ assigned to arm 1 by the algorithm $\cA$ must be converging to 1, and the score of $s_T(2)$ assigned to arm 2 must be converging to 0.5.
By the strong law of large numbers, it suffices to show that $\Psymb(\vp(t) = \{1,0\}) = 1, \forall t\geq T+1$, which implies the algorithm $\cA$ would suffer linear regret after $T$ time rounds and thus completes the proof.
\end{proof}

\section{Missing Proofs for Action-Dependent Bandits} 
\label{all_proof_of_action_dependent_bandit}

\subsection{The naive method that directly utilize techniques from Lipschitz bandits} 
We first give a naive approach which directly applies Lipschitz bandit technique to our action-dependent setting.
Recall that each meta arm $\vp$ specifies the probability $p_k\in[0,1]$ for choosing each base arm $k$.
We \emph{uniformly} discretize each $p_k$ into intervals of a fixed length $\epsilon$, with carefully chosen $\epsilon$ such that $1/\epsilon$ is an positive integer.
Let $\cP_\epsilon$ be the space of discretized meta arms,
i.e., for each $\vp =\{p_1,\ldots,p_K\}\in\cP_\epsilon$, 
$\sum_{k=1}^K p_k=1$ and $p_k \in \{0, \epsilon, 2\epsilon, \ldots, 1\}$ for all $k$.
We then run standard bandit algorithms on the finite set $\cP_\epsilon$.

There is a natural trade-off on the choice of $\epsilon$, which controls the complexity of arm space and the discretization error.  
show that, with appropriately chosen $\epsilon$, this approach can achieve sublinear regret (with respect to the optimal arm in the non-discretized space $\cP$).

\begin{lemma} \label{lemma:regret_bound_naive}
Let $\epsilon = \Theta \big(\big(\frac{\ln T}{T }\big)^\frac{1}{K+1}\big)$.
Running a bandit algorithm which achieves optimal regret $\cO(\sqrt{|\cP_\epsilon|T\ln T})$ on the strategy space $\cP_\epsilon$ attains the following regret (w.r.t. the optimal arm in non-discretized $\cP$):
$\reg(T) = \cO\big(T^{\frac{K}{K+1}}{(\ln T)}^{\frac{1}{K+1}}\big)$.
\end{lemma}
\begin{proof}
As mentioned, we \emph{uniformly} discretize the interval $[0, 1]$ of each arm into interval of a fixed length $\epsilon$. 
The strategy space will be reduced as $\cP_\epsilon$, which we use this as an approximation for the full set $\cP$.
Then the original infinite action space will be reduces as finite $\cP_\epsilon$, and we run an off-the-shelf MAB algorithm $\cA$, such as UCB1 or Successive Elimination, that only considers these actions in $\cP_\epsilon$. 
Adding more points to $\cP_\epsilon$ makes it a better approximation of $\cP$, but also increases regret of $\cA$ on $\cP_\epsilon$. 
Thus, $\cP_\epsilon$ should be chosen so as to optimize this tradeoff.
Let 
$\vp_\epsilon^* := \sup_{\vp \in \cP_\epsilon} \sum_{k=1}^K p_k\rewfn_k(p_k)$
denote the best strategy in discretized space $\cP_\epsilon$. 
At each round, the algorithm $\cA$ can only hope to approach expected reward $U(\vp_\epsilon^*)$, and together with additionally suffering \emph{discretization error}:
\[
\DE_\epsilon := U(\vp^*) - U(\vp_\epsilon^*).
\]
Then the expected regret of the entire algorithm is:
\begin{align*}
\reg(T) & = T \cdot U(\vp^*) - \texttt{Reward}(\cA) \\
& = T \cdot U(\vp_\epsilon^*) - \texttt{Reward}(\cA) + T(U(\vp^*) - U(\vp_\epsilon^*)) \\
& = \E[\reg_{\epsilon} (T)]+ T \cdot \DE_\epsilon,
\end{align*}
where $\texttt{Reward}(\cA)$ is the total reward of the algorithm, and $\reg_{\epsilon} (T)$ is the regret relative to $U(\vp_\epsilon^*)$.
If $\cA$ attains optimal regret $\cO(\sqrt{KT \ln T} )$ on any problem instance with time horizon $T$ and $K$ arms, then, 
\[
\reg(T) \leq \cO(\sqrt{\big|\cP_{\epsilon}\big| T \ln T} ) + T \cdot \DE_\epsilon.
\]
Thus, we need to choose $\epsilon$ to get the optimal trade-off between the size of $\cP_\epsilon$ and its discretization error.
Recall that $r_k(\cdot)$ 
is Lipschitz-continuous with the constant of $L_k$,
thus, we could bound the $\DE_\epsilon$ by restricting $\vp_\epsilon^*$ to be nearest w.r.t $\vp^*$. 
Let $L^* = \max_{k\in[K]} (1+L_k)$, 
then it's easy to see that
\[
\DE_\epsilon = \Omega(KL^*\epsilon).
\]
Thus, the total regret can be bounded above from:
\[
\reg(T) \leq \cO\left(\sqrt{(1/\epsilon + 1)^{K-1} T \ln T} \right) + \Omega(TKL^*\epsilon).
\]
By choosing $\epsilon = \Theta\left(\left( \frac{\ln T}{T(L^*)^2}\right)^{\frac{1}{K+1}}\right)$ we obtain:
\[
\reg(T) \leq \cO(cT^{\frac{K}{K+1}}{(\ln T)}^{\frac{1}{K+1}}).
\]
where $c = \Theta\left(K(L^*)^{\frac{K-1}{K+1}}\right)$.
\end{proof}

\subsection{Missing Discussions and Proofs of Theorem~\ref{theorem:key_theorem_vanilla}} \label{proof_regret_bound_action_bandit}

\paragraph{\textbf{Step $1$: Bounding the error of $|\overbar{U}(\vp)- U(\vp)|$.}}
For any $\vp = \{p_1, \ldots, p_K\}$, define the empirical reward 
$\overbar{U}_t(\vp) = \sum_{k=1}^K p_k\bar{r}_t(p_k)$.
The first step of our proof is to bound $\Psymb(|\overbar{U}_t(\vp)- U(\vp)| \leq \delta)$ for each meta arm $\vp = \{p_1, \ldots, p_K\}$ with high probability.\footnote{We use $\delta$ to denote the estimation error, as $\epsilon$ has been used as the discretization parameter.}
Using the Hoeffding's inequality, we obtain 
\begin{align*}
\Psymb\big(|\overbar{U}_t(\vp)- U(\vp)| \geq \delta\big) 
& = \Psymb\bigg(\bigg|\sum_k \frac{\sum_{s\in\cT_t(p_k)}\hat{\rewfn}_s(p_k)}{n_t(p_k)}- \sum_k p_k r(p_k)\bigg| \geq \delta\bigg) \\
& \leq 2\exp \bigg(-\frac{2\delta^2}{\sum_k \frac{1}{n_t(p_k)}} \bigg)
\le 2\exp \bigg(-\frac{2\delta^2 n_t(p_{\min}(\vp))}{K} \bigg),
\end{align*}
where $p_{\min}(\vp) := \argmin_{p_k\in \vp} n_t(p_k)$. 
By choosing 
$\delta = \sqrt{\frac{K\ln t}{n_t(p_{\min}(\vp))}}$
in the above inequality, 
for each meta arm $\vp$ at time $t$, we have that
$|\overbar{U}_t(\vp) - U(\vp)| \leq \sqrt{K\ln t/n_t(p_{\min}(\vp))}$, 
with the probability at least $1-2/t^2$.

\paragraph{\textbf{Step $2$: Bounding the probability on deploying suboptimal meta arm.}}
With the above high probability bound we obtain in Step $1$, 
we can construct an $\UCB$ index for each meta arm $\vp \in \cP_\epsilon$:
\begin{align} \label{ucb}
& \UCB_t(\vp) =  \overbar{U}_t(\vp) + \sqrt{\frac{K\ln t }{n_t(p_{\min}(\vp))}}.
\end{align}
The above constructed $\UCB$ index gives the following guarantee:
\begin{lemma} \label{lemma:key_lemma_vanilla}
At any time round $t$, for a suboptimal meta arm $\vp$, if it satisfies $n_t(p_{\min}(\vp)) \geq {4K\ln t}/{\Delta_{\vp}^2}$, then $\textup{\UCB}_t(\vp) < \textup{\UCB}_t(\vp_\epsilon^*)$ with the probability at least $1-4/t^2$.
Thus, for any $t$,
\begin{align*}
& \Psymb\left(\vp(t) = \vp|n_t(p_{\min}(\vp))\geq {4K\ln t}/{\Delta_{\vp}^2}\right) \leq 4t^{-2},
\end{align*}
where $\Delta_{\vp}$ denotes the badness of meta arm $\vp$. 
\end{lemma}
\begin{proof}\label{Proof_lemma_key_lemma_vanilla}
We prove this lemma by considering two ``events'' which occur with high probability: 
(1) the $\UCB$ index of each meta arm will concentrate on the true mean utility of $\vp$;
(2) the empirical mean utility of each meta arm $\vp$ will also concentrate on the true mean utility of $\vp$.
We then show that the probability of either one of the events not holding is at most $4/t^2$.
By a union bound we prove above desired lemma.
\begin{align*}
\UCB_t(\vp) & = \sum_{k=1}^K p_k\bar{\rewfn}_t(p_k)  + \sqrt{\ln t \frac{K}{n_t(p_{\min}(\vp))}} \\
& \labelrel\leq{ineq21} \sum_{k=1}^K p_k\bar{\rewfn}_t(p_k) + \Delta_{\vp}/2   < \left(\sum_{k=1}^K p_k\rewfn_k(p_k) + \Delta_{\vp}/2\right) + \Delta_{\vp}/2 & \tag*{By \text{Event 1}}\\
& = \sum_{k=1}^K p_{k,\epsilon}^*\rewfn_k(p_{k, \epsilon}^*)  < \sum_{k=1}^K p_{k, \epsilon}^*\bar{\rewfn}_t(p_{k, \epsilon}^*) +  \sqrt{\ln t \frac{K}{n_t(p_{\min}(\vp_\epsilon^*))}}  & \tag*{By \text{Event 2}}\\
& = \UCB_{t}(\vp_\epsilon^*),
\end{align*}
where $\vp_\epsilon^* = (p_{1, \epsilon}^*, \ldots, p_{K, \epsilon}^*)$.
The first inequality~\eqref{ineq21} comes from that $n_t(p_{\min}(\vp))\geq \frac{4K\ln t}{\Delta_{\vp}^2}$ and the probability of third inequality or fifth inequality not holding is at most $4/t^2$.
\end{proof}

Intuitively, Lemma \ref{lemma:key_lemma_vanilla} 
essentially shows that for a meta arm $\vp$, if its $n_t(p_{\min}(\vp))$ is sufficiently sampled with respect to $\Delta_{\vp}$, that is, sampled at least ${4K\ln t}/{\Delta_{\vp}^2}$ times, we know that the probability that we hit this suboptimal meta arm is very small.

\paragraph{\textbf{Step $3$: Bounding the $\Esymb[n_T(p_{\min}(\vp))]$.}}
Ideally, we would like to bound the number of the selections on deploying the suboptimal meta arm, i.e., $N_T(\vp)$, in a logarithmic order of $T$.
However, if we proceed to bound this by separately considering each meta arm, the final regret bound will have an order with exponent in $K$ since the number of meta arms grows exponentially in $K$.
Instead, we turn to bound $\Esymb[n_T(p_{\min}(\vp))]$. 
Recall that by the definitions of $n_T(p)$ and $p_{\min}(\vp)$, the pulls of $\vp$ is upper bounded by its $n_T(p_{\min}(\vp))$.
This quantity will help us to reduce the exponential $K$ to the polynomial $K$.
This is formalized in the following lemma.
\begin{lemma} \label{lemma: sufficient_sample}
For each suboptimal meta arm $\vp \neq \vp_\epsilon^*$, we have that $\Esymb[n_T(p_{\min}(\vp))] \leq \frac{4K\ln T}{\Delta_{\vp}^2} + \cO(1)$.
\end{lemma}
\begin{proof} \label{Proof_lemma_sufficient_sample}
To simplify notations, for each discretized arm $p_k$, we define the notion of \emph{super set} $\cS(p_k) = \{\vp: p_k \in \vp\}$ which contains all the meta arms that include this discretized arm.
For suboptimal meta arm $\vp \neq \vp_\epsilon^*$ and its $p_{\min}(\vp)$, we have 
\begin{align*}
& \Esymb[n_T(p_{\min}(\vp))]  \\
\labelrel={eq3}~& 1 + \Esymb \bigg[\sum_{t = \left \lceil K/\epsilon \right \rceil + 1}^T \mathbbm{1}\left(\vp(t) = \vp, \vp\in \cS(p_{\min}(\vp))\right)\bigg] \\
=~& 1 + \Esymb \bigg[\sum_{t = \left \lceil K/\epsilon \right \rceil + 1}^T \mathbbm{1}\left(\vp(t) = \vp, \vp\in \cS(p_{\min}(\vp)); n_t(p_{\min}(\vp)) < \frac{4K\ln t}{\Delta_{\vp}^2}\right)\bigg] \\
& + \Esymb \bigg[\sum_{t = \left \lceil K/\epsilon \right \rceil + 1}^T \mathbbm{1}\left(\vp(t) = \vp, \vp\in \cS(p_{\min}(\vp)); n_t(p_{\min}(\vp)) \geq \frac{4K\ln t}{\Delta_{\vp}^2}\right)\bigg] \\
\labelrel\leq{ineq4}~& \frac{4K\ln T}{\Delta_{\vp}^2} +  \Esymb \bigg[\sum_{t = \left \lceil K/\epsilon \right \rceil + 1}^T \mathbbm{1}\left(\vp(t) = \vp, \vp\in \cS(p_{\min}(\vp)); n_t(p_{\min}(\vp)) \geq \frac{4K\ln t}{\Delta_{\vp}^2}\right)\bigg] \\
=~& \frac{4K\ln T}{\Delta_{\vp}^2} +  \sum_{t = \left \lceil K/\epsilon \right \rceil + 1}^T\Psymb \left(\vp(t) = \vp, \vp\in \cS(p_{\min}(\vp)); n_t(p_{\min}(\vp)) \geq \frac{4K\ln t}{\Delta_{\vp}^2}\right) \\
=~& \frac{4K\ln T}{\Delta_{\vp}^2} +  \sum_{t = \left \lceil K/\epsilon \right \rceil + 1}^T\Psymb \left(\vp(t) = \vp, \vp\in \cS(p_{\min}(\vp))\bigg| n_t(p_{\min}(\vp)) \geq \frac{4K\ln t}{\Delta_{\vp}^2} \right)\Psymb \left(n_t(p_{\min}(\vp)) \geq \frac{4K\ln t}{\Delta_{\vp}^2}\right)\\
\labelrel\leq{ineq5}~& \frac{4K\ln T}{\Delta_{\vp}^2} + \frac{2\pi^2}{3}.
\end{align*}
We add $1$ in the first equality to account for $1$ (step~\eqref{eq3}) initial pull of every discretized arm by the algorithm (the initialization phase).
In step~\eqref{ineq4}, suppose for contradiction that the indicator $\mathbbm{1}\left(\vp(t) = \vp, \vp\in \cS(p_{\min}(\vp)); n_t(p_{\min}(\vp)) < S\right)$ takes value of $1$ at more than $S-1$ time steps, where $S = \frac{4K\ln T}{\Delta_{\vp}^2}$. 
Let $\tau$ be the time step at which this indicator is $1$ for the $(S-1)$-th time.
Then the number of pulls of all meta arms in $\cS(p_{\min}(\vp))$ is at least $L$ times until time $\tau$ (including the initial pull), and for all $t \ge  \tau$, $n_t(p_{\min}(\vp)) \geq S$ which implies $n_t(p_{\min}(\vp)) \geq \frac{4K\ln t}{\Delta_{\vp}^2}$.
Thus, the indicator cannot be $1$ for any $t\geq \tau$, contradicting the assumption that the indicator takes value of $1$ more than $L$ times.
This bounds $1 + \Esymb \left[\sum_{t\ge\left \lceil K/\epsilon \right \rceil + 1} \mathbbm{1}\left(\vp(t) = \vp, \vp\in \cS(p_{\min}(\vp)); n_t(p_{\min}(\vp)) < S\right)\right]$ by $S$.
In step~\eqref{ineq5}, we apply the lemma \ref{lemma:key_lemma_vanilla} to bound the first conditional probability term and use the fact that the probabilities cannot exceed $1$ to bound the second probability term.
\end{proof}
We use this connection in the following step to reduce the computation of regret on pulling all suboptimal meta arms so that to calculate the regret via the summation over discretized arms.

\paragraph{\textbf{Wrapping up: Proof of Theorem~\ref{theorem:key_theorem_vanilla}.}}
We are now ready to prove Theorem~\ref{theorem:key_theorem_vanilla}. 
We first define notations that are helpful for our analysis.
To circumvent the summation over all feasible suboptimal arms $\{\vp\}$,
for each discretized arm $p_k$, we define the notion of \emph{super set} $\cS(p_k) := \{\vp: p_k \in \vp\}$ which contains all suboptimal meta arms that include this discretized arm.
With a slight abuse of notations, we also sort all meta arms in $\cS(p_k)$ as $\vp_1, \vp_2, \ldots, \vp_{I(p_k)}$
in ascending order of their expected rewards, where $I(p_k) := |\cS(p_k)|$ is the cardinality of the super set $\cS(p_k)$.
For $\vp_l \in \cS(p_k)$, we also define $\Delta^{p_k}_l := \Delta_{\vp_l}$ where $l \in [I(p_k)]$, and specifically $\Delta^{p_k}_{\min}  := \min_{\vp\in \cS(p_k)} \Delta_{\vp} =\Delta_{I(p_k)}^{p_k}$; $\Delta^{p_k}_{\max}  := \max_{\vp\in \cS(p_k)} \Delta_{\vp} = \Delta_1^{p_k}$.
Let $\reg_{\epsilon} (T)$ denote the regret relative to the best strategy in the discretized space parameterized by $\epsilon$.
With these notations, we first establish the following instance-dependent regret.
\begin{lemma}\label{lemma: instance-dependent regret}
Following the $\UCB$ designed in (\ref{ucb}), we have the following instance-dependent regret on the discretized arm space:
$\reg_\epsilon(T) \leq \left\lceil K/\epsilon \right\rceil\cdot\left(\Delta_{\max}+\cO(1)\right)+ \sum_{p_k: \Delta_{\min}^{p_k} > 0}{8K\ln T}/{\Delta_{\min}^{p_k}}$,
where $\Delta_{\max} := \max_{p_k}\Delta_{\max}^{p_k}$.
\end{lemma}

\begin{proof}
Note that by definition, we can compute the regret $\reg_{\epsilon} (T)$ as follows:
\begin{align}
\reg_\epsilon(T) = \sum_{\vp \in \cP_\epsilon} \Esymb[N_T(\vp)] \Delta_{\vp} \leq \sum_{p_k} \sum_{l\in[I(p_k)]}  \Esymb[N_T(\vp_l)] \Delta_l^{p_k}.
\label{regret_2}
\end{align}
Observe that, by Lemma \ref{lemma: sufficient_sample}, for each discretized arm $p_k$, there are two possible cases:
\squishlist
  \item There exists a meta arm $\vp_l\in \cS(p_k)$, and its $p_{\min}(\vp_l) = p_k$. 
  Then by linearity of expectation, we can bound the expectation of total number of pulls for all $\vp_{l'} \in \cS(p_k)$ as follows
  \begin{align*}
  \sum_{\vp_{l'} \in \cS(p_k)} \Esymb[N_T(\vp_{l'})]  = \Esymb[n_T(p_k)] \leq \frac{4K\ln T}{(\Delta^{p_k}_{\min})^2} + \cO(1).
  \end{align*}
  \item There exists no meta arm $\vp\in \cS(p_k)$, and $p_{\min}(\vp)$ for each $\vp$ is $p_k$. 
  In this case, for each $\vp_l \in \cS(p_k)$, there always exists another discretized arm $p'$ that is included in $\vp_l$ such that $p' = p_{\min}(\vp_l)$ but $p'\neq p_k$.
  Thus, for each $\vp_l \in \cS(p_k)$, together with other meta arms which also include discretized arm $p'$ as $\vp_l$, we have that
  \begin{align*}
  & \sum_{\vp \in \bigcup_{p' \in \vp} \vp} \Esymb[N_T(\vp)]  = \sum_{\vp\in \cS(p')}  \Esymb[N_T(\vp)] \\
  = ~~~ &  \Esymb[n_T(p')]\leq  \frac{4K\ln T}{(\Delta^{p'}_{\min})^2} + \cO(1).
  \end{align*}
\squishend
The above observations imply that even though we can not find any meta arm $\vp$ in $\cS(p_k)$ such that $p_{\min}(\vp) =p_k$, we can always carry out similar analysis by finding another discretized arm $p' \in \vp $ but $p' \neq p_k$, such that $p' = p_{\min}(\vp)$. 
Thus, for each discretized arm $p_k$, we can focus on the case where $p_k$ is able to attain the minimum $n_t(p_k)$ for some $\vp \in \cS(p_k)$.
For analysis convenience, instead of looking at the counter of $\vp$, i.e., $n_t(p_{\min}(\vp))$, we will define a counter $c(p_k)$ for each discretized arm $p_k$ and the value of $c(p_k)$ at time $t$ is denoted by $c_t(p_k)$.
The update of $c_t(p_k)$ is as follows: 
For a round $t > \left \lceil K/\epsilon \right \rceil$ (here $\left \lceil K/\epsilon \right \rceil$ is the number of rounds needed for initialization), let $\vp(t)$ be the meta arm selected in round $t$ by the algorithm. 
Let $p_k = \argmin_{p_k\in\vp(t)} c_{t-1}(p_k)$. We increment $c(p_k)$ by one, i.e., $c_t(p_k) = c_{t-1}(p_k) + 1$. 
In other words, we find the discretized arm $p_k$ with the smallest counter in $\vp(t)$ and increment its counter. 
If such $p_k$ is not unique, we pick an arbitrary discretized arm with the smallest counter. 
Note that the initialization gives $\sum_{p_k}c_{\left \lceil K/\epsilon \right \rceil}(p_k) = \left \lceil K/\epsilon \right \rceil$. 
It is easy to see that for any $p_k = p_{\min}(\vp)$, we have $n_t(p_k) = c_t(p_k)$.

With the above change of counters, Lemma~\ref{lemma:key_lemma_vanilla} and Lemma~\ref{lemma: sufficient_sample} then have the implication on selecting discretized arm $p_k\notin \vp^*_\epsilon$ given its counter $c_t(p_k)$. 
To see this, for each $\vp_l \in \cS(p_k)$, we define sufficient selection of discretized arm $p_k$ with respect to $\vp_l$ as $p_k$ being selected ${4K\ln T}/\big(\Delta^{p_k}_l\big)^2$ times and $p_k$'s counter $c(p_k)$ being incremented in these selected instances. 
Then Lemma~\ref{lemma:key_lemma_vanilla} tells us when $p_k$ is sufficiently selected with respect to $\vp_l$, the probability that the meta arm $\vp_l$ is selected by the algorithm is very small.
On the other hand, when $p_k$'s counter $c(p_k)$ is incremented, but if $p_k$ is under-selected with respect to $\vp_l$, we incur a regret of at most $\Delta^{p_k}_j$ for some $j\le l$. 

Define $C_T(\Delta) := \frac{4K\ln T}{\Delta^2}$, the number of selection that is considered sufficient for a meta arm with reward $\Delta$ away from the optimal strategy $\vp_\epsilon^*$ with respect to time horizon $t$.
With the above analysis, we define following two situations for the counter of each discretizad arm: 
\begin{align*}
    c_T^{l, \suf}(p_k) 
    &:=  \sum_{t = \left \lceil K/\epsilon \right \rceil + 1}^T \mathbbm{1}\left(\vp(t) = \vp_l, c_t(p_k) > c_{t-1}(p_k) > C_T(\Delta_l^{p_k})  \right), \\
     c_T^{l, \und}(p_k) 
    &:= \sum_{t = \left \lceil K/\epsilon \right \rceil + 1}^T \mathbbm{1}\left(\vp(t) = \vp_l, c_t(p_k) > c_{t-1}(p_k),  c_{t-1}(p_k) \le C_T(\Delta_l^{p_k})  \right).
\end{align*}
Clearly, we have $c_T(p_k) = 1 + \sum_{l\in I(p_k)}\big(c_T^{l, \suf}(p_k) + c_T^{l, \und}(p_k)\big)$.
With these notations, we can write~\eqref{regret_2} as follows:
\begin{align}
    \reg_\epsilon(T) \le \Esymb\bigg[\sum_{p_k} \bigg(\Delta^{p_k}_{\max} + \sum_{l\in [I(p_k)]}\left(c_T^{l, \suf}(p_k) + c_T^{l, \und}(p_k)\right)\cdot\Delta_l^{p_k}\bigg)\bigg]. \label{regret_3}
\end{align}
The proof of this lemma will complete after establishing following two claims:
\begin{align}
\text{Claim 1:}  ~~ &\Esymb\bigg[\sum_{p_k}\sum_{l\in [I(p_k)]} c_T^{l, \suf}(p_k) \bigg] \le \left \lceil K/\epsilon \right \rceil \cdot \cO(1).  \label{Claim1}\\
\text{Claim 2:}  ~~ &\Esymb\bigg[\sum_{p_k}\sum_{l\in [I(p_k)]} c_T^{l, \und}(p_k) \Delta_l^{p_k}\bigg] 
\le \sum_{p_k}\left({(4K\ln T)}/{\Delta_\text{min}^{p_k}}+ 4K\ln T \left({1}/{\Delta_\text{min}^{p_k}} - {1}/{\Delta_\text{max}^{p_k}} \right) \right)  \label{Claim2}. 
\end{align}
We now first prove the Claim 1 as in~\eqref{Claim1}, i.e., for any $t > \left \lceil K/\epsilon \right \rceil$, we have following upper bound over counters of sufficiently selected discretized arms.
To see this, by definition of $c_T^{l, \suf}(p_k)$, it reduces to show that for any $T \ge t > \left \lceil K/\epsilon \right \rceil$,
\begin{align*}
      & \Esymb\bigg[\sum_{p_k}\sum_{l\in [I(p_k)]}\mathbbm{1}\left(\vp(t) = \vp_l, c_t(p_k) > c_{t-1}(p_k) > C_T(\Delta_l^{p_k})  \right) \bigg] \\
     =~&  \sum_{p_k}\sum_{l\in [I(p_k)]} \Psymb\left(\vp(t) = \vp_l, p_k = p_{\min}(\vp_l); \forall p  \in \vp_l, c_{t-1}(p) > C_T(\Delta_l^{p_k})\right)\\
     \labelrel\le{ineq20}~& \left \lceil 4K/\epsilon \right \rceil\cdot t^{-2},
\end{align*}
where the last step~\eqref{ineq20} is due to Lemma~\ref{lemma:key_lemma_vanilla}, thus~\eqref{Claim1} follows from a simple series bound.

We now proceed to analyze the discretized arms that are not sufficiently included in the meta arm chosen by the algorithm and prove the Claim 2 as in~\eqref{Claim2}. 
For any under-selected discretized arm $p_k$, its counter $c(p_k)$ will increase from $1$ to $C_T(\Delta_{\min}^{p_k})$. 
To simplify the notation, we set $C_T(\Delta_0^{p_k}) = 0$.
Suppose that at round $t$, $c(p_k)$ is incremented, and $c_{t-1}(p_k) \in (C_T(\Delta^{p_k}_{j-1}), C_T(\Delta^{p_k}_j)]$ for some $j\in [I(p_k)]$. 
Notice that we are only interested in the case that $p_k$ is under-selected. 
In particular, if this is indeed the case, $\vp(t) = \vp_l$ for some $l\ge j$. 
(Otherwise, $\vp(t)$ is sufficiently selected based on the counter value $c_{t-1}(p_k)$.) 
Thus, we will suffer a regret of $\Delta_l^{p_k} \le \Delta_j^{p_k}$ (step~\eqref{ineq2}). 
As a result, for counter $c_t(p_k)\in (C_T(\Delta^{p_k}_{j-1}), C_T(\Delta^{p_k}_j)]$, we will suffer a total regret for those playing suboptimal meta arms that include under-selected discretized arms at most $(C_T(\Delta^{p_k}_j) - C_T(\Delta^{p_k}_{j-1}))\cdot \Delta_j^{p_k}$ in rounds that $c_t(p_k)$ is incremented (step~\eqref{ineq3}).
In what follows we establish the above analysis rigorously.
\begin{align*}
    &   \sum_{l\in[I(p_k)]} c_T^{l, \und}(p_k) \Delta_l^{p_k} \\
    =~& \sum_{t = \left \lceil K/\epsilon \right\rceil+1}^T\sum_{l\in [I(p_k)]} \mathbbm{1}\left(\vp(t) = \vp_l, c_t(p_k) > c_{t-1}(p_k),  c_{t-1}(p_k) \le C_T(\Delta_l^{p_k})  \right)\cdot \Delta_l^{p_k} \\
    =~& \sum_{t = \left \lceil K/\epsilon \right\rceil+1}^T\sum_{l\in [I(p_k)]} \sum_{j=1}^l\mathbbm{1}\left(\vp(t) = \vp_l, c_t(p_k) > c_{t-1}(p_k),  c_{t-1}(p_k) \in (C_T(\Delta^{p_k}_{j-1}), C_T(\Delta^{p_k}_j)]  \right)\cdot \Delta_l^{p_k} \\
    \labelrel\le{ineq2}~& \sum_{t = \left \lceil K/\epsilon \right\rceil+1}^T\sum_{l\in [I(p_k)]} \sum_{j=1}^l\mathbbm{1}\left(\vp(t) = \vp_l, c_t(p_k) > c_{t-1}(p_k),  c_{t-1}(p_k) \in (C_T(\Delta^{p_k}_{j-1}), C_T(\Delta^{p_k}_j)]  \right)\cdot \Delta_j^{p_k} \\ 
    \le~& \sum_{t = \left \lceil K/\epsilon \right\rceil+1}^T\sum_{l, j\in [I(p_k)]}\mathbbm{1}\left(\vp(t) = \vp_l, c_t(p_k) > c_{t-1}(p_k),  c_{t-1}(p_k) \in (C_T(\Delta^{p_k}_{j-1}), C_T(\Delta^{p_k}_j)]  \right)\cdot \Delta_j^{p_k} \\ 
    =~& \sum_{t = \left \lceil K/\epsilon \right\rceil+1}^T\sum_{j\in [I(p_k)]}\mathbbm{1}\left(\vp(t) \in \cS(p_k), c_t(p_k) > c_{t-1}(p_k),  c_{t-1}(p_k) \in (C_T(\Delta^{p_k}_{j-1}), C_T(\Delta^{p_k}_j)]  \right)\cdot \Delta_j^{p_k} \\ 
    \labelrel\le{ineq3}~& \sum_{j\in [I(p_k)]} (C_T(\Delta^{p_k}_j) - C_T(\Delta^{p_k}_{j-1}))\cdot \Delta_j^{p_k}.
\end{align*}
Now, we can compute the regret incurred by selecting the meta arm which includes under-selected discretized arms:
\begin{align}
\sum_{p_k}\sum_{l\in [I(p_k)]} c_T^{l, \und}(p_k) \Delta_l^{p_k}  & \leq \sum_{p_k}\sum_{j\in [I(p_k)]} (C_T(\Delta^{p_k}_j) - C_T(\Delta^{p_k}_{j-1}))\cdot \Delta_j^{p_k} \nonumber\\
& = \sum_{p_k} \bigg(C_T(\Delta_\text{min}^{p_k})\Delta_\text{min}^{p_k} + \sum_{j \in [I(p_k) - 1]}C_T(\Delta^{p_k}_j)\cdot(\Delta^{p_k}_j - \Delta_{j+1}^{p_k})\bigg) \nonumber\\
& \le \sum_{p_k}  \bigg(C_T(\Delta_\text{min}^{p_k})\Delta_\text{min}^{p_k} + \int_{\Delta_\text{min}^{p_k}}^{\Delta_\text{max}^{p_k}}C_t(x)dx\bigg) \nonumber\\
& = \sum_{p_k}\left(\frac{4K\ln T}{\Delta_\text{min}^{p_k}}  + 4K\ln T \left(\frac{1}{\Delta_\text{min}^{p_k}} - \frac{1}{\Delta_\text{max}^{p_k}} \right) \right). \label{und_bound}
\end{align}
Equipped with the above set of results, the bound of regret~\eqref{regret_3} follows by combing the bounds in~\eqref{Claim1} and~\eqref{Claim2}. 
\end{proof}

To achieve instance-independent regret bound, we need to deal with the case when the meta-arm gap $\Delta_{\min}^{p_k}$ is too small, leading the regret to approach infinite. 
Nevertheless, one can still show that when $\Delta_{\min}^{p_k} \le 1/\sqrt{T}$, 
the regret contributed by this scenario scales at most $\cO(\sqrt{T})$ at time horizon $T$.
\begin{lemma}\label{lemma: instance-independent regret}
Following the $\UCB$ designed in (\ref{ucb}), we have:
$\reg_\epsilon(T) \leq \cO\big(K\sqrt{T\ln T/\epsilon}\big)$.
\end{lemma}
\begin{proof}\label{Proof_lemma_instance-independent_regret}
Following the proof of Lemma~\ref{lemma: instance-dependent regret}, we only need to consider the meta arms that are played when they are under-sampled. 
We particularly need to deal with the situation when $\Delta_{\min}^{p_k}$ is too small. 
We measure the threshold for $\Delta_{\min}^{p_k}$ based on $c_T(p_k)$, i.e., the counter of disretized arm $p_k$ at time horizon $T$. 
Let $\{T(p_k), \forall p_k\}$  be a set of possible counter values at time horizon $T$. 
Our analysis will then be conditioned on the event that $\cE(p_k) = \{c_T(p_k) = T(p_k)\}$. 
By definition,
\begin{align}
    &   \Esymb\big[\sum_{l\in[I(p_k)]} c_T^{l, \und}(p_k) \cdot \Delta^{p_k}_l  \mid \cE(p_k) \big] \nonumber\\
    =~& \sum_{t = \left \lceil K/\epsilon \right\rceil+1}^T\sum_{l\in [I(p_k)]} \mathbbm{1}\left(\vp(t) = \vp_l, c_t(p_k) > c_{t-1}(p_k),  c_{t-1}(p_k) \le C_T(\Delta_l^{p_k})   \mid \cE(p_k)\right)\cdot \Delta_l^{p_k} \label{lem_3_6_eq}.
\end{align}
We define 
$\Delta^*(T(p_k)) := \left(\frac{4K\ln T}{T(p_k)}\right)^{1/2}$, 
i.e., $C_T(\Delta^*(T(p_k))) = T(p_k)$.
To achieve \emph{instance-independent} regret bound, we consider following two cases:\\
\textbf{Case 1:} $\Delta_{\min}^{p_k} > \Delta^*(T(p_k))$, we thus have
\begin{align}
     \Esymb\big[\sum_{l\in[I(p_k)]} c_T^{l, \und}(p_k) \cdot \Delta^{p_k}_l  \mid \cE(p_k) \big] \le \cO\left(\sqrt{4K\ln T\cdot T(p_k)}\right). \label{up_case_1}
\end{align}
\textbf{Case 2:} $\Delta_{\min}^{p_k} < \Delta^*(T(p_k))$. Let $l^* := \min\{l\in I(p_k): \Delta^{p_k}_l >\Delta^*(T(p_k)) \}$. 
Observe that we have $\Delta_{l^*}^{p_k}\le \Delta^*(T(p_k))$ and the counter $c(p_k)$ never go beyond $T(p_k)$, we thus have
\begin{align}
    \eqref{lem_3_6_eq} &\le (C_T(\Delta^*(T(p_k))) - C_T(\Delta_{l^*-1}^{p_k}))\cdot \Delta^*(T(p_k)) + \sum_{j\in[l^* - 1]}(C_T(\Delta^{p_k}_j) - C_T(\Delta^{p_k}_{j-1}))\cdot \Delta_j^{p_k} \nonumber \\
    &\le C_T(\Delta^*(T(p_k)))\cdot \Delta^*(T(p_k)) + \int_{\Delta^*(T(p_k))}^{\Delta_{\max}^{p_k}} C_T(x)dx \le \cO\left(\sqrt{K\ln T \cdot T(p_k)}\right).\label{up_case_2}
\end{align}
Thus, combining~\eqref{up_case_1} and \eqref{up_case_2}, we have 
\begin{align*}
    \Esymb\big[\sum_{p_k: \Delta_{\min}^{p_k}>0}\sum_{l\in[I(p_k)]} c_T^{l, \und}(p_k) \cdot \Delta^{p_k}_l  \mid \cE(p_k) \big] 
    & \le  \sum_{p_k: \Delta_{\min}^{p_k}>0} \cO(\sqrt{K \ln T \cdot T(p_k)})\\ 
    & \labelrel\le{ineq15} \cO(K\sqrt{ T\ln T/\epsilon}),
\end{align*}
where~\eqref{ineq15} is by Jesen's inequality and $\sum_{p_k} T(p_k) \le KT/\epsilon$. 
Put all pieces together, we have the instance-independent regret bound as stated in the lemma.
Observe that the final inequality does not depend on the event $\cE(p_k)$, we thus can drop this conditional expectation.
\end{proof}
With the above lemma in hand, picking $\epsilon = \Theta ((\ln T/T)^{1/3})$ will give us desired result in Theorem~\ref{theorem:key_theorem_vanilla}.
\footnote{Here the choice of $\epsilon$ absorbs Lipschitz constant of $r_k(\cdot)$.}

\begin{remark}
\label{rmk_new_Hoeffding}
When only one arm is activated according to $\vp(t)$, the 
Hoeffding's inequality is adapted as follows:
\begin{align*}
\Psymb\big(|\overbar{U}_t(\vp)- U(\vp)| \geq \delta\big) 
& \le \sum_k \Psymb\big(|p_k\bar{r}(p_k) -  p_kr(p_k)| \geq \delta/K\big) \\
& \le \sum_k 2\exp\left(-2\delta^2n_t(p_k)/K^2\right)
\le 2K\exp\left(-2\delta^2n_t(p_{\min}(\vp))/K^2\right).
\end{align*}
The below analysis carries over with accordingly changing $\delta =  \sqrt{\frac{K\ln t}{n_t(p_{\min}(\vp))}}$ to $\delta = \sqrt{\frac{K^2\ln (\sqrt{K} t)}{n_t(p_{\min}(\vp))}}$,
and the condition of $n_t(p_{\min}(\vp))$ in Lemma~\ref{lemma:key_lemma_vanilla} is changed to ${4K^2 \ln (\sqrt{K}t)}/{\Delta_{\vp}^2}$ to account for larger $\delta$. 
As a result, the instance-independent regret bound in Lemma~\ref{lemma: instance-independent regret} is changed to $\cO\left(K\sqrt{KT\ln(\sqrt{K}T)/\epsilon}\right)$.
Together with the discretization error, one can then optimize the choice of $\epsilon$ to 
get $\tilde{\cO}(K^{4/3}T^{2/3})$ regret bound.
\end{remark}

\subsubsection{Regret Bound Comparison with~\cite{chen2016combinatorial}}\label{cf_regret}
In the work~\cite{chen2016combinatorial}, the authors study the setting when pulling the meta arm, each base arm in (or possibly other base arm) this meta arm will be triggered and played as a result. 
Back to our setting, this is saying that when pulling a meta arm $\vp = (p_1, \ldots, p_K)$, each base arm $k$ will be triggered with its corresponding probability (discretized arm) $p_k$. 
The authors in~\citep{chen2016combinatorial} discuss a general setting which allows complex reward structure where only requires two mild conditions.
In particular, one of the condition they need for expected reward of playing a meta arm is the bounded smoothness (cf., Definition 1 in~\cite{chen2016combinatorial}.). 
In the Theorem 2 of~\cite{chen2016combinatorial}, the authors give results when the function used to characterize bounded smoothness is $f(x) = \gamma \cdot x^\omega$ for some $\gamma > 0$ and $\omega \in (0, 1]$. 
In more detail, they achieve a regret bound $\cO\left(\frac{2\gamma}{2 - \omega}\left(\frac{12|\cM|\ln T}{p^*}\right)^{\omega/2}\cdot T^{1 - \omega/2} + |\cM|\cdot \Delta_{\max}\right)$ 
where $p^* \in (0, 1)$ is the minimum triggering probability across all base arms and $\Delta_{\max}$ is the largest badness of the suboptimal meta arm in discretized space. \footnote{For simplicity, the bound we present here omits a non-significant term.}
Adapt to our setting, by inspection, we have $\gamma = L^*, \omega = 1$, $p^* = \epsilon$, $|\cM| = \Theta(K/\epsilon)$, and $\Delta_{\max} = \Theta(KL^*)$. 
Substituting these values to the above bound, ignoring constant factors and combining with the discretization error, we have 
\begin{align*}
    \cO\left(\left(\frac{K\ln T}{\epsilon^2}\right)^{1/2}\cdot T^{1/2} + K^2/\epsilon\right) + \cO(TK\epsilon).
\end{align*}
Picking $\epsilon = \Theta(\ln T/(KT))^{1/4}$ will give us result.

\section{Proof of Theorem~\ref{theorem: key_theorem_neq_1} for History-dependent Bandits}
\label{appendix: proofs_history_dependent_bandit}

In this section, we provide the analysis of Theorem~\ref{theorem: key_theorem_neq_1}.
The analysis follows a similar structure to the one used in the proof of the regret bound in Theorem~\ref{theorem:key_theorem_vanilla}.
However, due to the existence of historical bias, we need to perform a careful computation when handling the high-probability bounds.
Specifically, we need to prove that, after deploying $\vp$ consecutively for moderate long rounds (tuning $s_a$), the approximation error $\big|U(\vp) - \overbar{U}_m^\est(\vp)\big|$ is small enough.
The analysis is provided below.


\paragraph{\textbf{Step 1: Bounding the small error of $\big|U(\vp) - \overbar{U}_m^\est(\vp)\big|$ with high-probability.}}
Our first step is to ensure the empirical mean reward estimation we obtain from the information we collected in all the estimation stages will approximate well the true mean of meta arm we want to deploy.

To return a high-probability error bound, we first bound the approximation error incurred due to the dependency of history of arm selection (``historical bias").
This is summarized below.
\begin{lemma} \label{lemma: approx_err_neq_1}
Keeping deploying $\vp = \{p_1, \ldots, p_K\}$ in the approaching stage with $s_a$ rounds, and collect all reward feedback in the following estimation stage for the empirical estimation of rewards generated by $\vp$, one can bound the approximation error as follows:
\begin{align*}
\Esymb \big[\big|\overbar{U}_m^\est(\vp) -  \overbar{U}(\vp)\big| \big]  \leq K\gamma^{s_a}(L^*+1),
\end{align*} 
where $ \overbar{U}(\vp)$ denote the empirical mean of rewards if the instantaneous reward is truly sampled from mean reward function according to $\vp$.
\end{lemma}
\begin{proof} \label{Proof_lemma_approx_err_neq_1}
The proof of this lemma is mainly built on analyzing the convergence of $\vp^{(\gamma)}$ via pulling the base arms with the same probability consistently. 
For the ease of presentation, let us suppose $t = mL$ and let $t^\est_m := \frac{t}{L}(L-s_a) = m(L-s_a)$ be the total number of estimation rounds in the first $m$ phases.
Thus, at the end of the approaching stage, we have
\begin{align*}
\hat{p}^{(\gamma)}_k(t+s_a) = \frac{p_k(t+s_a)\gamma^{0} + \ldots+ p_k(t+1)\gamma^{s_a-1} + (1 + \gamma +\ldots + \gamma^{t-1})\gamma^{s_a}\hat{p}^{(\gamma)}_k(t)}{1 + \gamma +\ldots + \gamma^{t + s_a-1}},
\end{align*}
where $\hat{p}^{(\gamma)}_k(t) = \frac{p_k(t)\gamma^{0} + \ldots + p_k(1)\gamma^{t-1}}{1 + \gamma +\ldots + \gamma^{t -1}}$.
Recall that during the approaching stage, we consistently pull arm $k$ with the same probability $p_k$.
Thus, the approximation error of $\hat{p}^{(\gamma)}_k(t+s_a)$ w.r.t. $p_k$ can be computed as:
\begin{align*}
\big|\hat{p}^{(\gamma)}_k(t+s_a)- p_k\big| =  \bigg|\frac{p_k(1-\gamma^{s_a}) + \hat{p}^{(\gamma)}_k(t)\gamma^{s_a}(1-\gamma^{t})}{1-\gamma^{t+s_a}} - p_k \bigg| \leq \frac{\gamma^{s_a}(1-\gamma^{t})}{1-\gamma^{t+s_a}} < \gamma^{s_a}.
\end{align*}
Recall that $U(\vp) = \sum_{p_k \in \vp} p_k\rewfn_k(p_k)$.
In the estimation stage, we approximate all the realized utility as the utility generated by the meta arm $\vp$.
However, note that we actually cannot compute the empirical value of $\overbar{U}(\vp)$, instead, we use $\overbar{U}_m^\est(\vp(t+s_a))$ of each phase as an approximation of $\overbar{U}(\vp)$, i.e., we approximate all $\vp^{(\gamma)}(t+s), \forall s \in (s_a, L]$ as $\vp(t+s_a)$ and use $\vp(t+s_a)$ as the approximation of $\vp$.
Recall that for any $s\in (s_a, L]$, we have:
\begin{align*}
\big|\hat{p}^{(\gamma)}_k(t+s) - p_k\big| & = \bigg|\frac{\gamma^{s}(1-\gamma^t)(\hat{p}^{(\gamma)}_k(t) - p_k)}{1-\gamma^{t+s}}\bigg| \leq \frac{\gamma^{s}(1-\gamma^{t})}{1-\gamma^{t+s}} < \frac{\gamma^{s_a}(1-\gamma^{t})}{1-\gamma^{t+s_a}} < \gamma^{s_a}.
\end{align*}
Thus, the approximation error on the empirical estimation can be computed as follows:
\begin{align*}
\Esymb \left[\big|\overbar{U}_m^\est(\vp(t+s_a)) -  \overbar{U}(\vp)\big| \right] & = \Esymb \bigg[\bigg|\sum_{p_k^{(\gamma)} \in \vp(t+s_a)} p_k^{(\gamma)}\bar{\rewfn}^\est_{t+s_a}(p_k^{(\gamma)})  - \sum_{p_k \in \vp} p_k\bar{\rewfn}^\est_{t+s_a}(p_k)\bigg|\bigg] \\
& =\bigg|\sum p_k^{(\gamma)}\Esymb\left[\bar{\rewfn}^\est_{t+s_a}(p_k^{(\gamma)})\right] - \sum p_k\Esymb\left[\bar{\rewfn}^\est_{t+s_a}(p_k)\right]\bigg|\\
& = \bigg|\sum p_k^{(\gamma)}\rewfn_k(p_k^{(\gamma)}) - \sum p_k\rewfn_k(p_k)  \bigg|\\
& = \bigg|\sum \left(p_k^{(\gamma)}\left(\rewfn_k(p_k^{(\gamma)}) -\rewfn_k(p_k)\right) + \rewfn_k(p_k)(p_k^{(\gamma)}- p_k)\right) \bigg|\\
& \leq \sum \left|\gamma^{s_a} L_kp_k^{(\gamma)} + \rewfn_k(p_k)\gamma^{s_a}\right| \leq K\gamma^{s_a}(L^*+1).
\end{align*}
\end{proof}

With the approximation error at hand, we can then bound the error of $\big|U(\vp) - \overbar{U}_m^\est(\vp)\big|$ with high probability:
\begin{lemma}\label{lemma:high_prob_error_neq_1}
With probability at least $1-\frac{6}{\big(L\rho m\big)^2}$, we have
\begin{align*}
\big|U(\vp) - \overbar{U}_m^\est(\vp)\big| \leq \err + 3\sqrt{\frac{K\ln \big(L\rho m\big)}{n_m^\est(p_{\min}(\vp))}},
\end{align*}
where $p_{\min}(\vp) = \argmin_{p_k \in \vp} n^\est_m(p_k)$.
\end{lemma}
\begin{proof}\label{Proof_lemma_high_prob_error_neq_1}
We first decompose $\big|U(\vp) - \overbar{U}_m^\est(\vp_{e}^{(\gamma)})\big|$ as $\big|U(\vp) - \overbar{U}(\vp)  \big|  + \big|\overbar{U}(\vp) - \overbar{U}_m^\est(\vp) \big|$ and then apply union bound.
\begin{align*}
& \Psymb\left( \big|U(\vp) - \overbar{U}_m^\est(\vp(t+s_a))\big| \geq \delta\right) \\
\leq~&  \Psymb\left(\big|U(\vp) - \overbar{U}(\vp)  \big|  + \big|\overbar{U}(\vp) - \overbar{U}_m^\est(\vp(t+s_a))  \big| \geq \delta\right) & \tag*{By triangle inequality}\\
=~& \Psymb\bigg(\big|U(\vp) - \overbar{U}(\vp)  \big|  + \big|\overbar{U}_m^\est(\vp(t+s_a)) - \Esymb[\overbar{U}_m^\est(\vp(t+s_a)) ] - \\
& \quad (\overbar{U}(\vp) - \Esymb[\overbar{U}(\vp)] ) + \Esymb[\overbar{U}(\vp)] - \Esymb[\overbar{U}_m^\est(\vp(t+s_a)) ] \big|  \geq \delta\bigg)\\
\leq~&    \Psymb\bigg(2\big|U(\vp) - \overbar{U}(\vp)  \big|  + \big|\overbar{U}_m^\est(\vp(t+s_a)) - \Esymb[\overbar{U}_m^\est(\vp(t+s_a))\big| \geq \delta - \err\bigg)\\
\labelrel\leq{ineq7}~&  3\Psymb\bigg(|U(\vp) - \overbar{U}(\vp)  \big|  \geq \frac{\delta - \err}{3}\bigg) \leq  6\exp\bigg(-\frac{2n^\est_m(p_{\min}(\vp))(\delta - \err)^2}{9K} \bigg),
\end{align*}
where in step~\eqref{ineq7}, we use the Hoeffding's Inequality on Weighted Sums and Lemma~\ref{lemma: approx_err_neq_1}. 
\end{proof}


\paragraph{\textbf{Step 2: Bounding the probability on deploying suboptimal meta arm.}}
Till now, with the help of the above high probability bound on the empirical reward estimation, the history-dependent reward bandit setting is largely reduced to an action-dependent one with a certain approximation error.  
Then, similar to our argument on upper bound of action-dependent bandits, we have the following specific Lemma for history-dependent bandits: 
\begin{lemma}\label{lemma:key_lemma_neq_1}
At the end of each phase, for a suboptimal meta arm $\vp$, if it satisfies $n_m^\est(p_{\min}(\vp)) \geq \frac{9K\ln \big(L\rho m\big)}{\big(\Delta_{\vp}/2 - \err\big)^2}$, 
then with the probability at least $1- \frac{12}{\big(L\rho m\big)^2}$, we have $\textup{\UCB}_{m}(\vp) < \textup{\UCB}_{m}(\vp^*)$, i.e.,
\begin{align*}
\Psymb\bigg(\vp(m+1) = \vp|n_m^\est(p_{\min}(\vp)) \geq \frac{9K\ln \big(L\rho m\big)}{\big(\frac{\Delta_{\vp}}{2} - \err\big)^2}\bigg) \leq \frac{12}{\big(L\rho m\big)^2}.
\end{align*}
\end{lemma}

\begin{proof}\label{Proof_lemma_key_lemma_neq_1}
To prove the above lemma, we construct two high-probability events. 
\texttt{Event 1} corresponds to that the $\UCB$ index of each meta arm concentrates on the true mean utility of $\vp$;
\texttt{Event 2} corresponds to that the empirical mean utility of each approximated meta arm $\vp^{(\gamma)}$ concentrates on the true mean utility of $\vp$.
The probability of \texttt{Event 1} or \texttt{Event 2} not holding is at most $4/t^2$.
By the definition of the constructed $\UCB$, we'll have
\begin{align*}
\UCB_m(\vp) & =  \overbar{U}_m^\est(\vp(t+s_a)) + \err + 3\sqrt{\frac{K\ln \left(L\rho m\right)}{n_m^\est(p_{\min}(\vp))}} \labelrel\leq{ineq8} \overbar{U}_m^\est(\vp(t+s_a)) + \Delta_{\vp}/2 \\
& \labelrel<{ineq9} \left(U(\vp) + \Delta_{\vp}/2\right) + \Delta_{\vp}/2  &\tag*{By \texttt{Event 1}}\\
& = U(\vp_\epsilon^*)  \labelrel<{ineq10} \UCB_m(\vp_\epsilon^*), &\tag*{By \texttt{Event 2}}
\end{align*}
where the first inequality~\eqref{ineq8} is due to  $n_m^\est(p_{\min}(\vp)) \geq \frac{9K\ln \left(L\rho m\right)}{\left(\Delta_{\vp}/2 - \err\right)^2}$,
and the probability of step~\eqref{ineq9} or~\eqref{ineq10} not holding is at most $12/{\left(L\rho m\right)^2}$.
\end{proof}
The above lemma implies that we will stop deploying suboptimal meta arm $\vp$ and further prevent it from incurring regret as we gather more information about it such that $\UCB_{m}(\vp) < \UCB_{m}(\vp_\epsilon^*)$.

\paragraph{\textbf{Step 3: Bounding the $\Esymb[n^\est_m(p_{\min}(\vp))]$.}}
The results we obtain in Step 2 implies following guarantee:
\begin{lemma} \label{lemma: sufficient_sample_reduction}
For each suboptimal meta arm $\vp \neq \vp^*$, we have following: 
\begin{align*}
& \Esymb[n^\est_m(p_{\min}(\vp))] \leq \frac{9K\ln \big(L\rho m\big)}{\big(\Delta_{\vp}/2 - \err\big)^2} + \frac{2\pi^2}{L-s_a}.
\end{align*}
\end{lemma}
\begin{proof} \label{Proof_lemma_sufficient_sample_reduction}
For notation simplicity, suppose $t = mL$.
For each suboptimal arm $\vp \neq \vp_\epsilon^*$, and suppose there exists $p_{\min}(\vp) \notin \vp_\epsilon^*$ such that $p_{\min}(\vp) = \argmin_{p_k \in \vp} n^\est_t(p_k)$, then
\begin{align*}
& \Esymb[n^\est_t(p_{\min}(\vp))] \\
=~& (L-s_a)\Esymb \left[\sum_{i = 1}^{m} \mathbbm{1}\left(\vp(i) = \vp, \vp\in \cS(p_{\min}(\vp))\right)\right] \\
=~& (L-s_a) \Esymb \left[\sum_{i = 1}^{m} \mathbbm{1}\left(\vp(i) = \vp, \vp\in \cS(p_{\min}(\vp)); n_i^\est(p_{\min}(\vp)) < \frac{9K\ln \left(i(L-s_a)\right)}{\left(\Delta_{\vp}/2 - \err\right)^2} \right)\right] + \\
&  (L-s_a)\Esymb \left[\sum_{i = 1}^{m} \mathbbm{1}\left(\vp(i) = \vp, \vp\in \cS(p_{\min}(\vp)); n_i^\est(p_{\min}(\vp)) \geq \frac{9K\ln \left(i(L-s_a)\right)}{\left(\Delta_{\vp}/2 - \err\right)^2}\right)\right] \\
\labelrel\leq{ineq14}~& \frac{9K\ln \left(t^\est_{m}\right)}{\left(\Delta_{\vp}/2 - \err\right)^2} +  (L-s_a)\Esymb \left[\sum_{i = 1}^{m} \mathbbm{1}\left(\vp(i) = \vp, \vp\in \cS(p_{\min}(\vp)); n_i^\est(p_{\min}(\vp)) \geq \frac{9K\ln \left(i(L-s_a)\right)}{\left(\Delta_{\vp}/2 - \err\right)^2} \right)\right] \\
=~& \frac{9K\ln \left(t^\est_{m}\right)}{\left(\Delta_{\vp}/2 - \err\right)^2} +  (L-s_a) \sum_{i = 1}^{m} \Psymb\left(\vp(i) = \vp, \vp\in \cS(p_{\min}(\vp))\bigg| n_i^\est(p_{\min}(\vp)) \geq \frac{9K\ln \left(i(L-s_a)\right)}{\left(\Delta_{\vp}/2 - \err\right)^2} \right) \cdot\\
& \qquad \qquad\qquad\qquad\qquad\quad\quad  \Psymb \left(n_i^\est(p_{\min}(\vp)) \geq \frac{9K\ln \left(i(L-s_a)\right)}{\left(\Delta_{\vp}/2 - \err\right)^2}\right)\\
\leq~&  \frac{9K\ln \left(t^\est_{m}\right)}{\left(\Delta_{\vp}/2 - \err\right)^2} + (L-s_a)\sum_{i=1}^{m} \frac{12}{\left(i(L-s_a)\right)^2} \leq \frac{9K\ln \left(t^\est_{m}\right)}{\left(\Delta_{\vp}/2 - \err\right)^2} + \frac{2\pi^2}{L-s_a}.
\end{align*}
In step~\eqref{ineq14}, suppose for contradiction that the indicator $\mathbbm{1}\left(\vp(i) = \vp, \vp\in \cS(p_{\min}(\vp)); n_i^\est(p_{\min}(\vp)) < S\right)$ takes value of $1$ at more than $S-1$ time steps, where $S = \frac{9K\ln \left(i(S-s_a)\right)}{\left(\Delta_{\vp}/2 - \err\right)^2}$. 
Let $\tau$ be the phase at which this indicator is $1$ for the $(S-1)$-th phase.
Then the number of pulls of all meta arms in $\cS(p_{\min}(\vp))$ is at least $L$ times until time $\tau$ (including the initial pull), and for all $i > \tau$, $n_i(p_{\min}(\vp)) \geq S$ which implies $n_i^\est(p_{\min}(\vp)) \geq \frac{9K\ln \left(i(S-s_a)\right)}{\left(\Delta_{\vp}/2 - \err\right)^2}$.
Thus, the indicator cannot be $1$ for any $i\geq \tau$, contradicting the assumption that the indicator takes value of $1$ more than $S$ times.
This bounds $1 + \Esymb \left[\sum_{i = 1}^m \mathbbm{1}\left(\vp(i) = \vp, \vp\in \cS(p_{\min}(\vp)); n_i^\est(p_{\min}(\vp)) < S\right)\right]$ by $S$.
\end{proof}


\paragraph{\textbf{Wrapping up: Proof of Theorem~\ref{theorem: key_theorem_neq_1}.}} 
Following the similar analysis in Section 3, we can also get an instance-dependent regret bound for history-dependent bandits:
\begin{lemma}\label{lemma: instance-dependent regret_hist}
Following the $\UCB$ designed in Algorithm~\ref{algo:history-dependent}, we have following instance-dependent regret on discretized arm space for history-dependent bandits:
\begin{align*}
\reg_\epsilon(T) \leq \cO\bigg(\frac{K\Delta_{\max}}{L\epsilon\rho^2}\bigg)  + 
\sum_{p_k}   \bigg(\frac{9K\ln \left(T\rho\right)}{\rho} \bigg(\frac{\Delta_{\min}^{p_k}}{\left(\Delta_{\min}^{p_k}/2 - \err\right)^2} + \frac{2}{\Delta_{\min}^{p_k}/2 - \err} \bigg)\bigg).
\end{align*}
\end{lemma}
\ignore{
    \begin{sproof}
    The proof of the above lemma follows from similar outline as in the proof of Lemma~\ref{lemma: instance-dependent regret}. 
    We emphasize differences here. 
    We define a counter $c^\est(p_k)$ for each discretized arm $p_k$ and the value of $c^\est(p_k)$ at phase $m$ is denoted by $c_m^\est(p_k)$. 
    But different from the proof of action-dependent bandits, we update the counter $c^\est(p_k)$ only when we start a new phase.
    In particular, for a phase $m \ge 1$, let $\vp(m)$ be the meta arm selected in the phase $m$ by the algorithm. 
    Let $p_k = \argmin_{p_k\in\vp(m)} c_m^\est(p_k)$, we then increment $c_m^\est(p_k)$ by one, i.e., $c_m^\est(p_k) = c_{m-1}^\est(p_k) + 1$. 
    In other words,  we find the discretized arm $p_k$ with the smallest counter in $\vp(m)$ and increment its counter. 
    Note that the initialization gives $\sum_{p_k}c_0^\est(p_k) = \left \lceil K/\epsilon \right \rceil$. 
    Then for any $p_k = p_{\min}(\vp)$, we have $n_m(p_k) = L\rho\cdot c_m(p_k)$.
    We also define $C_M^\est(\Delta) := \frac{9K\ln \left(ML\rho\right)}{L\rho\left(\Delta/2 - \err\right)^2}$,
    the number of selection that is considered sufficient for a meta arm with reward $\Delta$ away from the optimal strategy $\vp_\epsilon^*$ with respect to phase horizon $M$.
    With the above notations, we define following two situations for the counter of each discretized arm: 
    \begin{align*}
        c_M^{\est, l, \suf}(p_k) &:= \sum_{m =  1}^M \mathbbm{1}\left(\vp(m) = \vp_l, c_m^\est(p_k) > c_{m-1}^\est(p_k) >  C_M^\est(\Delta_l^{p_k}) \right) \\
        c_M^{\est, l, \und}(p_k) &:= \sum_{m= 1}^M \mathbbm{1}\left(\vp(m) = \vp_l, c_m^\est(p_k) > c_{m-1}^\est(p_k),  c_{m-1}^\est(p_k) \le  C_M^\est(\Delta_l^{p_k}) \right).
    \end{align*}
    Similarly, we have $c_M^\est(p_k) = 1 + \sum_{l\in I(p_k)}\big(c_M^{\est, l, \suf}(p_k) + c_M^{\est, l, \und}(p_k)\big)$.
    With these notations, we can write~\eqref{regret_hist} as follows:
    \begin{align}
        \reg_\epsilon(T) \le \Esymb\bigg[\sum_{p_k} \bigg(\Delta^{p_k}_{\max} + L\cdot\sum_{l\in [I(p_k)]}\left(c_M^{\est, l, \suf}(p_k) + c_M^{\est, l, \und}(p_k)\right)\cdot\Delta_l^{p_k}\bigg)\bigg]. \nonumber 
    \end{align}
    The proof of this lemma will complete after establishing following two claims:
    \begin{align*}
    \text{Claim 3:}  ~~ &\Esymb\bigg[L\sum_{p_k}\sum_{l\in [I(p_k)]} c_M^{\est, l, \suf}(p_k) \bigg] \le  \cO\left(\frac{K}{L\epsilon\rho^2}\right). \\
    \text{Claim 4:}  ~~ & \Esymb\bigg[ L \sum_{p_k}\sum_{l\in [I(p_k)]} c_M^{\est, l, \und}(p_k) \Delta_l^{p_k} \bigg] \leq   \sum_{p_k} \bigg(\frac{9K\ln \left(ML\rho\right)}{\rho} \bigg(\frac{\Delta_{\min}^{p_k}}{\left(\Delta_{\min}^{p_k}/2 - \err\right)^2} + \frac{2}{\Delta_{\min}^{p_k}/2 - \err} \bigg)\bigg). 
    \end{align*}
    The proof of the above two claims is provided in Appendix~\ref{lemma: instance-dependent regret_hist}. 

\end{sproof}
}
\begin{proof}\label{Proof_lemma_instance-dependent_regret_hist}
For notation simplicity, we include all initialization rounds to phase $0$ and suppose the time horizon $T = ML$.
Note that by definitions, we can compute the regret $\reg_{\epsilon} (T)$ as follows:
\begin{align}
\reg_\epsilon(T) = \sum_{\vp \in \cP_\epsilon} \Esymb[N_T(\vp)] \Delta_{\vp} \leq \sum_{p_k} \sum_{\vp_l\in\cS(p_k)}  \Esymb[N_T(\vp_l)] \Delta_l^{p_k}. \label{regret_hist}
\end{align}
where $N_t(\vp) = K + L \sum_{m = 1}^M \mathbbm{1}\left(\vp(m) = \vp\right)$, where $K$ here accounts for the initialization. 
Follow the same analysis in action-dependent bandits, we can also define a counter $c^\est(p_k)$ for each discretized arm $p_k$ and the value of $c^\est(p_k)$ at phase $m$ is denoted by $c_m^\est(p_k)$. 
But different from the action-dependent bandit setting, we update the counter $c^\est(p_k)$ only when we start a new phase.
In particular, for a phase $m \ge 1$, let $\vp(m)$ be the meta arm selected in the phase $m$ by the algorithm. 
Let $p_k = \argmin_{p_k\in\vp(m)} c_m^\est(p_k)$. 
We increment $c_m^\est(p_k)$ by one, i.e., $c_m^\est(p_k) = c_{m-1}^\est(p_k) + 1$. 
In other words,  we find the discretized arm $p_k$ with the smallest counter in $\vp(m)$ and increment its counter. 
If such $p_k$ is not unique, we pick an arbitrary discretized arm with the smallest counter. 
Note that the initialization gives $\sum_{p_k}c_0^\est(p_k) = \left \lceil K/\epsilon \right \rceil$. 
It is easy to see that for any $p_k = p_{\min}(\vp)$, we have $n_m(p_k) = L\rho\cdot c_m(p_k)$.

Like in action-dependent bandits, we also define 
$C_M^\est(\Delta) := \frac{9K\ln \left(ML\rho\right)}{L\rho\left(\Delta/2 - \err\right)^2}$,
the number of selection that is considered sufficient for a meta arm with reward $\Delta$ away from the optimal strategy $\vp_\epsilon^*$ with respect to phase horizon $M$.
With the above notations, we define following two situations for the counter of each discretized arm: 
\begin{align}
    c_M^{\est, l, \suf}(p_k) &:= \sum_{m =  1}^M \mathbbm{1}\left(\vp(m) = \vp_l, c_m^\est(p_k) > c_{m-1}^\est(p_k) >  C_M^\est(\Delta_l^{p_k}) \right) \\
    c_M^{\est, l, \und}(p_k) &:= \sum_{m= 1}^M \mathbbm{1}\left(\vp(m) = \vp_l, c_m^\est(p_k) > c_{m-1}^\est(p_k),  c_{m-1}^\est(p_k) \le  C_M^\est(\Delta_l^{p_k}) \right).
\end{align}
Clearly, we have $c_M^\est(p_k) = 1 + \sum_{l\in I(p_k)}\big(c_M^{\est, l, \suf}(p_k) + c_M^{\est, l, \und}(p_k)\big)$.
With these notations, we can write~\eqref{regret_hist} as follows:
\begin{align}
    \reg_\epsilon(T) \le \Esymb\left[\sum_{p_k} \left(\Delta^{p_k}_{\max} + L\cdot\sum_{l\in [I(p_k)]}\left(c_M^{\est, l, \suf}(p_k) + c_M^{\est, l, \und}(p_k)\right)\cdot\Delta_l^{p_k}\right)\right]. \label{regret_hist_3}
\end{align}
We now first show that for any $m \ge 1$, we have following upper bound over counters of sufficiently selected discretized arms:
\begin{align}
    \Esymb\bigg[L\cdot\sum_{p_k}\sum_{l\in [I(p_k)]} c_M^{l, \suf}(p_k) \bigg] \le  \cO\left(\frac{K}{L\epsilon\rho^2}\right). \label{suf_bound_hist}
\end{align}
To see this, by definition of $c_M^{\est, l, \suf}(p_k)$, it reduces to show that for any $M \ge m > 1$,
\begin{align*}
      & \Esymb\bigg[L\cdot\sum_{p_k}\sum_{l\in [I(p_k)]}\mathbbm{1}\left(\vp(m) = \vp_l, c_m^\est(p_k) > c_{m-1}^\est(p_k) >  C_M^\est(\Delta_l^{p_k})\right) \bigg] \\
     =~&  L\cdot \sum_{p_k}\sum_{l\in [I(p_k)]} \Psymb\bigg(\vp(m) = \vp_l, p_k = p_{\min}(\vp_l); \forall p  \in \vp_l, L\rho \cdot c_{m-1}^\est(p) >  \frac{9K\ln \left(ML\rho\right)}{\big(\Delta_l^{p_k}/2 - \err\big)^2}\bigg)\\
     \labelrel\le{ineq11}~& \left \lceil 12LK/\epsilon \right \rceil\cdot (ML\rho)^{-2},
\end{align*}
where the last step~\eqref{ineq11} is due to Lemma~\ref{lemma:key_lemma_neq_1}, thus~\eqref{suf_bound_hist} follows from a simple series bound.

We now proceed to analyze the discretized arms that are not sufficiently included in the meta arm chosen by the algorithm. 
For any under-selected discretized arm $p_k$, its counter $c^\est(p_k)$ will increase from $1$ to $ C_M^\est(\Delta_{\min}^{p_k})$. 
To simplify the notation, we set $C_M^\est(\Delta_0^{p_k}) = 0$.
Suppose that at phase $m \ge 1$, $c^\est(p_k)$ is incremented, and $c_{m-1}^\est(p_k) \in ( C_M^\est(\Delta^{p_k}_{j-1}), C_M^\est(\Delta^{p_k}_j)]$ for some $j\in [I(p_k)]$. 
Notice that we are only interested in the case that $p_k$ is under-selected. 
In particular, if this is indeed the case, $\vp(m) = \vp_l$ for some $l\ge j$. 
(Otherwise, $\vp(m)$ is sufficiently selected based on the counter value $c_{m-1}^\est(p_k)$.) 
Thus, we will suffer a regret of $\Delta_l^{p_k} \le \Delta_j^{p_k}$ (step~\eqref{ineq12}). 
As a result, for counter $c_m^\est(p_k)\in (C_M^\est(\Delta^{p_k}_{j-1}), C_M^\est(\Delta^{p_k}_j)/L]$, we will suffer a total regret for those playing suboptimal meta arms that include under-selected discretized arms at most $(C_M^\est(\Delta^{p_k}_j) - C_M^\est(\Delta^{p_k}_{j-1}))\cdot \Delta_j^{p_k}$ in rounds that $c_m^\est(p_k)$ is incremented (step~\eqref{ineq13}).
In what follows we establish the above analysis rigorously.
\begin{align*}
    &   \sum_{l\in[I(p_k)]} c_M^{\est, l, \und}(p_k) \Delta_l^{p_k} \\
    =~& \sum_{m = 1}^M\sum_{l\in [I(p_k)]} \mathbbm{1}\left(\vp(m) = \vp_l, c_m^\est(p_k) > c_{m-1}^\est(p_k),  c_{m-1}^\est(p_k) \le  C_M^\est(\Delta_l^{p_k})  \right)\cdot \Delta_l^{p_k} \\
    =~& \sum_{m=1}^M\sum_{l\in [I(p_k)]} \sum_{j=1}^l\mathbbm{1}\left(\vp(m) = \vp_l, c_m^\est(p_k) > c_{m-1}^\est(p_k),  c_{m-1}^\est(p_k) \in (C_M^\est(\Delta^{p_k}_{j-1}), C_M^\est(\Delta^{p_k}_j)]  \right)\cdot \Delta_l^{p_k} \\
    \labelrel\le{ineq12}~& \sum_{m=1}^M\sum_{l\in [I(p_k)]} \sum_{j=1}^l\mathbbm{1}\left(\vp(m) = \vp_l, c_m^\est(p_k) > c_{m-1}^\est(p_k),  c_{m-1}^\est(p_k) \in (C_M^\est(\Delta^{p_k}_{j-1}), C_M^\est(\Delta^{p_k}_j)]  \right)\cdot \Delta_j^{p_k} \\ 
    \le~& \sum_{m=1}^M\sum_{l, j\in [I(p_k)]}\mathbbm{1}\left(\vp(m) = \vp_l, c_m^\est(p_k) > c_{m-1}^\est(p_k),  c_{m-1}^\est(p_k) \in (C_M^\est(\Delta^{p_k}_{j-1}), C_M^\est(\Delta^{p_k}_j)]  \right)\cdot \Delta_j^{p_k} \\ 
    =~& \sum_{m=1}^M\sum_{j\in [I(p_k)]}\mathbbm{1}\left(\vp(m) \in \cS(p_k), c_m^\est(p_k) > c_{m-1}^\est(p_k),  c_{m-1}^\est(p_k) \in (C_M^\est(\Delta^{p_k}_{j-1}), C_M^\est(\Delta^{p_k}_j)]  \right)\cdot \Delta_j^{p_k} \\ 
    \labelrel\le{ineq13}~& \sum_{j\in [I(p_k)]} (C_M^\est(\Delta^{p_k}_j) - C_M^\est(\Delta^{p_k}_{j-1}))\cdot \Delta_j^{p_k}.
\end{align*}
Now, we can compute the regret incurred by selecting the meta arm which includes under-selected discretized arms:
\begin{align*}
& L \cdot \sum_{p_k}\sum_{l\in [I(p_k)]} c_M^{\est, l, \und}(p_k) \cdot\Delta_l^{p_k}  \\
\leq~ & L \cdot \sum_{p_k}\sum_{j\in [I(p_k)]} (C_M^\est(\Delta^{p_k}_j) - C_M^\est(\Delta^{p_k}_{j-1}))\cdot \Delta_j^{p_k} \nonumber\\
=~  & L \cdot \sum_{p_k} \bigg(C_M^\est(\Delta_{\min}^{p_k})\Delta_{\min}^{p_k} + \sum_{j \in [I(p_k) - 1]}C_M^\est(\Delta^{p_k}_j)\cdot(\Delta^{p_k}_j - \Delta_{j+1}^{p_k})\bigg) \nonumber\\
\le~&  L \cdot \sum_{p_k}  \bigg(C_M^\est(\Delta_{\min}^{p_k})\Delta_{\min}^{p_k} + \int_{\Delta_{\min}^{p_k}}^{\Delta_\text{max}^{p_k}}C_M^\est(x)dx\bigg) \nonumber\\
=~&  \sum_{p_k}\left(\frac{9K\ln \left(ML\rho\right)}{\rho\left(\Delta_{\min}^{p_k}/2 - \err\right)^2}  \cdot \Delta_{\min}^{p_k} + 9K\ln \left(ML\rho\right)/\rho \cdot\int_{\Delta_{\min}^{p_k}}^{\Delta_\text{max}^{p_k}} \frac{1}{(x/2 - \err)^2}dx \right) \nonumber \\
=~&  \sum_{p_k} \left( \frac{9\Delta_{\min}^{p_k}K\ln \left(ML\rho\right)}{\rho\left(\Delta_{\min}^{p_k}/2 - \err\right)^2} +  \frac{9K\ln \left(ML\rho\right)}{\rho} \left(\frac{2}{\frac{\Delta_{\min}^{p_k}}{2} - \err} - \frac{2}{\Delta_\text{max}^{p_k}/2 - \err}\right)\right) \nonumber\\
\leq~&   \sum_{p_k}   \left(\frac{9K\ln \left(ML\rho\right)}{\rho} \left(\frac{\Delta_{\min}^{p_k}}{\left(\Delta_{\min}^{p_k}/2 - \err\right)^2} + \frac{2}{\Delta_{\min}^{p_k}/2 - \err} \right)\right). \nonumber
\end{align*}
Combing the bound established in~\eqref{suf_bound_hist} will complete the proof. 
\end{proof}

The instance-independent regret on discretized arm space is summarized in following lemma:
\begin{lemma}\label{lemma: instance-independent regret_hist}
Following the $\UCB$ designed in Algorithm~\ref{algo:history-dependent}, the instance-independent regret is given as
$\reg_\epsilon(T) \leq \cO\left(K\cdot \sqrt{{T\ln (T\rho)}/{(\rho\epsilon)}} + K/(L\epsilon\rho^2)\right)$.
\end{lemma}
\begin{proof}\label{Proof_lemma_instance-independent_regret_hist}
Following the proof action-dependent bandits, we only need to consider the meta arms that are played when they are under-sampled. 
We particularly need to deal with the situation when $\Delta_{\min}^{p_k}$ is too small. 
We measure the threshold for $\Delta_{\min}^{p_k}$ based on $c_M^\est(p_k)$, i.e., the counter of disretized arm $p_k$ at phase horizon $M$. 
Let $\{M(p_k), \forall p_k\}$  be a set of possible counter values at time horizon $M$. 
Our analysis will then be conditioned on the event that $\cE(p_k) := \{c_M^\est(p_k) = M(p_k)\}$. 
By definition,
\begin{align}
    &   \Esymb\big[\sum_{l\in[I(p_k)]} c_M^{\est, l, \und}(p_k) \cdot \Delta^{p_k}_l  \mid \cE(p_k) \big] \nonumber\\
    =~& \sum_{m = 1}^M\sum_{l\in [I(p_k)]} \mathbbm{1}\left(\vp(m) = \vp_l, c_m^\est(p_k) > c_{m-1}^\est(p_k),  c_{m-1}^\est(p_k) \le  C_M^\est(\Delta_l^{p_k})  \mid \cE(p_k)\right)\cdot \Delta_l^{p_k} \label{lem_4_8_eq}.
\end{align}
We define $\Delta^*(M(p_k)) := 2\left(\frac{9K\ln (ML\rho)}{L\rho\cdot M(p_k)}\right)^{1/2} + 2\err$. 
thus we have $C_M^\est(\Delta^*(M(p_k))) =  M(p_k)$.
To achieve \emph{instance-independent} regret bound, we consider following two cases:\\
\textbf{Case 1:} $\Delta_{\min}^{p_k} > \Delta^*(M(p_k))$, clearly we have $\Delta_{\min}^{p_k}/2 > \err$. 
Thus,
\begin{align}
     &L \cdot\Esymb\big[\sum_{l\in[I(p_k)]} c_M^{\est, l, \und}(p_k) \cdot \Delta^{p_k}_l  \mid \cE(p_k) \big] \le \cO\left( \sqrt{\frac{K\ln (T\rho) \cdot LM(p_k)}{\rho}}\right). \label{up_case_1_hist}
\end{align}
\textbf{Case 2:} $\Delta_{\min}^{p_k} < \Delta^*(M(p_k))$. Let $l^* := \min\{l\in I(p_k): \Delta^{p_k}_l >\Delta^*(M(p_k)) \}$. 
Observe that we have $\Delta_{l^*}^{p_k}\le \Delta^*(M(p_k))$ and the counter $c^\est(p_k)$ never go beyond $M(p_k)$, we thus have
\begin{align}
    L \cdot\eqref{lem_4_8_eq} &\le L(C_M^\est(\Delta^*(M(p_k))) - C_M^\est(\Delta_{l^*-1}^{p_k}))\cdot \Delta^*(M(p_k)) + \sum_{j\in[l^* - 1]}L(C_M^\est(\Delta^{p_k}_j) - C_M^\est(\Delta^{p_k}_{j-1}))\cdot \Delta_j^{p_k} \nonumber\\
    &\le LC_M^\est(\Delta^*(M(p_k)))\cdot \Delta^*(M(p_k)) + L\int_{\Delta^*(M(p_k))}^{\Delta_{\max}^{p_k}} C_M^\est(x)dx \nonumber\\
    & \le \cO\left(\sqrt{\frac{K\ln (T\rho) \cdot LM(p_k)}{\rho}}\right) .
    \label{up_case_2_hist}
\end{align}
Combining~\eqref{up_case_1_hist} and \eqref{up_case_2_hist}, and with Jesen's inequality and $\sum_{p_k} M(p_k) \le KM/\epsilon$ will give us desired result. 
Put all pieces together, we have the instance-independent regret bound as stated in the lemma.
The final inequality does not depend on the event $\cE(p_k)$, we thus can drop this conditional expectation.
\end{proof} 
Combining with the discretization error, we have 
\begin{align*}
    \reg(T )\le \cO\left(K\cdot \sqrt{{T\ln (T\rho)}/{(\rho\epsilon)}}  + K/(L\epsilon\rho^2)\right) + \cO(K\epsilon T).
\end{align*}
Picking 
\begin{align*}
    \epsilon = \cO\left(\frac{\ln(T\rho)}{T\rho}\right)^{1/3}; \quad s_a = \cO\left(\frac{1/3\ln\left(\frac{\ln(T\rho)}{T\rho}\right) - \ln (L^*K)}{\ln \gamma}\right).
\end{align*}
We will obtain the results as stated in the theorem.

\section{Lower Bound of Action-Dependent Bandits} \label{sec:lower_bound}
In this section,
we derive the lower regret bound of bandits with action-dependent feedback, 
showing that the upper regret bound of our Algorithm~\ref{algo:action_UCB} is optimal in the sense that it matches this lower bound in terms of the dependency on $T$ and $K$.
Note that, by importance-weighting technique, we can construct an unbiased estimation of each base arm's reward. 
In the below discussion, we rephrase our problem as the {\em combinatorial Lipschitz bandit with constraint}, henceforth called~\CombLipBC, which directly operates on the observations of all base arms:
\begin{definition} [\CombLipBC]
Let action set $\cP$ available to the learner be a continuous space, consisted of $K$ unit-range base arms, i.e., $\cP \subset [0,1]^K$.
At each time, the learner needs to select a meta arm $\vp(t) = \{p_1(t), \ldots, p_K(t)\}$ in which each discretized arm $p_k(t)\in[0,1]$ is selected from $k$-th unit range, with the constraint such that $\sum_k p_k(t) = 1$.
And then the learner will observe rewards $\{\rewob_t(p_k(t))\}_{k\in[K]}$ for all base arms with the mean of each $\Esymb [\rewob_t(p_k(t))] = \rewfn_k(p_k(t))$.
\end{definition}
Our main result of this section is summarized in the following theorem:
\begin{theorem} \label{theorem:lower_bound}
Let $T\!>\!2K$ and $K\!\geq\!4$, there exists a problem instance such that for any algorithm $\cA$ for our \action~, we have $\inf_{\cA} \reg(T) \geq \Omega(K T^{2/3})$.
\end{theorem}
The high-level intuition for deriving the above lower bound is that we first construct a reduction from $\CombLipBC$ to a discretized {\em combinatorial bandit problem with the action constraint} $\sum_k p_k(t) = 1$ - we refer to this latter problem setting as $\CombBC$.
Then we show that the regret incurred within $\CombLipBC$ is lower bounded by the regret incurred with $\CombBC$.
To finish the proof, we bound the worst-case regret from below of $\CombBC$ by taking an average over a conveniently chosen class of problem instances.

\subsection{Randomized problem instances and definitions}
We now construct a reduction for proving the lower bound of $\CombLipBC$.
Specifically, we will construct a distribution $\cD$ over a set of problem instances 
(we also call each instance an adversary, since the instances are adversarially constructed) 
of $\CombLipBC$, while each problem instance will be uniquely mapped to a problem instance in $\CombBC$.
The construction is similar to the one used in~\citep{kleinberg2008multi}.

These new instances are associated with $0-1$ rewards. 
For each base arm $k\in [K]$, all the discretized arms $p$ have mean reward $\rewfn_k(p) = 1/2$ except those near the unique best discretized arm $p_k^*$ with $\rewfn_k(p_k^*) = 1/2+\epsilon$. 
Here $\epsilon > 0$ is a parameter to be adjusted later in the analysis. 
Due to the requirement of Lipschitz condition, a smooth transition is needed in the neighborhood of each $p_k^*$. 
More formally, we define the following function $\rewfn_k(\cdot)$ for base arm $k$:
\begin{equation}\label{eq: bump_reward}
  \rewfn_k(p)  = \left \{
  \begin{aligned}
    & 1/2, && \forall p\in[0,1]: |p - p_k^*| \geq \epsilon/L_k \\
    & 1/2 + \epsilon - L_k\cdot|p - p_k^*|, && \forall p\in[0,1]: |p - p_k^*| < \epsilon/L_k
  \end{aligned} \right.
\end{equation} 
Fix $\Np  \in \Nsymb$ and partition all base arms $[0,1]$ into $\Np$ disjoint intervals of length $1/\Np$.
Then the above functions indicate that each interval with the length of $2\epsilon$ will either contain a bump or be completely flat. 
For the sake of simplifying presentation, in the analysis below, we'll focus on the case where the Lipschitz constant is $L_k = 1, \forall k\in[K]$. 
Formally, 
\begin{definition} 
We define 0-1 rewards problem instances $\cI(\vp^*, \epsilon)$ for \textup{\CombLipBC} indexed by a random permutation $\vp^* = \{p_k^*\}_{k\in[K]}$, which satisfies following property:
\squishlist
    \item $\sum_{k} p_k^* = 1$ and each $p_k^*$ takes the value from $\{(2j-1)\epsilon\}_{j\in[\Np]}$.
    \item The reward function of base arm $k$ is defined in (\ref{eq: bump_reward}), and the optimal action of arm $k$ is $p_k^*$.
\squishend
\end{definition}
In combinatorial bandits, the learner selects a subset of ground arms subject to some pre-defined  constraints. 
Adapting to our model, we denote this discretized action space $\cM$ as the set of $K\times \Np $ binary matrices $\{0,1 \}^{K\times \Np }$:
\begin{align*}
\cM = \{\va \in \{0,1\}^{K \times \Np }: \forall k\in[K], \sum_{j=1}^{\Np}a_{k,j} = 1\},
\end{align*}
where $a_{k,j}\in\{0,1\}$ is the indicator random variable such that $a_{k,j} = 1$ means that the $j$-th discretized arm probability is selected for the $k$-th base arm.
Note that this space has not included the action constraint that we're planning to impose on  \CombLipBC. 

We now construct the problem instances for $\CombBC$ such that each problem instance $\cI(\vp^*, \epsilon)$ in $\CombLipBC$ has a corresponding problem instance in $\CombBC$.
\begin{definition} 
We define 0-1 rewards problem instances $\cJ(\vl^*, \epsilon)$ for \textup{\CombBC} indexed by $\vl^*= \{l^*_k\}_{k\in[K]}$, such that $l_k^* = (p_k^*/\epsilon + 1)/2$. 
Therefore, $l^*_k\in [\Np]$
and the mean reward of $\cJ(\vl^*, \epsilon)$ is defined as follows:
for any $t\in\{1, \ldots, T\}$, 
\begin{equation}\label{reward_fn_CombBC}
\Esymb[\rewob_t(l_k(t))]  = \left \{
  \begin{aligned}
    & 1/2, && l_k(t) \neq l^*_k \\
    & 1/2+\epsilon, && l_k(t) = l^*_k
  \end{aligned} \right.
\end{equation} 
\end{definition}
Observe that with one more action constraint, the feasible action space of $\CombBC$ will be a constrained space of $\cM$, which we denote by
$\Pi = \{\va  \in \{0,1\}^{K \times \Np }: \forall k\in[K], \sum_{j=1}^{\Np}a_{k,j} = 1, \sum_{i=1}^K l_{k,j}a_{k,j} = K-1 + \Np\}$.

We now next show that for any algorithm $\cA_{\cI}$ trying to solve the problem instance $\cI(\vp^*, \epsilon)$ in \CombLipBC, we can construct an algorithm $\cA_{\cJ}$ that needs to solve a corresponding problem instance $\cJ(\vp^*, \epsilon)$ in \CombBC.

The intuition of the construction routine is as follows. 
With the above defined $K\Np $ intervals in hand and the deliberately designed reward structure, whenever an algorithm chooses a meta arm $\vp = \{p_1, \ldots, p_K\}$ such that each discretized arm 
$p_k$ falls into an interval of this base arm $k$, choosing the center of this interval is best.
Thus, if we restrict to discretized arms that are centers of the intervals of all base arms, we then have a family of problem instances of \CombBC, where the reward function is exactly defined in (\ref{reward_fn_CombBC}).

\begin{routine}[H]
\renewcommand{\thealgorithm}{}
\caption{A routine inbetween $\cA_{\cI}$ and $\cA_{\cJ}$} \label{routine: ALG to ALG_J}
\begin{algorithmic}\label{routine}
\STATE \textbf{Input:} A $\CombLipBC$ instance $\cI$, a $\CombBC$ instance $\cJ$ and an algorithm $\cA_{\cI}$ for solving $\cI$.
\FOR{round $t = 1, \ldots $}
    \STATE $\cA_{\cI}$ selects a meta arm $\vp(t) = \{p_1(t),\ldots,p_K(t)\}$;
    \STATE $\cA_{\cJ}$ selects arm $\vl(t) = \{l_1(t),\ldots,l_K(t)\}$ such that $p_k(t) \in \big[(2l_k(t)-1)\epsilon - \epsilon, (2l_k(t)-1)\epsilon + \epsilon\big), \forall k \in [K]$;
    \STATE $\cA_{\cJ}$ observes $\{\rewob(l_k(t))\}$;
    \STATE $\cA_{\cI}$ observes $\{\rewob(p_k(t))\}$;
\ENDFOR
\end{algorithmic}
\end{routine}

Furthermore, with above construction routine, we have following guarantee:
\begin{lemma} \label{lemma: regret_mapping}
The regret incurred by $\cA_{\cI}$, which is for the problem instance $\cI(\vp^*, \epsilon)$, is lower bounded by the regret incurred by $\cA_{\cJ}$ for the problem instance $\cJ(\vl^*, \epsilon)$:
\begin{align} \label{ineq: regret_relations}
\Esymb[\reg_{2\epsilon}(T)|\cI, \cA_{\cI}] \geq \Esymb[\reg_{2\epsilon}(T)|\cJ, \cA_{\cJ}].
\end{align}
\end{lemma}
\begin{proof}
As we can see, each instance $\cJ(\vl^*, \epsilon)$ corresponds to an instance $\cI(\vp^*, \epsilon)$ of \CombLipBC.
In particular, each $k$-th base arm in $\cJ$ corresponds to the base arm $k$ in $\cI$, and more specifically, each discretized arm $j\in[\Np ]$ in $k$-th base arm corresponds to the all possible discretized arms $p$ such that $p \in [(2j-1)\cdot \epsilon -\epsilon, (2j-1)\cdot \epsilon + \epsilon)$.
In other words, we can view $\cJ$ as a discrete version of $\cI$. 
In particular, we have $\rewfn_{k}(j|\cJ) = \rewfn_k(p), \forall p \in [(2j-1)\epsilon -\epsilon, (2j-1)\epsilon + \epsilon)$, where $\rewfn_k(\cdot)$ is the reward function for base arm $k$ in $\cI$, and $\rewfn_{k}(\cdot|\cJ)$is the reward function for base arm $k$ in $\cJ$.

Given an arbitrary algorithm $\cA_{\cI}$ for a problem instance $\cI$ of \CombLipBC, we can use it to construct an algorithm $\cA_{\cJ}$ to solve the corresponding problem instance $\cJ$ in \CombBC.
To see this, at each round, $\cA_{\cI}$ is called and an action is selected $\vp(t)$. 
This action corresponds to an action $\vl(t)$ in $\CombBC$ such that for each discretized arm $p_k(t) \in \vp(t)$, it falls into the interval $[(2l_k(t)-1)\epsilon - \epsilon, (2l_k(t)-1)\epsilon + \epsilon)$ where $l_k(t)\in \vl(t)$.
Then algorithm $\cA_{\cJ}$ will observe $\{\rewob(l_k(t))\}$ and receive the reward $\sum_k \rewob(l_k(t))$.  
After that, $\sum_{k}\rewob(l_k(t))$ and $\vp(t)$ will be further used to compute reward $\sum_{k}\rewob(p_k(t))$ such that $\Esymb[\sum_{k}\rewob(p_k(t))] = \sum_{k\in[K]} \rewfn(l_k(t))$, and feed it back to $\cA_{\cI}$. 

At each round, let $\vp(t)$ and $\vl(t)$ denote the action chosen by the $\cA_{\cI}$ and $\cA_{\cJ}$, since we have $\rewfn_k(l_k(t)) \geq \rewfn_k(p_k(t))$ and best arm of the problem instance $\cI$ and $\cJ$ has the same mean reward $K(1/2 + \epsilon)$, this completes the proof.
\end{proof}

\subsection{Lower bound the \texorpdfstring{$\Esymb[\reg_{2\epsilon}(T)|\cJ, \cA_{\cJ}]$}{Lg}}
With Lemma \ref{lemma: regret_mapping} stating the relationship between $\Esymb[\reg_{2\epsilon}(T)|\cI, \cA_{\cI}]$ and $\Esymb[\reg_{2\epsilon}(T)|\cJ, \cA_{\cJ}]$ as derived in (\ref{ineq: regret_relations}), we can lower bound the $\Esymb[\reg_{2\epsilon}(T)|\cA_{\cI}]$ via deriving the lower bound for $\Esymb[\reg_{2\epsilon}(T)|\cA_{\cJ}]$.

The structure of the proof is similar to that of \cite{audibert2010regret}, while the main difference is that we construct a different set of adversaries to bound the  probability of the learner on achieving ``good event'' (will be specified later).
At a high level, our proof builds on the following 4 steps: from step 1 to 3 we restrict our attention to the case of deterministic strategies for the learner,
and then we show how to extend the results to arbitrary and randomized strategies by Fubini's theorem in step 4. 

\paragraph{Step 1: Regret Notions.}
We will also call that the learner is playing against the $\vl^*$-adversary when the current instance is $\cJ(\vl^*, \epsilon)$.
We denote by $\Esymb_{\vl^*}[\cdot]$ the expectation with respect to the reward generation process of the $\vl^*$-adversary.
Without the loss of generality, we assume $K$ is an even number.
We write $\Psymb_{(2h-1, 2h), \vl^*}$ for the probability distribution of $(j_{2h-1,t}, j_{2h,t})$ when the learner faces the $\vl^*$-adversary.
Thus, against the $\vl^*$-adversary, we have
\begin{align*}
\Esymb_{\vl^*}[\reg_{2\epsilon}(T)] = \Esymb_{\vl^*} \sum_{t = 1}^T \sum_{h= 1}^{K/2} 2\epsilon \mathbbm{1}(\{j_{2h-1,t} \neq l_{2h-1}^*, j_{2h,t} \neq l_{2h}^*\}) = T\cdot 2\epsilon\sum_{h=1}^{K/2}\bigg(1 - \Psymb_{(2h-1, 2h), \vl^*}(G_T)\bigg),
\end{align*}
where $G_T$ denotes the \emph{good event} such that $\{j_{2h-1,T} = l_{2h-1}^*, j_{2h,T} = l_{2h}^*\}$ holds simultaneously for base arm $2h-1$ and $2h$.
For a particular distribution $\vl^* \sim \cD$ for all random adversaries, and let $\Psymb(\vl^*)$ denote the support of the adversary $\vl^*$. 
Because the maximum value is always no less than the mean, we have
\begin{align} \label{eq: regret_lower_bound}
\sup_{\vl^*\in \cJ_\epsilon}\Esymb_{\vl^*}[\reg_{2\epsilon}(T)] & \geq  T\cdot 2\epsilon\sum_{h=1}^{K/2}\bigg(1 - \sum_{\vl^* \in \cJ_\epsilon}\Psymb(\vl^*)\cdot\Psymb_{(2h-1, 2h), \vl^*}(G_T)\bigg).
\end{align}

\paragraph{Step 2: Information Inequality}
Let $\Psymb_{-(2h-1, 2h), \vl^*}$ be the probability distribution of $(j_{2h-1,t} , j_{2h,t})$ against the adversary which plays like the $\vl^*$-adversary except that in the $(2h-1, 2h)-$th base arms, where the rewards of all discretized arms are drawn from a Bernoulli distribution of parameter $1/2$. 
We refer to it as $(-h, \vl^*)$-adversary. 
Let $\cJ_\epsilon$ denote the set of all possible $\vl^*$ adversaries and $\cD$ be the distribution over $\vl^*$ in which $\vl^*$ is sampled uniformly at random.
\begin{lemma}
Let $n_{-h}(K - 1 + \Np - m ), \forall m\in\{2,\ldots, 1 + \Np\}$ denote the total number of the combinations of $\big(j_k\big)_{k\neq 2h-1, 2h}$ such that $\sum_{i\neq 2h-1, 2h} j_k = K - 1 + \Np - m $.
Then we have 
\begin{align} \label{ineq: pinsker}
\frac{1}{|\cJ_\epsilon|}\sum_{\vl^*\in \cJ_\epsilon} \Psymb_{(2h-1, 2h), \vl^*}(G_T) & \leq \sum_{m=2}^{\Np+1} \frac{n_{-h}(K\!-\!1\!+\!\Np\!-\!m)}{|\cJ_\epsilon|} + c\epsilon\sqrt{\frac{T}{|\cJ_\epsilon|} \sum_{m=2}^{\Np+1} n_{-h}(K\!-\!1\!+\!\Np\!-\!m)}, 
\end{align}
where $c$ is a constant.
\end{lemma} 
\begin{proof}
Let \textup{\KL}$(\cdot)$ be the Kullback-Leibler divergence operator.
By Pinsker's inequality, we have
\begin{align*}
\Psymb_{(2h-1, 2h), \vl^*}(G_T) \leq \Psymb_{-(2h-1, 2h), \vl^*}(G_T) + \sqrt{\frac{1}{2}\KL(\Psymb_{-(2h-1, 2h), \vl^*}, \Psymb_{(2h-1, 2h), \vl^*})}, \quad \forall \vl^* \in \cJ_\epsilon.
\end{align*}
Then by the concavity of the square root,
\begin{align*}
&\frac{1}{|\cJ_\epsilon|}\sum_{\vl^*\in \cJ_\epsilon} \Psymb_{(2h-1, 2h), \vl^*}(G_T) \\
\leq ~&\frac{1}{|\cJ_\epsilon|}\sum_{\vl^*\in \cJ_\epsilon} \Psymb_{-(2h-1, 2h), \vl^*}(G_T)  + \sqrt{\frac{1}{2|\cJ_\epsilon|}\sum_{\vl^*\in \cJ_\epsilon}\KL(\Psymb_{-(2h-1, 2h), \vl^*}, \Psymb_{(2h-1, 2h), \vl^*})}.
\end{align*}
We introduce $n_h(m), \forall m\in\{2,\ldots, 1 + \Np\}$  to denote the total number of combinations of $(j_{2h-1}, j_{2h})$ such that $j_{2h-1} + j_{2h} = m$.
Then by definition, it is easy to see that $n_h(m) = m - 1$, and furthermore
\begin{align} \label{eq: relation_statisitcs}
\sum_{m=2}^{\Np+1} n_h(m) \cdot n_{-h}(K - 1 + \Np - m) = |\cJ_\epsilon|. 
\end{align}
Let $\cD$ be the distribution over $\vl^*$ in which $\vl^*$ is sampled uniformly at random, i.e., $\Psymb(\vl^*) = \frac{1}{|\cJ_\epsilon|}$, then by the symmetry of the adversary $(-h, \vl^*)$, we have 
\begin{align*}
\sum_{\vl^*\in \cJ_\epsilon} \Psymb(\vl^*) \cdot \Psymb_{-(2h-1, 2h), \vl^*}(G_T) & = \sum_{m=2}^{\Np+1} \sum_{\vl^*: \sum_{k\neq 2h-1, 2h}l_k^* = K - 1 + \Np - m } \Psymb(\vl^*) \cdot \Psymb_{-(2h-1, 2h), \vl^*}(G_T) \\
& = \sum_{m=2}^{\Np+1}  \frac{1}{n_h(m)}\sum_{\vl^*: \sum_{k\neq 2h-1, 2h}l_k^* =K - 1 + \Np - m} \Psymb(\vl^*) \\
& = \sum_{m=2}^{\Np+1}  \frac{1}{n_h(m)}\frac{n_h(m) \cdot n_{-h}(K - 1 + \Np - m )}{|\cJ_\epsilon|} \\
 & = \sum_{m=2}^{\Np+1} \frac{n_{-h}(K - 1 + \Np - m)}{|\cJ_\epsilon|}. & \tag*{By (\ref{eq: relation_statisitcs})}
\end{align*}
\end{proof}

\paragraph{Step 3: Bounding $\KL(\Psymb_{-(2h-1, 2h), \vl^*}, \Psymb_{(2h-1, 2h), \vl^*})$ via the chain rule.}
We now proceed to bound the value of $\KL(\Psymb_{-(2h-1, 2h), \vl^*}, \Psymb_{(2h-1, 2h), \vl^*})$.
\begin{lemma}
\textup{\KL}$( \Psymb^T_{-(2h-1, 2h), \vl^*}, \Psymb^T_{(2h-1, 2h), \vl^*}) \leq \frac{c\epsilon^2T}{1-4\epsilon^2} \Psymb_{-(2h-1, 2h), \vl^*}(G_T)$, where $c$ is the constant value.
\end{lemma}

\begin{proof}
Given any sequence of observed rewards up to time $T$, which denoted by $W_T \in \{1,\ldots, K\}^T$, the empirical distribution of plays, and, in particular, the probability distribution of $(j_{2h-1,t}, j_{2h,t})$ conditional on the fact that $W_T$ will be the same for all adversaries. 
Thus, if we denote by $\Psymb^T_{(2h-1, 2h), \vl^*}$ (or $\Psymb^T_{-(2h-1, 2h), \vl^*}$) the probability distribution of $W_T$ when the learner plays against the $\vl^*$-adversary (or the $(-h, \vl^*)$-adversary), we can easily show that $\KL(\Psymb_{-(2h-1, 2h), \vl^*}, \Psymb_{(2h-1, 2h), \vl^*}) \leq \KL(\Psymb^T_{-(2h-1, 2h), \vl^*}, \Psymb^T_{(2h-1, 2h), \vl^*})$. 
Then we apply the chain rule for Kullback-Leibler divergence iteratively to introduce the probability distributions $\Psymb^t_{(2h-1, 2h), \vl^*}$ of the observed rewards $W_t$ up to time $t$ and then will arrive desired result.
More formally, we reduce to bound the $\KL(\Psymb^T_{-(2h-1, 2h), \vl^*}, \Psymb^T_{(2h-1, 2h), \vl^*})$, 
\begin{align*}
& \quad \KL( \Psymb^T_{-(2h-1, 2h), \vl^*}, \Psymb^T_{(2h-1, 2h), \vl^*}) \\
= &\quad  \KL( \Psymb^1_{-(2h-1, 2h), \vl^*}, \Psymb^1_{(2h-1, 2h), \vl^*}) +  \\
&\quad \sum_{t=2}^T\sum_{w_{t-1}\in\{1,\ldots, K\}^{t-1}} \Psymb^{t-1}_{-(2h-1, 2h),\vl^*}(w_{t-1}) \KL\big(\Psymb_{-(2h-1, 2h), \vl^*}(\cdot|w_{t-1}), \Psymb_{(2h-1, 2h), \vl^*}(\cdot|w_{t-1})\big) \\
= & \quad\KL(\cB_{\emptyset}, \cB^{'}_{\emptyset})\mathbbm{1}(j_{2h-1, 1} = l_{2h-1}^*, j_{2h, 1} = l_{2h}^*) +  \\
& \quad \sum_{t=2}^T\sum_{w_{t-1}: j_{2h-1, t-1} = l_{2h-1}^*, j_{2h, t-1} = l_{2h}^* } \Psymb^{t-1}_{-(2h-1, 2h),\vl^*}(w_{t-1}) \KL(\cB_{w_{t-1}}, \cB^{'}_{w_{t-1}}) \\
= & \quad\KL(\cB_{\emptyset}, \cB^{'}_{\emptyset})\mathbbm{1}(G_1) +  \sum_{t=2}^T\sum_{w_{t-1}: G_{t-1}} \Psymb^{t-1}_{-(2h-1, 2h),\vl^*}(w_{t-1}) \KL(\cB_{w_{t-1}}, \cB^{'}_{w_{t-1}}),
\end{align*}
where $\cB_{w_{t-1}}$ and $\cB^{'}_{w_{t-1}}$ are two Bernoulli random variables with parameters in $\{1/2, 1/2 + \epsilon\}$.
Due to the fact that $\KL(p, q) \leq \frac{(p-q^2)}{q(1-q)}$, we will have
\begin{align*}
\KL(\cB_{w_{t-1}}, \cB^{'}_{w_{t-1}}) \leq c\frac{\epsilon^2}{1-4\epsilon^2},
\end{align*}
where $c$ is a constant. 
Taking the summation will complete the proof.
\end{proof}

\paragraph{Wrapping up: Proof of Theorem \ref{theorem:lower_bound} on Deterministic Strategies.} 
Observe that we can bound 
\begin{align*}
\sum_{m=2}^{\Np+1} n_{-h}(K - 1 + \Np - m)/|\cJ_\epsilon|  = \Omega(1/\Np),
\end{align*}
which follows the fact that: given $a_1\leq a_2 \leq \ldots \leq a_n$ and $b_1\leq b_2 \leq \ldots \leq b_n$, one will have $n\sum a_i b_i \geq \sum a_i \sum b_i$.
Plugging back into Eqs. (\ref{ineq: pinsker}) and (\ref{eq: regret_lower_bound}) and substituting $\epsilon = \Theta(T^{-1/3})$ will get the desired result.
\paragraph{Step 4: Fubini's theorem for Random Strategies.}
For a randomized learner, let $\Esymbr$ denote the expectation with respect to the randomization of the learner. 
Then 
\[
\frac{1}{|\cJ_\epsilon|}\sum_{\vl^*\in \cJ_\epsilon} \Esymb \sum_{t=1}^T
\big(\vl(t)^T \vr_t - (\vl^*)^T\vr_t\big) = \Esymbr \frac{1}{|\cJ_\epsilon|}\sum_{\vl^*\in \cJ_\epsilon} \Esymb_{\vl^*} \sum_{t=1}^T
\big(\vl(t)^T \vr_t - (\vl^*)^T\vr_t\big).
\]
where $\vr_t = (r_1(l_1(t), \ldots, r_K(l_K(t)))$, and value of the reward for not realized arms are computed from Eq (\ref{eq:reward_impor_weights}).
The interchange of the integration and the expectation is justified by Fubini's Theorem.
For every realization of learner's randomization, the results of all earlier steps still follow.
This will give us the same lower bound for $\Esymbr \frac{1}{|\cJ_\epsilon|}\sum_{\vl^*\in \cJ_\epsilon} \Esymb_{\vl^*} \sum_{t=1}^T
\big(\vl(t)^T \vr_t - (\vl^*)^T\vr_t\big)$ as we have shown above.

\section{Proof of the lower bound in \history}\label{sec:proof_lower_his} 
For \history, we show that for a general class of utility function which satisfies the \emph{strictly proper} property (we will shortly elaborate this property), solving \history~is as least hard as solving \action.
Armed with the above derived lower bound of action-dependent case, we can then conclude the lower bound of history-dependent case.
\emph{Strictly Proper Utility Function} is defined as below.
\begin{definition}[Strictly Proper Utility] \label{defn_strict_proper_utility}
For any mixed strategy $\vp \in \cP$ and any $\vq \neq \vp$, the functions $\{\rewfn_k\}$ are strictly proper if following holds,
\begin{align}
\sum_{p_k\in\vp} p_k \rewfn_k(p_k) > \sum_{p_k \in \vp, q_k \in \vq}p_k \rewfn_k(q_k).
\end{align} 
\end{definition}
With above defined strictly proper utility at hand, we now ready to prove the Theorem~\ref{theorem_low_bound_all} for history-dependent case.
\begin{proof}
Let $\cI^h$ denote a \history~instance whose utility function satisfies above defined strictly proper property, and $\cI^a$ denote the associated action-dependent bandit instance whose utility function is the same as that in $\cI^h$.
Let $\vf^*(t) = \{f_k^*\}_{k\in[K]}$ be the discounted frequency at time $t$ when the learner keeps deploying the best-in-hindsight strategy $\vp^*$ and $L^*  = \max L_k$.
Then we can show that
\begin{align*}
\E[\reg(T)|\cI^h]  & = \sum_{t = 1}^T U_t(\vp^*) - \sum_{t= 1}^T U_t(\vp(t))\\
& = \sum_{t = 1}^T \sum_{p_k^* \in \vp^*}p_k^*\cdot\rewfn_k(f^*_k(t)) - \sum_{t= 1}^T \sum_{k}p_k(t) \cdot\rewfn_k(f_k(t)) \\
& > \sum_{t = 1}^T \sum_{p_k^* \in \vp^*}p_k^*\cdot\rewfn_k(f^*_k(t)) - \sum_{t= 1}^T \sum_{k}p_k(t) \cdot\rewfn_k(p_k(t)) \\
& \geq \sum_{t = 1}^T \sum_{p_k^* \in \vp^*}p_k^*\cdot\rewfn_k(p_k^*) - \frac{\gamma^2(1-\gamma^{2T-2})KL^*}{1-\gamma^2} - \sum_{t= 1}^T \sum_{k}p_k(t) \cdot\rewfn_k(p_k(t)) \\
& = \E[\reg(T)|\cI^a] - \frac{\gamma^2(1-\gamma^{2T-2})KL^*}{1-\gamma^2} = \Omega(KT^{2/3}),
\end{align*}
where the first inequality is due to the strict proper property of utility function, and the third inequality is due to the fact that the \history~shares the same best-in-hindsight strategy as that in the action-dependent bandit and Lemma \ref{lemma: approx_err_neq_1}. 
By the regret reduction from the \history~to the action-dependent bandit, we can conclude the lower bound of the history-dependent case.
\end{proof}

\section{Optimal dynamic policy v.s. best policy in hindsight} \label{optimal_dynamic_vs_best_fixed}
As we mentioned, for \action, the optimal dynamic policy can be characterized by a best-in-hindsight (mixed) strategy computing from following constrained optimization problem: $\max_{\vp \in \cP} \sum_{k=1}^K p_k \rewfn_k(p_k)$.
While for \history, it is possible that the optimal policy $\vp^*$ may not be well-defined due to the fact of reward dependence on action history.
However, we argue that when competing against with best-in-hindsight policy, 
notwithstanding in the face of this kind of reward-history correlation, the value of the optimal strategy is always well-defined in the limit, and this limit value is also characterized by the best-in-hindsight (mixed) strategy computed from \action.
To gain intuition, note that the time-discounted frequency $\bm{\mathrm{f}}(t)$ will be exponentially approach to the fixed strategy $\vp$ the learner deploys.
As we explain in Section \ref{sec:history}, after consistently deploying $\vp$ with $s$ rounds, the frequency $\vf(t+s)$ will be converging to $\vp$ with the exponential decay error $\gamma^s$.
Thus, to achieve highest expected reward, the learner should deploy the optimal strategy computed as in action-dependent case.

\section{EVALUATIONS}\label{missing_exps}
We conducted a series of simulations to  empirically evaluate the performance of our proposed solution with a set of baselines.

\subsection{Evaluations for \action}
We first evaluate our proposed algorithm on \action~against the following state-of-the-art bandit algorithms.
\squishlist
\item \textbf{EXP3}:
One natural baseline is applying EXP3 \citep{auer2002nonstochastic} on the space of base arms.
While EXP3 is designed for adversarial rewards, it is competing with the best fixed arm in hindsight and might not work well in our setting since the optimal strategy is randomized.

\item \textbf{EXP3-Meta-Arm (mEXP3)}: 
To make a potentially more fair comparison, we also implement EXP3 on the meta-arm space. 
We denote it as mEXP3 in the following discussion.
\item \textbf{CUCB} \citep{chen2016combinatorial,wang2017improving}:
This algorithm is designed to solve combinatorial semi-bandit problem, 
which chooses $m$ arms out of $M$ arms at each round and receives only the rewards of selected arms. 
Mapping to our setting, $M = K / \epsilon$ represents the total number of discretized arms, 
$m=K$ is the number of base arms, and the selection of $m$ arms is constrained to satisfy the probability simplex constraint.
\squishend
In the simulations, we set $K = 2$ for simplicity.
Moreover, $\rewfn_k(p_k)$ is chosen such that $\rewfn_k(p_k)$ is maximized when $0<p_k<1$. 
In particular, we define $\rewfn_k(p_k)$ as
a scaled Gaussian function : $\rewfn_k(p_k) = f(p_k|\tau_k, 0.5)/C_k$, where $f(x|\tau, \sigma^2)$ is the pdf of Gaussian distribution with the mean $\tau$ and variance $\sigma^2$, and $C_k = f(\tau_k|\tau_k, \sigma^2)$ is a constant ensuring $\rewfn_k(p_k) \in [0,1], \forall p_k\in[0,1]$. 
For each arm $k$, $\tau_k$ is uniformly draw from 0.45 to 0.55 and the instantaneous reward is drawn from a Bernoulli distribution with the mean of $\rewfn_k(p_k(t))$, i.e., $\rewob_t(p_k(t)) \sim \Ber(\rewfn_k(p_k(t)))$.
And the ratio $\rho$ is set to 0.2.
For each algorithm we perform 40 runs for each of independent 40 values of the corresponding parameter, and we report the averaged results of these independent runs, where the error bars correspond to $\pm2$ standard deviations.

The results, shown in Figure~\ref{fig_1}, demonstrate that our algorithm significantly outperforms the baselines.
As expected, mEXP3 works better than EXP3 algorithm when $T$ is large, since the former searches the optimal strategies in the meta arm space. 
Our algorithm outperforms mEXP3 and CUCB since we utilize the problem structure, which reduces the amount of explorations.

\subsection{Evaluations for \history}
We now evaluate our proposed algorithm for \history~via comparing against the following baselines from non-stationary bandits.
Note that, while CUCB performs reasonably well in action-dependent case, it does not apply in history-dependent case, since we cannot \emph{select} the time-discounted frequency (which maps to the arm in CUCB) as required in CUCB.
\squishlist

\item \textbf{Discounted UCB} \citep{garivier2011upper,kocsis2006discounted}:
Discounted UCB (DUCB) is an adaptation of the standard UCB policies that relies on a discount factor $\gamma_\DUCB \in (0,1)$.
This method constructs an $\UCB: \bar{\rewfn}_{t}(k, \gamma_\DUCB) + c_t(k,\gamma_\DUCB)$ for the instantaneous expected reward, 
where the confidence is defined as $c_t(k, \gamma_\DUCB) = 2\sqrt{\frac{\xi \ln(n_t)}{N_t(k, \gamma_\DUCB)}}$, for an appropriate parameter $\xi$, 
$N_t(k, \gamma_\DUCB) = \sum_{s=1}^t\gamma_\DUCB^{t-s} \mathbbm{1}_{(a_s = k)}$,
and the discounted empirical average is given by
$\bar{\rewfn}_{t}(k, \gamma_\DUCB) = \frac{1}{N_t(k, \gamma_\DUCB)} \cdot\sum_{s =1}^t\gamma_\DUCB^{t-s} \rewob_{s}(k)\mathbbm{1}_{(a_s = k)}$.

\item \textbf{Sliding-Window UCB} \citep{garivier2011upper}:
Sliding-Window UCB (SWUCB) is a modification of DUCB, instead of averaging the rewards over all past with a discount factor, SWUCB relies on a local empirical average of the observed rewards, for example, using only the $\tau$ last plays.
Specifically, this method also constructs an $\UCB: \overbar{\rewfn}_{t}(k, \tau) + c_t(k,\tau)$ for the instantaneous expected reward. 
The local empirical average is given by
$\overbar{\rewfn}_{t}(k, \tau) = \frac{1}{N_t(k, \tau)}\sum_{s =t-\tau+1}^t \rewob_{s}(k)\mathbbm{1}_{(a_s = k)}$, $N_t(k, \tau) = \sum_{s =t-\tau+1}^t\gamma_\DUCB^{t-s} \mathbbm{1}_{(a_s = k)}$ and the confidence interval is defined as $c_t(k, \gamma_\DUCB) = 2\sqrt{\frac{\xi \ln(\min(t, \tau))}{N_t(k, \tau)}}$.
\squishend
We use \emph{grid searches} to determine the algorithms' parameters.
For example, in DUCB, the discount factor was chosen from $\gamma_\DUCB\in\{0.5, 0.6, \ldots 0.9\}$, while the window size of SWUCB was chosen from $\tau\in\{10^2,\ldots, 5\times 10^2\}$.
Besides above algorithms, we also implement the celebrated non-stationary bandit algorithm EXP3.

We chose $K$ and $\rewfn_k(p_k)$ to be the same as the experiments in action-dependent case.
And the discount factor is chosen as $\gamma_\DUCB = 0.8$ and the window size for SWUCB is chosen as 200 via the grid search, and $\xi$ is set to 1.
We examine the algorithm performances under different $\gamma$ (the parameter in time-discounted frequency), 
with smaller $\gamma$ indicating that arm rewards are more influenced by recent actions. 
As seen in Figures~\ref{fig:gamma_0.2}-\ref{fig:gamma_0.6}, our algorithm outperforms all baselines in all $\gamma$ but the improvement is more significant with small $\gamma$.
This is possibly due to that most non-stationary bandit algorithms have been focusing on settings in which the change of arm rewards over time is not dramatic. 

We also examine our algorithm with larger number of base arms $K$ with comparing to above baseline algorithms and the performance of our algorithm on different ratios $\rho$. 
The results are presented in Figure~\ref{fig:K_and_ratio}
and show that our algorithm consistently performs better than other baselines when $K$ goes large.
The results also suggest that our algorithm is not sensitive to different $\rho$, though one could see the regret is slightly lower when $\rho$ is increasing, which is expected from our regret bound.

\end{document}